\documentclass[manuscript,screen]{acmart}

\usepackage{tikz}
\usetikzlibrary{shapes,arrows.meta,chains,decorations.pathreplacing}
\usepackage{array}
\usepackage{venndiagram}
\usepackage{xcolor}
\usepackage{enumitem}

\usetikzlibrary{shapes.geometric, arrows.meta, positioning}

\definecolor{lightgray}{RGB}{211, 211, 211}
\definecolor{customPurple}{RGB}{123, 104, 238} % Replace with actual RGB values
\definecolor{customGreen}{RGB}{144, 238, 144} % Replace with actual RGB values
\definecolor{customOrange}{RGB}{255, 165, 0}  % Replace with actual RGB values
\definecolor{customBlue}{RGB}{135, 206, 250}  % Replace with actual RGB values

\tikzset{
    base/.style = {minimum height=1cm, text centered, rectangle, rounded corners, inner sep=0pt, outer sep=0pt},
    base_layer1/.style = {base, font=\sffamily, draw=black, text width=2.1cm},
    base_layer2/.style = {base, font=\sffamily\footnotesize, draw=black, text width=9cm},
    root/.style = {base_layer1, draw=lightgray, fill=lightgray!30, minimum height=1.4cm},
    layer1_1/.style = {base_layer1, draw=customPurple, fill=customPurple!30},
    layer1_2/.style = {base_layer1, draw=customGreen, fill=customGreen!30},
    layer1_3/.style = {base_layer1, draw=customOrange, fill=customOrange!30},
    layer1_4/.style = {base_layer1, draw=customBlue, fill=customBlue!30},
    layer2_1/.style = {base_layer2, draw=customPurple!20, fill=customPurple!20},
    layer2_2/.style = {base_layer2, draw=customGreen!20, fill=customGreen!20},
    layer2_3/.style = {base_layer2, draw=customOrange!20, fill=customOrange!20},
    layer2_4/.style = {base_layer2, draw=customBlue!20, fill=customBlue!20},
    evl_base_layer1/.style = {base, font=\sffamily, text width=2.0cm, minimum height=0.8cm},
    evl_base_layer2/.style = {base, font=\sffamily, text width=2.0cm, minimum height=0.6cm},
    evl_base_layer3/.style = {base, font=\sffamily\small, text width=2.3cm, minimum height=0.6cm},
    evl_base_layer4/.style = {base, font=\sffamily\small, text width=7.5cm, minimum height=1.2cm},
    evl_base_layer4_large/.style = {base, font=\sffamily\small, text width=10.1cm, minimum height=1.0cm},
    evl_layer1_1/.style = {evl_base_layer1, draw=customGreen, fill=customGreen!30},
    evl_layer1_2/.style = {evl_base_layer1, draw=customOrange, fill=customOrange!30},
    evl_layer2_2/.style = {evl_base_layer2, draw=customOrange, fill=customOrange!30},
    evl_layer3_2/.style = {evl_base_layer3, draw=customOrange, fill=customOrange!30},
    evl_layer4_2/.style = {evl_base_layer4, draw=customOrange!20, fill=customOrange!20},
    evl_layer4_2_large/.style = {evl_base_layer4_large, draw=customOrange!20, fill=customOrange!20},
    odd_base_layer1/.style = {base, font=\sffamily, text width=2.1cm},
    odd_base_layer2/.style = {base, font=\sffamily\small, text width=3.1cm, minimum height=0.5cm},
    odd_base_layer3/.style = {base, font=\sffamily\footnotesize, text width=5.8cm, minimum height=0.8cm},
    odd_layer1_1/.style = {odd_base_layer1, draw=customGreen, fill=customGreen!30},
    odd_layer1_2/.style = {odd_base_layer1, draw=customOrange, fill=customOrange!30},
    odd_layer1_3/.style = {odd_base_layer1, draw=customBlue, fill=customBlue!30},
    odd_layer2_1/.style = {odd_base_layer2, draw=customGreen, fill=customGreen!30},
    odd_layer2_2/.style = {odd_base_layer2, draw=customOrange, fill=customOrange!30},
    odd_layer2_3/.style = {odd_base_layer2, draw=customBlue, fill=customBlue!30},
    odd_layer3_1/.style = {odd_base_layer3, draw=customGreen!20, fill=customGreen!20},
    odd_layer3_2/.style = {odd_base_layer3, draw=customOrange!20, fill=customOrange!20},
    odd_layer3_3/.style = {odd_base_layer3, draw=customBlue!20, fill=customBlue!20},
    cis_base_layer1/.style = {base, font=\sffamily\footnotesize, text width=2.1cm},
    cis_base_layer2/.style = {base, font=\sffamily\footnotesize, text width=2.8cm, minimum height=0.7cm},
    cis_base_layer3/.style = {base, font=\sffamily\footnotesize, text width=2.9cm, minimum height=0.5cm},
    cis_base_layer4/.style = {base, font=\sffamily\footnotesize, text width=6.7cm, minimum height=0.8cm},
    cis_base_layer4_sm/.style = {base, font=\sffamily\footnotesize, text width=4cm, minimum height=0.8cm},
    cis_layer1_1/.style = {cis_base_layer1, draw=customGreen, fill=customGreen!30},
    cis_layer1_2/.style = {cis_base_layer1, draw=customOrange, fill=customOrange!30},
    cis_layer1_3/.style = {cis_base_layer1, draw=customBlue, fill=customBlue!30},
    cis_layer2_1/.style = {cis_base_layer2, draw=customGreen, fill=customGreen!30},
    cis_layer2_2/.style = {cis_base_layer2, draw=customOrange, fill=customOrange!30},
    cis_layer2_3/.style = {cis_base_layer2, draw=customBlue, fill=customBlue!30},
    cis_layer3_2/.style = {cis_base_layer3, draw=customOrange, fill=customOrange!30},
    cis_layer4_1/.style = {cis_base_layer4, draw=customGreen!20, fill=customGreen!30},
    cis_layer4_2/.style = {cis_base_layer4, draw=customOrange!20, fill=customOrange!30},
    cis_layer4_3/.style = {cis_base_layer4, draw=customBlue!20, fill=customBlue!30},
    arrow/.style={-Latex},
}

\usepackage{amsmath}
\usepackage{hyperref}
\usepackage{multirow}
\usepackage{accents}
\usepackage{wrapfig}
\usepackage{threeparttable}
\AtBeginDocument{%
  \providecommand\BibTeX{{%
    \normalfont B\kern-0.5em{\scshape i\kern-0.25em b}\kern-0.8em\TeX}}}

\newcommand{\miniskip}{\vspace*{-.5\baselineskip}}
\newcommand{\shrink}{\vspace*{-.9\baselineskip}}

\newcommand{\note}[1]{\textcolor{black}{#1}}
\newcommand{\shorten}[1]{\textcolor{black}{#1}}
\newcommand{\new}[1]{\textcolor{black}{#1}}

\acmJournal{CSUR}
\acmYear{2026} \acmVolume{1} \acmNumber{1} \acmArticle{}
\acmMonth{1} \acmDOI{10.1145/3795686}

\begin{document}
\title{A Survey on Recent Advances in Conversational Data Generation}

\author{Heydar Soudani}
\email{heydar.soudani@ru.nl}
\orcid{0000-0003-0393-8662}
\affiliation{%
  \institution{Radboud University}
  \city{Nijmegen}
  \country{The Netherlands}
}
\author{Roxana Petcu}
\email{r.m.petcu@uva.nl}
\orcid{0000-0002-2617-205X}
\affiliation{%
  \institution{University of Amsterdam}
  \city{Amsterdam}
  \country{The Netherlands}
}
\author{Evangelos Kanoulas}
\email{e.kanoulas@uva.nl}
\orcid{0000-0002-8312-0694}
\affiliation{%
  \institution{University of Amsterdam}
  \city{Amsterdam}
  \country{The Netherlands}
}
\author{Faegheh Hasibi}
\orcid{0009-0006-9986-482X}
\email{f.hasibi@cs.ru.nl}
\affiliation{%
  \institution{Radboud University}
  \city{Nijmegen}
  \country{The Netherlands}
}

\begin{abstract}

% V1: 
% In recent years, conversational systems have gained significant importance, particularly with the emergence of advanced chatbots such as ChatGPT and GPT-4. These systems have revolutionized the way humans interact with machines, offering new possibilities in various domains. However, a major obstacle in training these sophisticated chatbots is the scarcity of specialized dialogue data. Traditionally, the creation of conversational datasets has relied on crowdsourcing methods. Despite its utility, this method presents several drawbacks, including high costs, limited scalability, and the need for extensive human effort, rendering it far from an optimal solution. Recently, the generation of synthetic dialogue data has emerged as a promising alternative. This approach includes efforts to augment existing dialogue datasets or to transform other textual resources into conversational data, offering a more feasible and efficient method for dataset creation. 

% V2:
Recent advancements in conversational systems have significantly enhanced human-machine interactions across various domains. However, training these systems is challenging due to the scarcity of specialized dialogue data. Traditionally, conversational datasets were created through crowdsourcing, but this method has proven costly, limited in scale, and labor-intensive. As a solution, the development of synthetic dialogue data has emerged, utilizing techniques to augment existing datasets or convert textual resources into conversational formats, providing a more efficient and scalable approach to dataset creation.
%
% V1:
% In this survey, we provide a systematic and comprehensive review of research on the creation of multi-turn conversational data. We discuss existing works for three types of dialogue systems: open domain, task-oriented, and information-seeking conversations.
% % We also classify existing works based on distinct features of each system, aiming to simplify the understanding of these complex systems. 
% We group existing works based on distinct features, such as the data that conversation grounded on, the turn generation approaches and the filtering methods, and we propose a general framework aiming to introduce main principles of a conversation data generation system.
% Moreover, we discuss evaluation metrics and methods used for assessing the synthetic conversational data. Moreover, we delve into the current challenges facing this research area and explore potential directions for future work. 
% V2
In this survey, we offer a systematic and comprehensive review of multi-turn conversational data generation, focusing on three types of dialogue systems: open domain, task-oriented, and information-seeking. We categorize the existing research based on key components like seed data creation, utterance generation, and quality filtering methods, and  introduce a general framework that outlines the main principles of conversation data generation systems. Additionally, we examine the evaluation metrics and methods for assessing synthetic conversational data, address current challenges in the field, and explore potential directions for future research.
Our goal is to accelerate progress for researchers and practitioners by presenting an overview of state-of-the-art methods and highlighting opportunities to further research in this area.

\end{abstract}




\maketitle

\section{Introduction}\label{sec:introduction}

Conversational AI focuses on enabling machines to interact with humans using natural language~\cite{Zamani2022CIS}. These systems, particularly enhanced by neural networks and deep learning, aim to closely mimic human conversational behaviors~\cite{deng2023goalAwareness}. This enhancement is manifested in the system's ability to handle complicated challenges, such as discussing specialized topics~\cite{ruggeri2023ArgSciChat, li2023newsdialogues, qi2023pragmaticqa}, shifting between conversation topics~\cite{adlakha2022topiocqa}, asking proactive questions~\cite{Wu22INSCIT}, and integrating emerging topics~\cite{Soudani23Data}. To develop systems capable of addressing these sophisticated challenges, a robust and large-scale source of training data is essential.

Crowdsourcing is the primary method for creating conversational datasets, where human workers generate data based on provided instructions~\cite{Budzianowski18MultiWOZ, li2017dailydialog, zhang2018personachat, choi2018quac, Reddy2019CoQA, Feng2020Doc2Dial, fengl2021multidoc2dial, qu2018msdial, Yang18MSDialog2, Penha19MANtIS, liu2021durecdial, Soudani25Enhancing}. This approach, however, is costly, time-consuming, and challenging to scale or adapt to new domains~\cite{rashkin2019towardsEmpathetic, He18CraigslistBargains, sevegnani2021otters}. It also introduces annotation biases~\cite{Yang2020DAug} and is ineffective for emerging topics, especially in specialized fields or languages with limited development resources~\cite{lee2022personachatgen, Soudani24Fine}. As an alternative, generating synthetic conversational datasets has emerged as a promising solution. This method involves transforming existing textual resources—like documents, tables, and knowledge graphs—into conversational formats or augmenting existing dialogue data with new instances~\cite{Soudani23Data}. Synthetic conversation generation is especially beneficial in resource-scarce domains, enabling models to leverage all available resources to enhance learning~\cite{Ovadia23Injection, gupta2024rag}. Data Generation denotes the process of expanding or enhancing a dataset by creating entirely new data points, as compared to data augmentation, which involves applying transformations or modifications to existing data points. Generation results in more robust and diverse training samples as compared to augmentation, where the new data points are not present in the original dataset, however, are plausible and relevant. 

\begin{figure}[t]
\centering % Centers the figure
\miniskip
\begin{tikzpicture}[node distance=0.5cm]

% Main Nodes
\node (convgen) [root] {Conversation Generation};
\node (tod) [layer1_2, above right=-0.65cm and 0.5cm of convgen] {Task-oriented};
\node (odd) [layer1_3, below right=-0.65cm and 0.5cm of convgen] {Open Domain};
\node (eval) [layer1_1, above=0.2 of tod] {Evaluation};
\node (cis) [layer1_4, below=0.2 of odd] {Information Seeking};

\node (eval_papers) [layer2_1, right=of eval] {BERTScore~\cite{Zhang20BERTScore}, BARTScore~\cite{Yuan21BARTScore}, Coverage~\cite{wu2022dg2, kim2021neuralwoz}, Exact Match~\cite{wu2022dg2, kim2022simseek}, Dist-n~\cite{li2016distn}, Ent-n~\cite{Zhang18entn}, SentBERT~\cite{Reimers19SentBERT}, USR~\cite{Mehri20USR}, Self-BLEU~\cite{Zhu18selfbleu}, GEval~\cite{Liu23GEval}, UniEval~\cite{zhong2022UniEval}, Human Evaluation~\cite{Smith22Human}};

\node (tod_papers) [layer2_2, right=of tod] {ABUS~\cite{li2017ABUS}, VHDA~\cite{yoo2020VHDA}, NUS~\cite{kreyssig2018neural}, M2M~\cite{Shah2018M2M}, NeuralWOZ~\cite{kim2021neuralwoz}, SGD~\cite{Rastogi2020SGD}, HUS~\cite{gur2018HUS}, VHUS~\cite{gur2018HUS}, TUS~\cite{lin2021TUS}, JOUST~\cite{tseng2021DSUSRL}, Unified-US~\cite{wan2022unified}, Simulated-Chat~\cite{mohapatra2021simulatedchats}, ICL-US~\cite{Terragni2023ICLUS}, Dialogic~\cite{li2022DIALOGIC}, INA~\cite{ahmad2023ina}};

\node (odd_papers) [layer2_3, right=of odd] {AUGESC~\cite{Zheng23AugESC}, PLACES~\cite{chen2023places}, SPC~\cite{Jandaghi23spc}, SODA~\cite{kim2022soda}, PERSONACHATGEN~\cite{lee2022personachatgen}, BOTSTALK~\cite{kim2022botstalk}, GCN~\cite{Lin22GCN}, WEAKDAP~\cite{chen2022weakly}};

\node (cis_papers) [layer2_4, right=of cis] {MultiCQAG~\cite{Hwang22MultiCQAG}, CQAG-AR~\cite{HwangL22CQAGAR}, DG2~\cite{wu2022dg2}, SimSeek~\cite{kim2022simseek}, Inpainting~\cite{Dai22Inpainting}, SynDG~\cite{bao2023SynDG}, LAPS~\cite{Joko2024LAPS}, TtWMusic~\cite{leszczynski2023talk}, MusicSyn~\cite{leszczynski2022conversational}, SimQuAC~\cite{abbasiantaeb2024let}, SOLID~\cite{askari2024self}, TopDial~\cite{Wang23topdial}};
%MATHDIAL~\cite{macina2023mathdial}};

% Arrows
\draw (convgen.east) -- ++(5pt,0) |- (eval.west);
\draw (convgen.east) -- ++(5pt,0) |- (tod.west);
\draw (convgen.east) -- ++(5pt,0) |- (odd.west);
\draw (convgen.east) -- ++(5pt,0) |- (cis.west);
\draw (eval) -- (eval_papers);
\draw (tod) -- (tod_papers);
\draw (odd) -- (odd_papers);
\draw (cis) -- (cis_papers);

\end{tikzpicture}
\caption{An overview of multi-turn conversation generation sections and papers.}
\label{fig:conversationgen_secs}
% \shrink
\end{figure}

% === History of ConvAI and Dialogue system types
The history of conversational AI traces back to the 1960s with the creation of the first chatbot, Eliza~\cite{weizenbaum1966eliza}. Eliza operated on rule-based principles, generating responses based on predefined rules tied to specific keywords or facets.
Following this, the rise of systems that integrate rule-based and machine learning components has been observed~\cite{Ni2023Recent, deng2023goalAwareness}, exemplified by platforms such as Apple's Siri, Amazon's Alexa, and Microsoft's Xiaobing. These systems have found commercial and industrial applications, relying on machine learning algorithms to improve response accuracy by learning from historical user interactions and conversation data~\cite{Jinjie2023Recent}.
Further advancements occurred with the advent of deep learning-based systems, leveraging large models and extensive datasets to enhance their capabilities~\cite{Ni2023Recent}. Recently, the emergence of Large Language Models (LLMs) has taken this a step further. These models excel in generating contextually relevant, informative, and diverse responses that mirror human language, thanks to their ability to analyze vast amounts of data and the sophistication of their underlying architectures~\cite{deng2023goalAwareness, yi2024survey}.

Dialog systems can be categorized into three paradigms: Task-oriented Dialogs (TOD), Open Domain Dialogs (ODD), and Conversational Information Seeking (CIS)~\cite{Yang:2025:survey, Soudani23Data, Zamani2022CIS}; 
\note{ see the electronic appendix for examples.}
% see Table~\ref{tab:dialog_types_examples}. 
\note{Traditionally, these systems have been distinguished based on methodological differences. However, with the advancements of LLMs, the boundaries between them are increasingly blurring.}
The first paradigm, TOD, seeks to execute tasks at the user's request by accurately identifying their intentions and providing appropriate responses~\cite{wang2017TODModels}. TOD systems are built to comprehend user queries and generate responses that help achieve the user's objective. They find applications in various domains and use cases, such as flight booking, restaurant reservations, hotel accommodations, taxi services, movie ticket purchases, weather inquiries, navigation assistance, scheduling, and customer service~\cite{wen2017Applications, fellows2021Applications}. TOD faces multiple challenges, such as ensuring robust and reliable dialogue state tracking~\cite{fellows2021Applications}, integrating external databases for extracting specific conversational entities~\cite{yang2020KnowledgeBaseChallenges}, e.g., restaurant names, flight numbers, and cinema schedules, and optimizing response generation for user satisfaction and task completion~\cite{wen2017woz}. With a focus on these challenges, we propose a taxonomy for TOD generation based on four steps: identifying input sources as external knowledge, such as knowledge bases, schemas, and ontologies, or internal, such as knowledge extracted from training data, dialogue generation methods, training, and quality filtering. This multi-faceted analysis offers an intuition of TOD generation, emphasizing the critical roles of input handling, generation and training approaches, and quality control in their development and deployment.

The second paradigm, ODD, aims to facilitate casual conversations with users across a broad range of topics and domains, without being confined to specific tasks or objectives~\cite{Ni2023Recent, kim2022soda}.
ODD systems encounter several challenges, such as maintaining coherent and contextually appropriate responses~\cite{Liu21ESConv}, ensuring response diversity to engage users~\cite{zhao2019LaRL}, displaying proactivity by steering discussions toward certain topics~\cite{chen2023controllable, yang2022topkg, Liu23MTGP}, personalizing responses based on user profiles~\cite{zhang2018personachat}, and generating informative replies from external knowledge bases~\cite{Lin22GCN}. 
Addressing these challenges requires systems to be trained with datasets that specifically include them, making synthetic data generation a promising approach.
This survey proposes a taxonomy for synthetic dataset creation, structured in three steps: seed generation, turn generation, and quality filtering. This taxonomy is introduced to abstract choices in existing methods in the literature to produce diverse, informative~\cite{kim2022soda, chen2023places}, multi-skill~\cite{kim2022botstalk}, and personalized conversational data~\cite{lee2022personachatgen, Jandaghi23spc}.

\begin{figure}[t]
    \centering
    \includegraphics[width=0.92\textwidth]{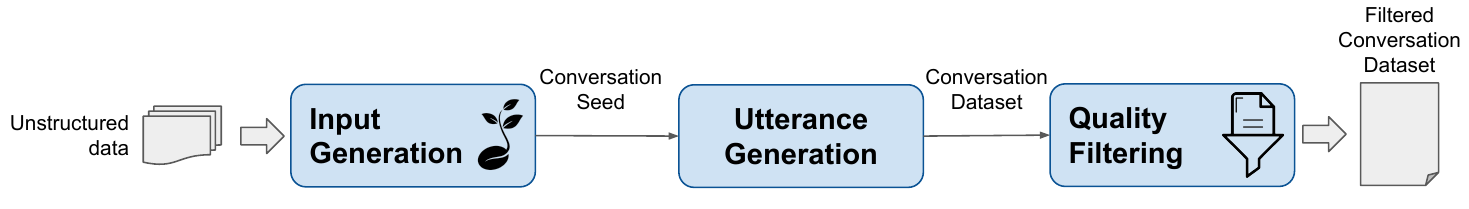}
    % \shrink
    \caption{The general framework for conversation generation in TOD, ODD, and CIS systems. \note{This framework consists of three main components: \textbf{1) Input Generation} creates a dialog seed , \textbf{2) Utterance Generation} uses the dialog seed as an input to a model and generates a multi-turn conversation sample, and \textbf{3) Quality Filtering} eliminates dialogs that do not achieve a desirable representation; filtering can be \textit{lexical} or \textit{consistency} related.}}
    \label{fig:main_pipeline}
    \shrink
\end{figure}

% Definition
The third paradigm, CIS, is designed to assist users in seeking and retrieving information through natural language dialogue interactions. The primary objective of a CIS system is to satisfy the information needs of users by engaging in dynamic conversations, which may include text as well as other modalities such as voice, clicks, or touch~\cite{Zamani2022CIS}. CIS encompasses three main areas: conversational search, conversational question answering (QA), and conversational recommendation.
\textit{Conversational Search} and \textit{Conversational QA} involve interacting with a system in natural language to find specific information, allowing users to pose multiple questions based on their interaction history, with the system providing relevant answers~\cite{wu2022dg2, abbasiantaeb2024let}. \textit{Conversational Recommendation Systems} suggest items to users based on their previous interactions, serving as personalized information-seeking tools~\cite{liu2021durecdial, Wang23topdial}.
% Challenges:
CIS systems face several challenges, including maintaining the conversation state to better respond to user needs, managing mixed-initiative interactions where the system sometimes leads the conversation and sometimes responds to user queries, and adapting to user preferences and profiles to enhance satisfaction by personalizing the conversation.
Based on these challenges, tasks such as intent prediction, asking clarification questions, target-oriented recommendation, and topic switching are defined. These tasks help clarify uncertainties, lead conversations towards specific items, and adapt the discussion to cover various subjects and entities based on the user's informational needs~\cite{adlakha2022topiocqa}. Effective handling of these tasks requires a substantial amount of well-curated data for training the models.
% contribution
In this work, we systematically review the literature on information-seeking conversational datasets and provide a three-step framework similar to those used for ToD and ODD. 
\note{The three-step framework used in each subsection is illustrated in Figure \ref{fig:main_pipeline}.}

\note{In this work, we also differentiate between LLM-based and non LLM-based approaches.
Specifically, we define \textit{Language Models (LMs)} as encoder-only, decoder-only, or encoder–decoder architectures,  typically with fewer than one billion parameters. %,trained or fine-tuned with limited, domain-specific data.
In contrast, we define \textit{Large Language Models (LLMs)} as foundation models trained on broad, general-purpose corpora with more than one billion parameters, and capable of generalization through prompting, few-shot learning, or in-context learning (ICL). Based on these definitions, we categorize DA approaches into the following groups and provide detailed coverage of these categories in the corresponding subsections.}

% \begin{itemize}
\begin{itemize}%[leftmargin=*]
    \item \note{LLM-based approaches: Methods that rely on prompting or in-context learning of LLMs (such as GPT-3, PaLM, or LLaMA) to generate full dialogue responses without supervised fine-tuning. }
    \item \note{Non LLM-based approaches: Methods that use LMs, such as BART and RoBERTa, or do not use LMs.  
    % on LLM-based approaches: Approaches using LMs, such as BART and RoBERTa. 
    This category also includes methods that use LLMs only for language generation but not for generating the core structure of the dialog; e.g., methods that create TOD intents from a knowledge base but generate language around them using LLMs.}
    %Thhis category, we also consider approaches that use LLMs for language generation but not for generating the core structure of the dialog, such as sampling TOD intents from a knowledge base but generating language around them using LLMs.}    
\end{itemize}

% Our contributions
% In our survey, we systematically examine methods developed for generating new conversational data. 
% We list all the synthetic datasets discussed in Table~\ref{table:synthetic_datasets}.
\new{Figure~\ref{fig:conversationgen_secs} presents an overview of the paper’s structure.}
The contributions of this work are summarized as follows:
% \begin{itemize}
\begin{itemize}%[leftmargin=*]
    \item We present a comprehensive review of the methodologies employed for creating synthetic conversational data, highlighting the innovative techniques and processes that facilitate their generation.
    \item We introduce a general framework consisting of three steps: input generation, utterance generation, and quality filtering, which are used to abstract and describe dialogue data generation across three distinct types of dialogue systems: open-domain, task-oriented, and information-seeking.
    \item We offer a detailed description of data generation evaluation methods, detailing the metrics and criteria used to assess the quality, relevance, and utility of conversational datasets.
    \item We discuss prospective areas of focus and recent research challenges that have emerged due to the growing demands for data generation.
\end{itemize}

\section{Evaluation}~\label{sec:eval}

% === List of papers
% [Dist-n] \cite{li2016distn}
% [Ent-n] \cite{Zhang18entn}
% [Self-BLEU] \cite{Zhu18selfbleu}
% [GEval] \cite{Liu23GEval}
% [UniEval] \cite{zhong2022UniEval}
% [BERTScore] \cite{Zhang20BERTScore}
% [Human] \cite{Smith22Human}
% [Simulator] \cite{}
% [USR] \cite{Mehri20USR}

% === Introduction ====
Before exploring data generation methods, it's essential to understand how we can evaluate the outcomes dialogue generation methods. While this paper does not focus on conversation evaluation, we find it necessary to provide a comprehensive overview of how generated conversations are evaluated. There are primarily two approaches for assessing synthetically generated conversational data: \textbf{\textit{extrinsic}} and \textbf{\textit{intrinsic}} evaluation, described below.

% In the extrinsic approach, the quality of the generated data is evaluated based on its performance in downstream tasks. In the intrinsic approach, the synthetic data is directly evaluated using human or automatic metrics to assess various qualities of the generated conversations, such as naturalness, understandability, and coherence. To demonstrate generality, existing works often employ a combination of both approaches. In this section, we review existing methods for evaluating the results of conversational data generation.

\medskip
\noindent
\textbf{\emph{Extrinsic Evaluation}} measures the quality of the generated data based on the performance of downstream tasks they are used for.
% Definition
Given that the ultimate purpose of synthetic conversational data generation is to augment training data and improve the performance of conversational agents, a dialogue model is trained using all or part of the synthetic data and the final performance of the model is used to indirectly assess the quality of the data generation model~\cite{kim2022simseek, xu2023baize}.
% conversation generation models are indirectly assessed based on the effectiveness of these downstream tasks, where the dialogue model is fine-tuned using all or part of the synthetic data.% Consequently, in the extrinsic approach, the quality of synthetic data is 
% In this scenario, the performance of the downstream task directly reflects the quality of the synthetic data.
% Examples
For example, AUGESC~\cite{Zheng23AugESC} and SODA~\cite{kim2022soda} aim to create an ODD conversational dataset for fine-tuning an emotional support conversational system and a social conversational agent, respectively, using  synthetically generated data. Similarly, within the CIS system, SOLID~\cite{askari2024self} focuses on enhancing the intent prediction task with a portion of the generated data and Inpainting~\cite{Dai22Inpainting} highlights that its generated dataset serves as a valuable training resource for ConvQA systems.
% 
% As another example within CIS, DG2~\cite{wu2022dg2} generates its dataset by mirroring the process used to create a crowdsourced dataset known as Doc2Dial~\cite{Feng2020Doc2Dial}, and it adopts the same downstream tasks introduced in Doc2Dial. The first task is span selection, where an agent must identify relevant text spans based on the dialogue context. This task requires the agent to interpret user queries within the context of the dialogue history and select the appropriate span from the related document. The second task, agent response prediction, involves producing coherent responses based on extracted rationales. This task demands that the agent generate a natural language response to the user query, taking into account both the dialogue context and the document.

\begin{figure}[t]
\centering % Centers the figure
\begin{tikzpicture}[node distance=0.5cm]

% Main Nodes
\node (convevl) [root] {Conversation Evaluation};
% \node (1_ext) [evl_layer1_1, above right=-0.2cm and 0.3cm of convevl] {Extrinsic};
% \node (2_int) [evl_layer1_2, below right=-0.2cm and 0.3cm of convevl] {Intrinsic};

\node (2_1_aut) [evl_layer2_2, above right=0.03cm and 0.3cm of convevl] {Automatic};
\node (2_2_hmn) [evl_layer2_2, below right=0.03cm and 0.3cm of convevl] {Human};

\node (2_1_1_bse) [evl_layer3_2, above right=0.1cm and 0.3cm of 2_1_aut] {Reference-based};
\node (2_1_2_fre) [evl_layer3_2, below right=0.1cm and 0.3cm of 2_1_aut] {Reference-free};

\node (2_1_1_1_ppr) [evl_layer4_2, right=0.2cm of 2_1_1_bse] {Word overlap, BERTScore~\cite{Zhang20BERTScore}, BARTScore~\cite{Yuan21BARTScore}, Exact Match~\cite{wu2022dg2, kim2022simseek}, Coverage~\cite{wu2022dg2, kim2021neuralwoz}, Coreference Alignment~\cite{gao2019CFnet}}; 
\node (2_1_2_1_ppr) [evl_layer4_2, right=0.2cm of 2_1_2_fre] {Dist-n~\cite{li2016distn}, Ent-n~\cite{Zhang18entn}, SentBERT~\cite{Reimers19SentBERT}, Self-BLEU~\cite{Zhu18selfbleu}, USR~\cite{Mehri20USR}, UniEval~\cite{zhong2022UniEval}, GEval~\cite{Liu23GEval}, Simulation~\cite{tang2019target, yang2022topkg}};
\node (2_2_1_1_ppr) [evl_layer4_2_large, right=0.2cm of 2_2_hmn] {Single-model and pair-wise evaluation, per-turn and per-dialogue evaluation \cite{Smith22Human}};

\draw (convevl.east) -- ++(5pt,0) |- (2_1_aut.west);
\draw (convevl.east) -- ++(5pt,0) |- (2_2_hmn.west);
% \draw (2_int.east) -- ++(5pt,0) |- (2_1_aut.west);
% \draw (2_int.east) -- ++(5pt,0) |- (2_2_hmn.west);
\draw (2_1_aut.east) -- ++(5pt,0) |- (2_1_1_bse.west);
\draw (2_1_aut.east) -- ++(5pt,0) |- (2_1_2_fre.west);
\draw (2_1_1_bse) -- (2_1_1_1_ppr);
\draw (2_1_2_fre) -- (2_1_2_1_ppr);
\draw (2_2_hmn) -- (2_2_1_1_ppr);

\end{tikzpicture}
\caption{An overview of methods for evaluating generated multi-turn conversations. 
% \todo{Please remove extrinsic and intrinsic boxes, directly branch out to Automatic and Human.}
}
\label{fig:eval_all}
% \shrink
\end{figure}

\medskip
\noindent
\textbf{\emph{Intrinsic Evaluation}} measures synthetic data quality directly  using human or automatic evaluation metrics, assessing various qualities of the generated conversations, such as naturalness, understandability, and coherence. 
the quality of the generated dialogue is directly evaluated either automatically using various evaluation metrics and through human evaluation. 
Automatic evaluation metrics of conversations  are designed to either compare the model's output against a reference dataset (ground truth) or assess it based on inherent characteristics without reference.
To demonstrate generality, existing studies often employ a combination of both intrinsic and extrinsic evaluation methodologies. 
In this section, we review existing methods for evaluating conversations, generated either during the data augmentation process or by a dialogue model. An overview of these methods is presented in Figure~\ref{fig:eval_all} \new{and Table~\ref{tab:eval_description}}.

% Description table ...
% Table 3: different action setup
\renewcommand{\arraystretch}{0.85}
\begin{table*}[t]
\centering
\caption{\note{Overview of evaluation metrics for conversational systems, categorized to Reference-based and Reference-free metrics.}}
\label{tab:eval_description}
\begin{tabular}{l |l |m{10.5cm}}
\hline
Metric & Category &Description \\ \hline\hline

Exact Match~\cite{wu2022dg2} & Ref-based & A strict accuracy metric that measures the percentage of predictions that match the reference text exactly, with no partial credit. \\
BLEU & Ref-based & A precision-based metric that measures the overlap of n-grams between a generated text and reference texts, emphasizing exact word matches. \\
ROUGE-L &  Ref-based& A recall-oriented metric that evaluates the longest common subsequence between a generated text and reference texts, capturing sentence-level structure similarity. \\
METEOR & Ref-based & A metric that scores text similarity based on unigram matches, stemming, synonyms, and word order, balancing precision and recall. \\% with higher correlation to human judgment. \\
BERTScore~\cite{Zhang20BERTScore} & Ref-based & A semantic similarity metric that computes token-level cosine similarity between contextual embeddings from BERT for generated and reference texts. \\
BARTScore~\cite{Yuan21BARTScore} & Ref-based & An evaluation metric that uses the log-likelihood from a pretrained BART model to assess the quality of generated text given the reference or source text. \\
Dist-n~\cite{li2016distn} & Ref-free & A diversity metric that calculates the ratio of distinct n-grams to the total number of generated n-grams, measuring lexical diversity. \\
Ent-n~\cite{Zhang18entn} & Ref-free & A diversity metric that computes the entropy of the n-gram distribution in generated text, capturing the unpredictability and variety of word usage. \\
SentBERT~\cite{Reimers19SentBERT} & Ref-free & A semantic similarity metric that uses sentence embeddings from BERT to measure the cosine similarity between generated and reference texts. \\
Self-BLEU~\cite{Zhu18selfbleu} & Ref-free & A diversity metric that computes BLEU scores between each generated sentence and the rest of the generated sentences, where lower scores indicate higher diversity. \\
USR~\cite{Mehri20USR} & Ref-free & An unsupervised and reference-free metric that evaluates dialogue quality based on user satisfaction through multiple model-based evaluators (e.g., relevance, fluency, and engagement). \\
UniEval~\cite{zhong2022UniEval} & Ref-free & A unified evaluation framework that uses a single model to assess multiple aspects of text generation quality (such as relevance, coherence, and fluency) in a reference-free or reference-based manner. \\
G-EVAL~\cite{Liu23GEval} & Ref-free & An LLM-based evaluation framework that leverages GPT models with carefully designed prompts to assess multiple dimensions of text quality, such as coherence, consistency, and relevance. \\
\hline
\end{tabular}
% \shrink
\end{table*}

\subsection{Automatic Evaluation}
Conversations can be evaluated automatically using reference-based and reference-free approaches.

\subsubsection{Reference-based} These methods assess the quality and relevance of generated text by comparing it against a set of predefined reference texts. These metrics are designed to measure how closely the generated dialogue aligns with expected responses, providing insights into the accuracy and coherence of the conversation. Dialogue data evaluation using these methods can be categorized into three main groups based on the metrics employed.

The first group, \textit{Word Overlap Metrics,} assesses the n-gram similarity between the generated text and the ground truth, focusing on the overlap of word sequences to evaluate the quality of generated questions against a predefined standard. Some examples of these metrics include BLEU (1-3), ROUGE-L (R-L), and METEOR.

The second group, \textit{Embedding-based Metrics}, relies on a score derived from comparing the embeddings of the generated text and the ground truth. Two examples from this group are BERTScore~\cite{Zhang20BERTScore} and BARTScore~\cite{Yuan21BARTScore}.
% 
% BERTScore utilizes the BERT model to generate embeddings for words in both the reference and generated texts. It calculates the cosine similarity between the contextual embeddings of words in the reference text and the generated text~\cite{do2022cohsCQG}. This allows BERTScore to measure the semantic similarity between the texts rather than relying solely on lexical matches, making it suitable for tasks where the meaning of the words is more important than their exact order or choice.
% 
\shorten{BERTScore uses BERT-based contextual embeddings to compute cosine similarity between words in reference and generated texts~\cite{do2022cohsCQG}. By measuring semantic similarity rather than exact lexical overlap, it better captures meaning beyond word choice or order.}
% 
% BARTScore~\cite{Yuan21BARTScore} is based on the BART model~\cite{lewis-2020-bart}, which is trained for both denoising and sequence-to-sequence tasks, giving it a unique approach to understand and generate text. BARTScore computes scores using a generative approach. It treats the evaluation of text as a text generation problem, where the score is derived from the likelihood of regenerating the reference given the candidate text as input, or vice versa. This can involve generating the likelihood of the reference text conditioned on the candidate text or scoring the candidate text directly based on its fluency and coherency.
% 
\shorten{BARTScore leverages the BART sequence-to-sequence model~\cite{lewis-2020-bart} to evaluate text using a generative scoring approach. It treats evaluation as a generation task, computing scores from the likelihood of generating the reference from the candidate (or vice versa).} %, capturing fluency and coherence.}
Compared to BERTScore, which focuses on semantic similarity using embeddings from BERT, BARTScore leverages the generative capabilities of BART to assess text quality from a more holistic perspective, incorporating fluency, coherence, and even factual accuracy~\cite{Zhang20BERTScore, Yuan21BARTScore}.

% One example of this group is BERTScore~\cite{Zhang20BERTScore}, which calculates similarity scores between generated and ground-truth texts by utilizing deep contextualized embeddings~\cite{do2022cohsCQG}. The aim is to capture the semantic similarity beyond mere word overlap, providing a more nuanced evaluation of text generation quality.
% BARTScore~\cite{Yuan21BARTScore} also is based on this idea that models trained to convert the generated text to/from a reference output or the source text will achieve higher scores when the generated text is better.

%
% The last group, \textit{Subtask Evaluation Metrics} includes specific metrics that assess how effectively dialogue models handle distinct components or subtasks of the conversation generation process. The first example is Exact Match~\cite{wu2022dg2}, defined as a condition where the predicted span exactly aligns with the actual span in the document. This metric is primarily used in generation methods that use a document as the reference for conversation data. It is important to note that since answers in a conversation can be lengthy, the exact match may be too stringent criterion. Consequently, the F1 word overlap metric is often used as an alternative~\cite{kim2022simseek}.
% 
\shorten{The last group, \textit{Subtask Evaluation Metrics}, assess how well dialogue models handle specific components of conversation generation. For example, Exact Match \cite{wu2022dg2} requires the predicted span to exactly match the reference span, and is mainly used in document-grounded generation. Since this criterion is often too strict for long conversational answers, F1 word overlap is commonly used instead \cite{kim2022simseek}.}
Another metric is Coverage~\cite{wu2022dg2, kim2021neuralwoz}, which is used in both ToD and CIS conversation generation. In CIS, Span Coverage~\cite{wu2022dg2} assesses how effectively a model captures the rationale spans within a document. The premise here is that dialogues covering a larger portion of the document indicate a more effective data augmentation method. Span Coverage is calculated by measuring the proportion of the document that the generated dialogue spans cover.
% The formula for Span Coverage is as follows:
% \[
% \text{Coverage}=\frac{\sum_{\text{span}}\left|\bigcup_{d \in \text{doc}_i} \bigcup_{s \in d} s\right|}{|\text{document}_i|}
% \]
% In this formula, \(s\) represents spans within a document, and \(doc\) refers to the number of documents in the corpus.
% 
In the context of ToD, zero-shot coverage~\cite{kim2021neuralwoz} is evaluated, which measures the accuracy ratio between zero-shot learning outcomes in a target domain and a fully trained model that includes that domain.
Another metric in this group is Coreference Alignment~\cite{gao2019CFnet}, which is critical in conversational QA where using coreferences, such as pronouns 'he' and 'she', is common—almost half of the questions include explicit coreference markers~\cite{Reddy2019CoQA}. Specifically, coreference alignment modeling instructs the decoder to focus on the correct non-pronominal coreferent mention in the conversation's attention distribution to accurately generate the pronominal reference word~\cite{gao2019CFnet}. To evaluate this feature, the Precision (P), Recall (R), and F-score (F) of pronouns in the generated questions are calculated and compared with those in the ground truth questions.

\subsubsection{Reference-free} These methods assess the generated dialogue without comparing them to ground truth data, evaluating inherent properties such as diversity~\cite{li2016distn, Zhang18entn}, semantic coherence~\cite{Reimers19SentBERT}, and user satisfaction~\cite{Mehri20USR}. These methods provide a flexible evaluation of the dialogue's originality and engagement, allowing for a broader interpretation of quality without the constraint of matching to a specific ground truth.

For diversity-based metrics, Dist-n~\cite{li2016distn} calculates the diversity of generated text by determining the ratio of unique unigrams and bigrams to the total number of words generated~\cite{pan2019ReDR}. Ent-n~\cite{Zhang18entn} measures the uniformity of the n-gram distribution across all generated text, revealing the variability in n-gram usage.
Another metric for evaluating diversity is Self-BLEU~\cite{Zhu18selfbleu}. While BLEU typically measures the similarity between two sentences, Self-BLEU uses one sentence from a set as a hypothesis and the rest as references, calculating a BLEU score for each sentence. The average of these scores is termed Self-BLEU. A higher Self-BLEU score indicates lower diversity among the sentences.
% 
% Sent-BERT~\cite{Reimers19SentBERT} assesses the semantic diversity of responses by computing the average negative cosine similarity between the SentenceBERT embeddings of each response pair, thus gauging how semantically varied the responses are. It should be noted that diversity metrics are often dependent on the length of the text, and tend to yield higher scores in texts with fewer total words due to the increased likelihood of repetitive words in longer texts compare to smaller ones. To address this disparity, LAPS~\cite{Joko2024LAPS} imposes a word cutoff for all evaluated datasets until the threshold of the word count is reached.
% 
\shorten{Sent-BERT \cite{Reimers19SentBERT} measures semantic diversity by computing the average negative cosine similarity between SentenceBERT embeddings of response pairs. We note that diversity metrics are length-sensitive and often yield higher scores for shorter texts due to reduced word repetition. To mitigate this effect, LAPS \cite{Joko2024LAPS} applies a word-count cutoff across datasets.}

% USR~\cite{Mehri20USR} is a reference-free measure designed to evaluate the quality of dialogue, incorporating five sub-metrics that together assess the overall quality of a conversation. These sub-metrics are: (1) Understandable, which evaluates the coherence of a response within its context; (2) Natural, which assesses whether a response mirrors natural human speech; (3) Maintains Context, which checks if a response appropriately continues the conversation; (4) Interesting, which determines whether a response is engaging or mundane; and (5) Uses Knowledge, which examines if a response appropriately includes relevant facts. The overall quality score is derived by aggregating these sub-metrics through a regression model trained on human evaluation.
% 
\shorten{USR~\cite{Mehri20USR} is a reference-free dialogue evaluation metric composed of five sub-metrics: Understandable (response coherence), Natural (human-likeness), Maintains Context (conversation continuity), Interesting (engagement), and Uses Knowledge (appropriate factual content). A regression model trained on human judgments combines these components into an overall dialogue quality score.}
% 
% 
% UniEval -> \cite{zhong2022UniEval}
% UniEval~\cite{zhong2022UniEval} is another aspect-based reference-free evaluator that provides a unified, multi-dimensional method for evaluating Natural Language Generation (NLG) tasks. Specifically, UniEval utilizes T5 as a backbone model and cast each evaluation aspect (e.g., coherence or consistency) to a Boolean QA problem; e.g., to assess the coherence of a summary it poses a boolean question, such as "Is this a coherent summary of the document?" Furthermore, it performs an intermediate training phase on four types of NLG, building a single evaluator for each task. For the dialogue generation task, UniEval follows USR and considers the four aspects of Naturalness, Coherence, Engagingness, and Groundedness.
% 
\shorten{UniEval \cite{zhong2022UniEval} is another reference-free, aspect-based NLG evaluator that uses T5 to unify multiple evaluation dimensions. It frames each aspect (e.g., coherence, consistency) as a Boolean QA task. For example, it asks whether a summary is coherent with its source document. UniEval is trained on four NLG task types, producing one evaluator per task. For dialogue generation, it follows USR and evaluates Naturalness, Coherence, Engagingness, and Groundedness.}
% 
% GEval -> \cite{Liu23GEval}
% Another metric, G-EVAL~\cite{Liu23GEval}, utilizes LLMs within a chain-of-thought (CoT) and form-filling framework to evaluate the quality of NLG outputs. By providing only the "Task Introduction" and "Evaluation Criteria" as prompts, G-EVAL prompts LLMs to generate a detailed CoT comprising Evaluation Steps. Subsequently, G-EVAL uses both the prompt and the generated CoT to assess the NLG outputs. The evaluation results are formatted as a form. Additionally, the probabilities associated with the output rating tokens can be leveraged to refine the final metric.
% 
\shorten{G-EVAL~\cite{Liu23GEval} evaluates NLG outputs by prompting an LLM with only a task description and evaluation criteria. The model then generates a chain-of-thought outlining evaluation steps, which, together with the prompt, is used to assess the output and produce a form-style rating. The metric can also incorporate the probabilities of the rating tokens to refine the final score.}

Simulation is another form of dialogue evaluation, which is primarily used for assessing target-guided open-domain dialogue systems~\cite{tang2019target, yang2022topkg}. This method involves two dialogue agents engaging in a conversation, after which the success rate in reaching the predefined target is automatically calculated.

\subsection{Human Evaluation}
% === Intro
The analysis of a conversational model's performance becomes more comprehensive with the inclusion of human evaluations~\cite{Smith22Human}. Although automatic metrics assess certain aspects of model performance and offer speed, efficiency, and reproducibility, they often correlate weakly with human judgment and may not capture all the nuances of conversational competence~\cite{Dinan2019ConvAI2}. In human evaluation, raters evaluate how effectively models manage realistic and engaging conversations~\cite{Deriu21Surveyeval}.
However, human evaluations have their own limitations. A significant challenge is the lack of comparability across different studies due to variations in experimental settings. These variations can include differing criteria such as naturalness, informativeness, context relevance, and answer accuracy.%, and scoring rates. 
\shorten{Various methods are used for human evaluation of conversations. \citet{Smith22Human} categorize these methods along two features: per-turn vs.\ per-dialogue, and pairwise vs.\ single-model evaluation. Per-turn ratings provide fine-grained analysis, while per-dialogue evaluations better capture overall coherence. Pairwise comparisons highlight subtle differences and reduce scoring bias, whereas single-model evaluations are suitable when direct comparison is unnecessary.}
% 
% ==
Five evaluation techniques have been used by combining these features:
% \begin{itemize}
\begin{itemize}[leftmargin=*]
    \item Pairwise Per-Turn Evaluation (PW-Turn): This technique provides annotations for every turn of a conversation. It requires a crowdworker to choose between a pair of model responses after each message.

    \item Pairwise Per-Dialogue Evaluation (PW-Dialog): This method asks evaluators to choose between two models by presenting a pair of complete conversations.

    \item Pairwise Per-Dialogue Self-Chat Evaluation: This approach compares two conversations that involve only the two bots talking to each other, assessing their interaction without human input.

    \item Single-Model Per-Turn Evaluation (SM-Turn): In this technique, a crowdworker interacts with a conversational agent powered by a single model. The worker annotates each response from the model based on engagement, human-likeness, and interest.

    \item Single-Model Per-Dialogue Evaluation (SM-Dialog): Similar to SM-Turn, this method involves evaluating a single model’s performance over an entire conversation, with the assessment occurring after the conversation has concluded.
\end{itemize}

After comparing various evaluation techniques, \citet{Smith22Human} draws several conclusions. Firstly, the PW-Turn are effective when differences in models’ responses are easily noticeable, such as when models are trained on different datasets. Secondly, PW-Dialog are most effective when differences between models become apparent over several conversation turns, for example, differences in the average length of conversations. Lastly, single-model evaluations are well-suited for comparing models that are similar but vary slightly in quality, such as models with different numbers of parameters.

\section{Task-oriented Conversational Data Generation}\label{sec:tod}

Task-Oriented Dialogues (TOD) represent a specific type of conversational data that contains structured interactions aimed at accomplishing specific tasks or objectives, such as booking a flight ticket or making a restaurant reservation~\cite{wen2017Applications, fellows2021Applications}. These dialogues play a pivotal role in areas such as customer service and virtual assistance, where they must adhere to particular requirements and constraints. For instance, successfully booking a restaurant table requires verifying the location's availability, matching the cuisine with the user's preference, and securing a reservation that accommodates the size of the user's party.

TOD dialogues are often used in areas such as customer service and virtual assistance, and consequently, they are constructed around an \textbf{entity} such as \textit{Restaurant}, \textit{Customer}, or \textit{Movie}. As a result, factuality is a very important aspect of TOD systems, meaning all generated entity-centric facts must be constructed or verified using existing \textbf{predefined knowledge}. 
%Entity-centric facts are monitored using a dialogue state tracking task, forming a list of entity-based \textbf{slots} and \textbf{slot values} that is stored in memory. A \textbf{slot} is a domain or task-specific predefined attribute, while the \textbf{slot value} is its associated piece of information. An example of slots and slot values for the restaurant reservation task is the following: \textit{\{party\_size: two people, cuisine: french, date: January 25th, time: 7:00 PM\}}. 
The \textbf{predefined knowledge} used for fact verification is represented using complex graphical structures, which in addition to entities, also contain information on the relationship between them. The relationship between entities shows dependencies of the task, for example, a \textit{customer} must have at least one attribute clearly defined, such as \textit{name}, before making a \textit{reservation}. Relationships between entities are crucial for generating dialogue sequentially and logically, enhancing the system's ability to manage complex interactions and dependencies effectively.

\begin{figure}[t]
    \centering
    \includegraphics[width=0.8\textwidth]{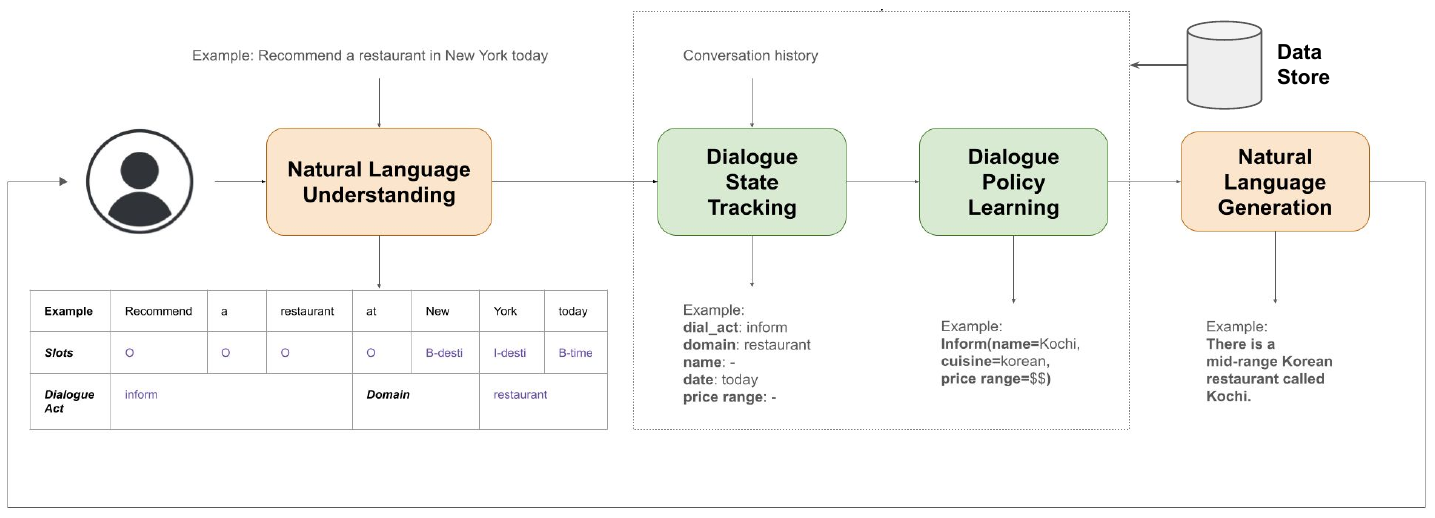}
    % \shrink
    \caption{Architecture of a Task-Oriented Dialogue (TOD) system, composed of four modules communicating in a pipeline-fashion. The input and output of each module are illustrated, indicating the process of receiving a user conversational turn, identifying intents, slots and values, extracting response attributes from the data store, and generating them into a natural language response.}
    \label{fig:tod_pipeline}
    % \shrink
\end{figure}

\medskip \noindent
\emph{\textbf{TOD Pipeline Modules.}} 
\note{TOD systems are mainly composed of four modules, namely Natural Language Understanding, Dialogue State Tracking, Dialogue Policy Learning, and Natural Language Generation. These modules can be either compressed into an end-to-end pipeline, or often follow a modular approach as shown in Figure \ref{fig:tod_pipeline}. We will describe the function, input, and output of each of these modules:}
% \begin{itemize}
\begin{itemize}[leftmargin=*]
    \item \textbf{Natural Language Understanding (NLU).} This module receives as input a conversational turn in natural language form. The goal is to process the input and extract intents, slots and values for the identified slots. 
    \item \textbf{Dialogue State Tracking (DST).} This module receives as input the conversation history and output of the NLU module (which corresponds to the current turn of the dialog) and produces the necessary slots that should be filled to approach the user goal.
    \item \textbf{Dialogue Policy Learning (DPL).} This module receives as input the slots that must be filled in, and outputs values that would be satisfactory next actions based on the current dialogue state.
    \item \textbf{Data Store.} When predefined knowledge is provided to the system (in the form of a Database, Knowledge base, ontology or schema), both the DST and DPL modules extract slots and values from these sources. If this predefined knowledge is not provided, extra modules are added to learn how to extract these slots and values from provided training data.
    \item \textbf{Natural Language Generation} receives as input the DPL output, and converts it into natural language representation.
\end{itemize}

\if 0
\begin{figure}[t]
    \centering
    \includegraphics[width=0.85\textwidth]{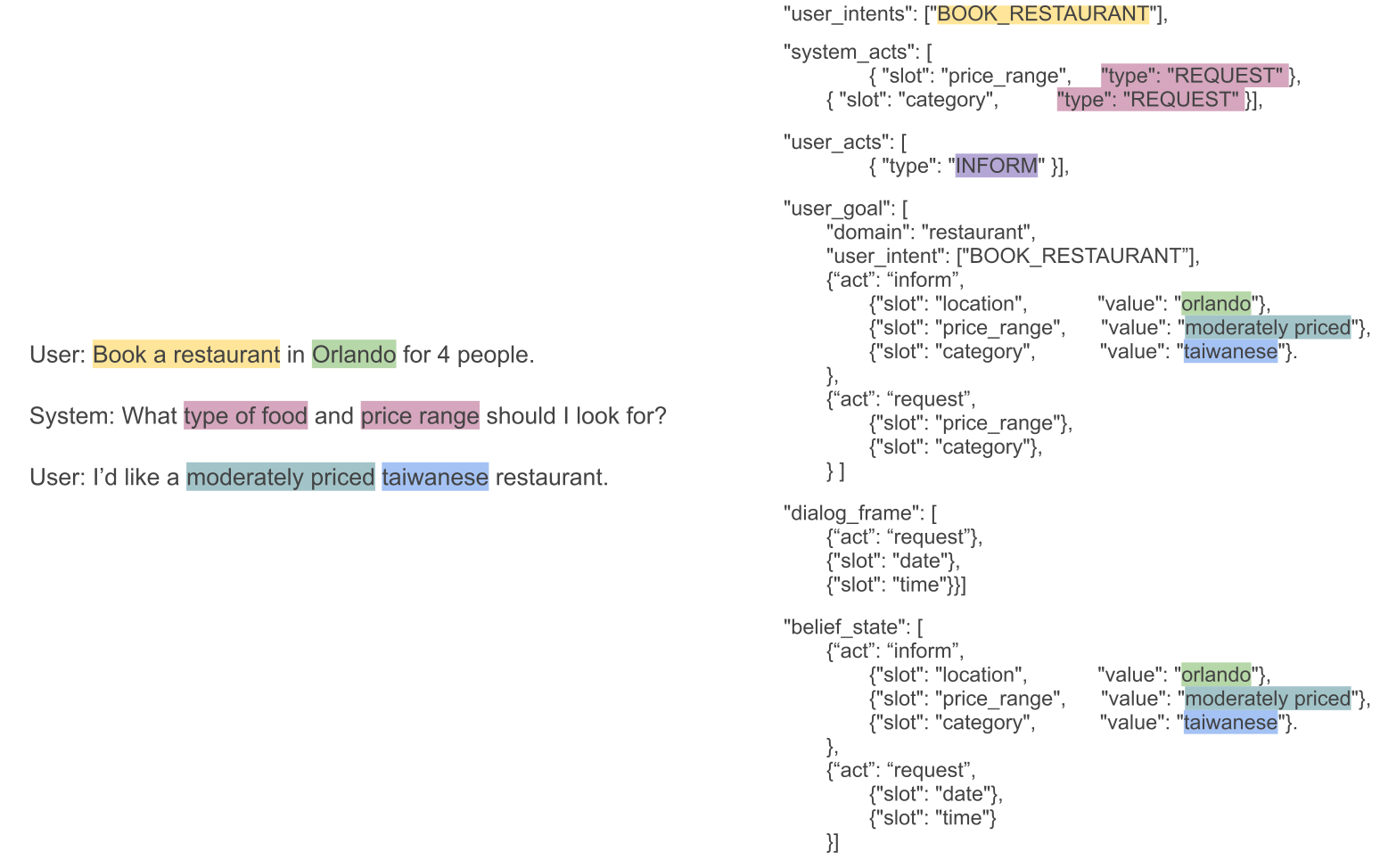}
    % \shrink
    \caption{On the left, a TOD example; On the right, the associated key terms, slots and their values.}
    \label{fig:tod_structure}
    % \shrink
\end{figure}
\fi

\paragraph{\textbf{Key terms in TOD}} 
\note{Having introduced the main graphical representations used for entity-centric dialogue generation, we will now define additional key terms that will recurrently appear throughout this section (see an illustrative example in the electronic appendix). 
% \todo{Figure~\ref{fig:tod_structure} illustrates on the left side an example dialogue composed of three turns, and on the right side the associated values for key terms such as intent, dialogue act, user goal, dialog frame and belief state}, which we will define as follows:
}
% \begin{itemize}
\begin{itemize}[leftmargin=*]
    \item \textbf{Intent.}  Represents the possible actions that a user can pursue. Intents are task-specific. Examples are \textit{Book Movie}, \textit{Reserve Restaurant}, \textit{Set Alarm}.
    \item \textbf{Dialogue Act.} A label showing the speech act of a conversational turn; A dialogue act can be associated to either the system or the user, i.e., they can be referred to as \textbf{System Act} or \textbf{User Act} respectively. Common values are \textit{Inform}, and \textit{Request}.
    \item \textbf{(User) Goal.} A set of slot and slot values that indicate what the user wants to achieve from the conversation. Example: <date=``tomorrow”, time=``8PM”, restaurant=``LaCongerie”, cuisine=``french”, party\_size=``6">.
    \item \textbf{Dialogue Frame.} A tuple containing a dialogue act and a slot-value map.
    Example: \textit{Inform}<date=``tomorrow”, time=``8PM”, restaurant=``LaCongerie”, cuisine=``french”>.
    \item \textbf{Belief State / Dialogue State.} A tuple of dialogue acts (\textit{Inform}, \textit{Request}) that keeps track of what information has been achieved, and what yet has to be requested in order to reach the user goal. \textit{Inform} represents a set of slots and slot values that indicate the user constrains. \textit{Request} represents a set of slots, without the values, that indicate what the agent wants to extract in the remainder of the dialogue.
    Example: \textit{Inform}<date=``tomorrow”, time=``8PM”, restaurant=``LaCongerie'', cuisine=``french”>, \textit{Request}<party\_size>.
    \item \textbf{Slot and Value} A domain or task-specific predefined attribute and its associated piece of information that are monitored using dialogue state tracking. An example of slots and slot values for the restaurant reservation task is the following: \textit{\{party\_size: two people, cuisine: french, date: January 25th, time: 7:00 PM\}}.  
\end{itemize}

% \paragraph{\textbf{Turn Generation}} 
% Dialog turns are generated either at the utterance-level (Turn By Turn) or dialog-level (One Go). All TOD papers discussed in this study have utterance-level generation.

% \paragraph{\textbf{TOD Split}} 
% As previously mentioned, a non-trivial aspect of TOD dialogues is factuality: DBs and KGs define attributes (slots) that are bounded by real values, while schemas and ontologies only outline the semantic dependencies between different entities. Therefore, both DBs and KGs constrict the slots to have specific values, while a schema/ontology can generate text in which the hour 25AM exists and is valid. 

\begin{figure}[t]
    \centering
\includegraphics[width=0.85\textwidth]{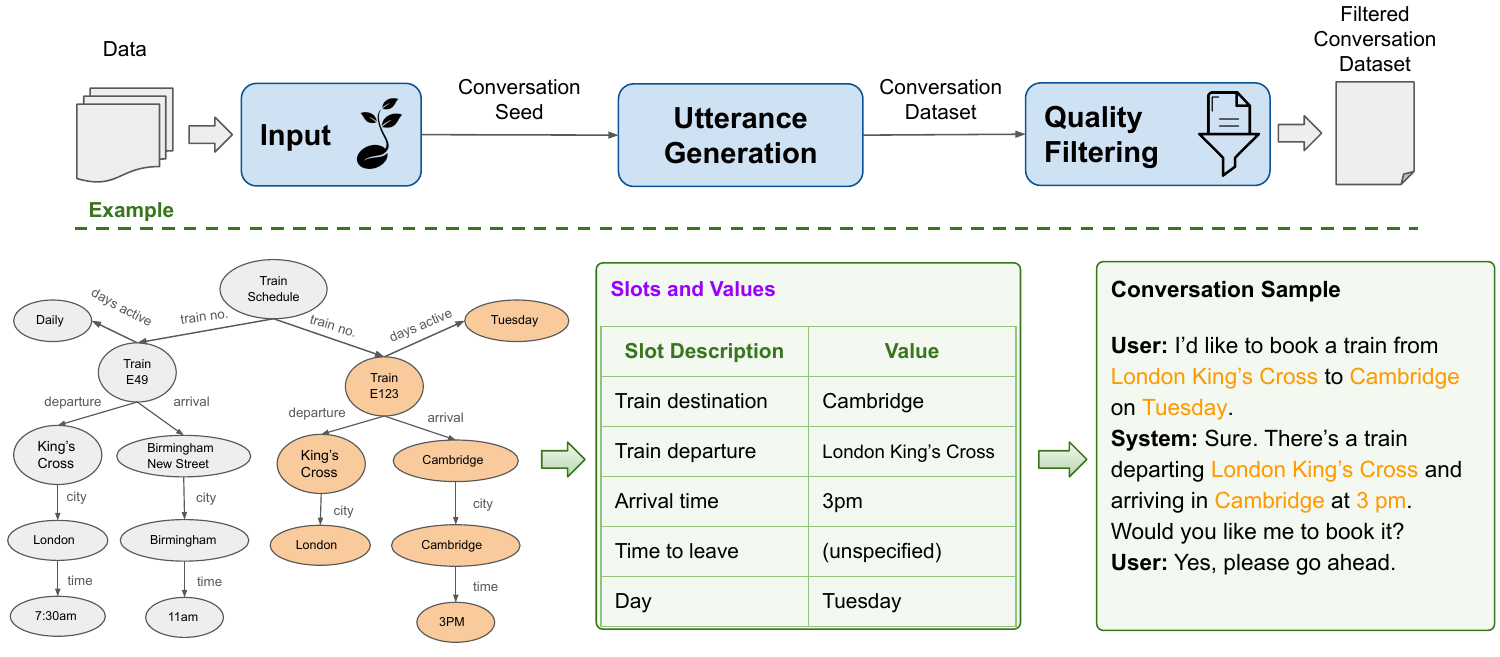}
    % \shrink
    \caption{\note{Overview of data generation in TOD. In this example, the system has access to a knowledge base containing predefined slots and real-world values (e.g., \textit{departure}, \textit{destination}, \textit{time}), which are sampled or extracted as input. The system uses the sampled slots and values to generate natural language utterances. While many methods rely on predefined templates or structured values, not all do; alternative approaches can instead learn slot–value structures directly from training data.}}
    \label{fig:tod_questions}
    % \shrink
\end{figure}

\begin{figure}[t]
\centering % Centers the figure
\begin{tikzpicture}
% [node distance=0.75cm]

% === Main Nodes (layer 1) ==
\node (odd_root) [root] {\small Task-Oriented Conversation Generation};
\node (2_utr) [odd_layer1_2, right=-0.25cm and 0.5cm of odd_root] {\small (2) Utterance Generation};
\node (1_inp) [odd_layer1_1, above=2.4cm of 2_utr] {\small (1) Input};
\node (3_flt) [odd_layer1_3, below=2.8cm of 2_utr] {\small (3) Quality Filtering};

% === Layer 2 ===============
\node (1_2_tkg) [odd_layer2_1, below right=-0.35cm and 0.5cm of 1_inp, inner sep=2pt] {\fontsize{7}{8.4}\selectfont No Input (from training data)};
\node (1_1_dia) [odd_layer2_1, above right=-0.35cm and 0.5cm of 1_inp, inner sep=2pt] {\fontsize{7}{8.4}\selectfont Provided (KG, DB, Schema, Ontology)};
% \node (1_3_prf) [odd_layer2_1, below=0.15cm of 1_2_tkg] {Personalized Profile};
% 

\node (2_1_1go) [odd_layer2_2, above right=1.2cm and 0.5cm of 2_utr, inner sep=2pt] {\fontsize{7}{8.4}\selectfont Provided Language Template};
\node (2_1_2go) [odd_layer2_2, above right=0.4cm and 0.5cm of 2_utr, inner sep=2pt] {\fontsize{7}{8.4}\selectfont Natural Language Around Input};
\node (2_1_3go) [odd_layer2_2, above right=-0.4cm and 0.5cm of 2_utr, inner sep=2pt] {\fontsize{7}{8.4}\selectfont Natural Language Conditioned on Input};

\node (2_2_4tbt) [odd_layer2_2, below right=-0.4cm and 0.5cm of 2_utr, inner sep=2pt] {\fontsize{7}{8.4}\selectfont Variational Inference};
\node (2_2_5tbt) [odd_layer2_2, below right=0.4cm and 0.5cm of 2_utr, inner sep=2pt] {\fontsize{7}{8.4}\selectfont Few-shot Learning};
\node (2_2_6tbt) [odd_layer2_2, below right=1.2cm and 0.5cm of 2_utr, inner sep=2pt] {\fontsize{7}{8.4}\selectfont In-Context Learning};

\node (3_2_con) [odd_layer2_3, below right=-0.4cm and 0.5cm of 3_flt, inner sep=2pt] {\fontsize{7}{8.4}\selectfont Factuality Check during Post-Processing};
\node (3_1_nos) [odd_layer2_3, above=0.5cm of 3_2_con, inner sep=2pt] {\fontsize{7}{8.4}\selectfont Ensured with Provided Input};
\node (3_3_per) [odd_layer2_3, below=0.15cm of 3_2_con, inner sep=2pt] {\fontsize{7}{8.4}\selectfont None};

% === Layer 3 ===============
\node (1_1_1_ppr) [odd_layer3_1, right=0.5cm of 1_1_dia, inner sep=2pt] {\fontsize{7}{8.4}\selectfont ABUS~\cite{li2017ABUS}, M2M~\cite{Shah2018M2M}, SGD~\cite{Rastogi2020SGD}, NUS~\cite{kreyssig2018neural}, NeuralWOZ~\cite{kim2021neuralwoz}, HUS~\cite{gur2018HUS}, VHUS~\cite{gur2018HUS}, TUS~\cite{lin2021TUS}, Joust~\cite{tseng2021DSUSRL}, Unified-US~\cite{wan2022unified}, *IND~\cite{ahmad2023ina}};

\node (1_2_1_ppr) [odd_layer3_1, right=0.5cm of 1_2_tkg, inner sep=2pt] {\fontsize{7}{8.4}\selectfont VHDA~\cite{yoo2020VHDA}, *Simulated-Chat~\cite{mohapatra2021simulatedchats}, *ICL-US~\cite{Terragni2023ICLUS}, *Dialogic~\cite{li2022DIALOGIC}, *INA~\cite{ahmad2023ina}};
% \node (1_3_1_ppr) [odd_layer3_1, right=0.5cm of 1_3_prf] {PERSONACHATGEN~\cite{lee2022personachatgen}, SPC~\cite{Jandaghi23spc}};

\node (2_1_1_ppr) [odd_layer3_2, right=0.5cm of 2_1_1go, inner sep=2pt] {\fontsize{7}{8.4}\selectfont ABUS~\cite{li2017ABUS}, M2M~\cite{Shah2018M2M}, SGD~\cite{Rastogi2020SGD}};
\node (2_1_2_ppr) [odd_layer3_2, right=0.5cm of 2_1_2go, inner sep=2pt] {\fontsize{7}{8.4}\selectfont NUS~\cite{kreyssig2018neural}, NeuralWOZ~\cite{kim2021neuralwoz}};

\node (2_1_3_ppr) [odd_layer3_2, right=0.5cm of 2_1_3go, inner sep=2pt] {\fontsize{7}{8.4}\selectfont HUS~\cite{gur2018HUS}, VHUS~\cite{gur2018HUS}, TUS~\cite{lin2021TUS}, Joust~\cite{tseng2021DSUSRL}, Unified-US~\cite{wan2022unified}, *IND~\cite{ahmad2023ina}};
\node (2_2_4_ppr) [odd_layer3_2, right=0.5cm of 2_2_4tbt, inner sep=2pt] {\fontsize{7}{8.4}\selectfont VHDA~\cite{yoo2020VHDA}};

\node (2_2_5_ppr) [odd_layer3_2, right=0.5cm of 2_2_5tbt, inner sep=2pt] {\fontsize{7}{8.4}\selectfont *Simulated-Chat~\cite{mohapatra2021simulatedchats}};
\node (2_2_6_ppr) [odd_layer3_2, right=0.5cm of 2_2_6tbt, inner sep=2pt] {\fontsize{7}{8.4}\selectfont *ICL-US~\cite{Terragni2023ICLUS}, *Dialogic~\cite{li2022DIALOGIC}, *INA~\cite{ahmad2023ina}};

\node (3_1_1_ppr) [odd_layer3_3, right=0.5cm of 3_1_nos, inner sep=2pt] {\fontsize{7}{8.4}\selectfont ABUS~\cite{li2017ABUS}, M2M~\cite{Shah2018M2M}, SGD~\cite{Rastogi2020SGD}, NUS~\cite{kreyssig2018neural}, NeuralWOZ~\cite{kim2021neuralwoz}, HUS~\cite{gur2018HUS}, VHUS~\cite{gur2018HUS}, TUS~\cite{lin2021TUS}, Joust~\cite{tseng2021DSUSRL}, Unified-US~\cite{wan2022unified}, IND~\cite{ahmad2023ina}, *Simulated-Chat~\cite{mohapatra2021simulatedchats}};
\node (3_2_1_ppr) [odd_layer3_3, right=0.5cm of 3_2_con, inner sep=2pt] {\fontsize{7}{8.4}\selectfont *ICL-US~\cite{Terragni2023ICLUS}, *Dialogic~\cite{li2022DIALOGIC}, *IND~\cite{ahmad2023ina}, *INA ~\cite{ahmad2023ina}};
\node (3_3_1_ppr) [odd_layer3_3, right=0.5cm of 3_3_per, inner sep=2pt] {\fontsize{7}{8.4}\selectfont VHDA~\cite{yoo2020VHDA}};

% % Arrows
\draw (odd_root.east) -- ++(5pt,0) |- (1_inp.west);
\draw (odd_root.east) -- ++(5pt,0) |- (2_utr.west);
\draw (odd_root.east) -- ++(5pt,0) |- (3_flt.west);

\draw (1_inp.east) -- ++(5pt,0) |- (1_1_dia.west);
\draw (1_inp.east) -- ++(5pt,0) |- (1_2_tkg.west);

\draw (2_utr.east) -- ++(5pt,0) |- (2_1_1go.west);
\draw (2_utr.east) -- ++(5pt,0) |- (2_1_2go.west);
\draw (2_utr.east) -- ++(5pt,0) |- (2_1_3go.west);
\draw (2_utr.east) -- ++(5pt,0) |- (2_2_4tbt.west);
\draw (2_utr.east) -- ++(5pt,0) |- (2_2_5tbt.west);
\draw (2_utr.east) -- ++(5pt,0) |- (2_2_6tbt.west);

\draw (3_flt.east) -- ++(5pt,0) |- (3_1_nos.west);
\draw (3_flt.east) -- ++(5pt,0) |- (3_2_con.west);
\draw (3_flt.east) -- ++(5pt,0) |- (3_3_per.west);

\draw (1_1_dia) -- (1_1_1_ppr);
\draw (1_2_tkg) -- (1_2_1_ppr);

\draw (2_1_1go) -- (2_1_1_ppr);
\draw (2_1_2go) -- (2_1_2_ppr);
\draw (2_1_3go) -- (2_1_3_ppr);

\draw (2_2_4tbt) -- (2_2_4_ppr);
\draw (2_2_5tbt) -- (2_2_5_ppr);
\draw (2_2_6tbt) -- (2_2_6_ppr);

\draw (3_1_nos) -- (3_1_1_ppr);
\draw (3_2_con) -- (3_2_1_ppr);
\draw (3_3_per) -- (3_3_1_ppr);

\end{tikzpicture}
\caption{Overview of approaches used for Task-Oriented Dialogue generation. The generation process is composed of three components, described in sections \ref{sec:tod_input} - \ref{sec:quality_check}. \note{The models marked with an asteriks * are LLM-based.}}
\label{fig:tod_split}
% \shrink
\end{figure}

Having defined the key components and notation of TOD systems, we outline three stages in the TOD generation process: (1) input, (2) utterance generation, and (3) quality filtering, as shown in Figure \ref{fig:tod_questions}. 
In Section \ref{sec:tod_input}, we provide an overview of different resources that are used as input for TOD systems. We then describe  data generation methods for TOD systems in Section~\ref{sec:generation_training}.
% Having defined the possible inputs, key notations, training methods and the importance of graphical structures for guiding TOD generation, identifying an effective system bottles down to addressing three base questions as presented in Figure \ref{fig:tod_questions}. Moreover, we can outline three stages in the TOD generation process: 1) Input, 2) Generation Method, and 3) Quality Filtering. 
% In Section \ref{sec:generation_training} we will connect 2) Generation Method together with training and simulation, as they are strongly tied with each other. 
We first split TOD approaches with respect to the external knowledge available, i.e., the input of the data generation system in the form of slots and their possible values. As presented in Section ~\ref{sec:tod_input}, this external knowledge appears in the form of a knowledge base, database, schema, or ontology. If provided, the corresponding slots and slot values are plugged into the dialogue system, and natural utterances are generated around them (see Sections \ref{sec:predefined_templates}-\ref{sec:conditioned_on_input}). If they are not provided, the dialogue system learns them through training data (see Sections \ref{sec:variational_inference}-\ref{sec:in_context_learning}). However, it is important to note that generating slots and values from the mapped latent space does not guarantee factual/truthful generations. Lastly, we describe the quality filtering step in Section~\ref{sec:quality_check}. An overview alongside existing work is presented in Figure \ref{fig:tod_split}.

\subsection{Input}\label{sec:tod_input}
There are several key graphical representations that TOD systems need as their input: (1) the \textbf{Schema} defines the entities, their attributes (slots) and which attributes link the entities; (2) the \textbf{Ontology} builds on top of the schema by adding information on the relationship between entities, and is often used for querying a knowledge graph; (3) the \textbf{Knowledge Graph (KG)} or \textbf{Data Base (DB)} serves as a repository of domain-specific information. The KG is an instantiation of an ontology with all available entities, while the DB is an instantiation of a schema. Understanding these elements is essential for developing robust and efficient TOD systems. We will outline each of these components, as they will re-appear in some of the methodologies we further discuss in this study. \note{Concrete examples of schemas, ontologies, and knowledge graphs used in TOD systems are provided in the electronic appendix.}

\begin{enumerate}[leftmargin=*]
    \item \textbf{Schema and Ontology}.
    These are general structures that describe entities, such as \textit{Restaurant} and \textit{Reservation}, and entity attributes, such as \textit{Location} for \textit{Restaurant}, and \textit{Party Size} for \textit{Reservation}. We differentiate between two commonly used structures:

    \textbf{Schema}: Represents a set of entities, where each entity has attributes (slots). The schema models the task in hand, e.g., a restaurant reservation involves information about the date, time, party size, and any special requests. 
    % \todo{An example is illustrated in Figure \ref{fig:schema_example}}.

    \textbf{Ontology}: Represents a set of entities, where each entity has attributes (slots), \textit{and the relationship between entities}, for example \textit{Reservation} "is made by" a \textit{Customer}. Often ontologies are considered more semantic-forward than schemas, the reason being the added meaning they inherit through these entity relationships. 
    % \todo{An example is illustrated in Figure \ref{fig:ontology_example}}. \\
    
    % \underline{Goal}: The dialog system can use these general structures to ask relevant questions. They contain more information about the semantics of the dialogue than about specific instantiations of entities.
    
    % \underline{Limitation}: General structures do not contain real-world data or restrictions on the possible slot values. For data generation, this means that a dialogue may evolve around combinations of slot values that do not exist, e.g. a restaurant called \textit{Moeders} that specializes in \textit{japanese cuisine}.

    \item \textbf{Knowledge Graph and Data Base}.
    Structured data representations that model real-world entities (instantiated entities), their attributes, and the relationship between them. Knowledge graphs can be class-specific, or integrated (multiple class objects in the same graph). 
    % \todo{Figure \ref{fig:KG_example} shows an example of a KG.}\\
    % \underline{Goal}: Links to real entities and is updated in real-time. \\
    % \underline{Limitation}: Difficult to build for every problem.
\end{enumerate}

In addition to these graphical representations, some TOD systems are built to operate without a predefined input. Such methods often rely on training datasets from which they capture domain or task-specific knowledge. Not relying on input offers flexibility, however, it does not guarantee faithful generations or a method of verifying the information.

% \begin{figure}[t]
%     \centering
%     \includegraphics[width=0.96\textwidth]{figs/TOD/TOD_split.png}
%     \caption{TOD split.}
%     \label{fig:tod_split}
% \end{figure}

\subsection{Utterance Generation}\label{sec:generation_training}

\if 0
\begin{wrapfigure}{r}{0.55\columnwidth}  % r = right, l = left
    \centering
    \begin{tikzpicture}[scale=0.55, every node/.style={scale=0.70}]
        % Nodes
        \node[align=center] (restaurant) at (0,0) {Restaurant: Name, Location, Cuisine Type, \\ Price Range, Rating, Opening Hours};
        \node[align=center] (reservation) at (8,0) {Reservation: Date, Time, Party Size, \\ Special Requests};
        \node[align=center] (customer) at (4,-2) {Customer: Name, Contact Information, Preferences};
    
        % Arrows with text
        \draw[->, >=latex] (restaurant) to[bend left=20] node[above] {offers} (reservation);
        \draw[->, >=latex] (reservation) to[bend left=20] node[below right] {is made by} (customer);
    \end{tikzpicture}
    \caption{Example of an ontology for the restaurant reservation task.}
    \label{fig:ontology_example}
    % \shrink
\end{wrapfigure}
\fi

Various techniques have emerged for constructing dialogue from \textbf{provided input} in the form (semi)structured, predefined knowledge, which can be grouped into three main approaches: (1) predefined language templates, (2) wrapping the slots and values in natural language, and (3) generating natural language conditioned on the input. There are also methods that operate \textbf{without predefined input}. These methods adopt an approach centered on discovering entity-based slots and values through learning.
These strategies typically rely on training datasets often curated by human annotators; alternatively, systems may be trained or fine-tuned on artificially generated dialogues. We identify three distinct approaches for achieving latent-space discovery of TOD-specific entity attributes, (1) variational inference, (2) few-shot learning, and (3) in-context, discussed in Sections \ref{sec:variational_inference}--Section \ref{sec:in_context_learning}.
% :   namely in (1) Section \ref{sec:variational_inference} through variational inference, (2) Section \ref{sec:few_shot} few-shot learning and (3) Section \ref{sec:in_context_learning} in-context learning. 
\note{Methods that rely on provided input do not require training, while variational inference, few-shot learning, and in-context learning require training, fine-tuning, or contextual learning at inference time.}
%\note{We note that the methods t available input naturally introduces different generation paradigms: template and input-conditioned methods do not require training, while variational inference, few-shot learning, and in-context learning require training, fine-tuning, or contextual learning at inference time.}

\if 0
\begin{wrapfigure}{r}{0.55\columnwidth}
    \centering
    \includegraphics[width=0.55\textwidth]{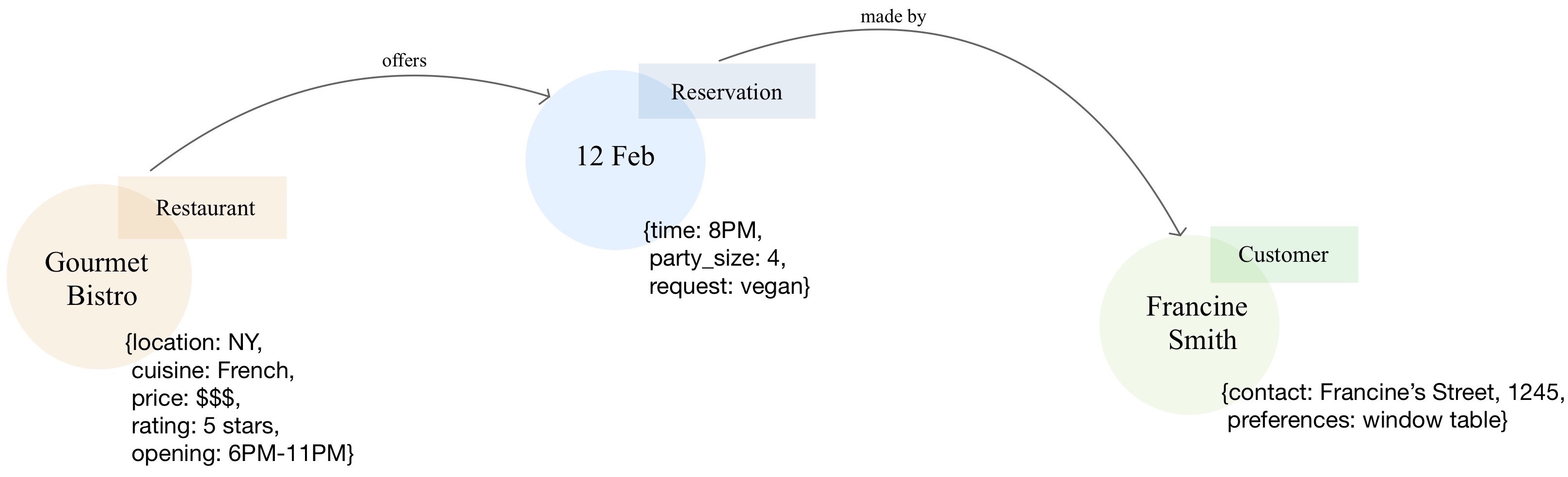}
    % \shrink
    \caption{Example of a KG for the restaurant reservation task.}
    \label{fig:KG_example}
    % \shrink
\end{wrapfigure}
\fi
\note{We note at the outset that a conversation inherently involves at least two participants, any TOD system requires a conversational partner (something akin to a user) to produce replies.  
% \note{Since a conversation involves at least two participants, any TOD system can generate utterances, but also requires a conversational partner - something akin to a user - to produce replies. 
The concept of \textbf{simulating} a user to interact with the TOD has been previously established~\cite{meng2003Simu, georgila2006Simu, jaderberg2017Simu, narayanan2002Simu, schatzmann2007Simu}. In TOD data generation, the conversation partner can be a human or a simulator, which can be another pre-trained dialogue system, or a dialogue system that is co-trained alongside the TOD, a paradigm which is often referred to as a two-sided simulation. This method is frequently used in conjunction with reinforcement learning policies, allowing the two simulators to iteratively enhance their performance through mutual interaction. 
In this paper, we focus on the data generation aspects, and refer the reader to the book by ~\citet{Balog:2024:fntir} for further details on user simulation.}

% \begin{figure}[t]
%     \centering
%     \begin{tikzpicture}[scale=0.8, every node/.style={scale=0.8}]
%         % Nodes
%         \node[align=center] (restaurant) at (0,0) {Restaurant: Name, Location, Cuisine Type, \\ Price Range, Rating, Opening Hours};
%         \node[align=center] (reservation) at (8,0) {Reservation: Date, Time, Party Size, \\ Special Requests};
%         \node[align=center] (customer) at (4,-2) {Customer: Name, Contact Information, Preferences};
    
%         % Arrows with text
%         \draw[->, >=latex] (restaurant) to[bend left=20] node[above] {offers} (reservation);
%         \draw[->, >=latex] (reservation) to[bend left=20] node[below right] {is made by} (customer);
%     \end{tikzpicture}
%     \caption{Example of an ontology for the restaurant reservation task.}
%     \label{fig:ontology_example}
%     \shrink \miniskip
% \end{figure}

% \todo{mention the path from input to slot-value pairs}

% \todo{update training and simulation part and align it with current categories mentioned here.}
%%%%%%%%%%%%%%%%%%%%%%%%%%%%%%%%%%
% \paragraph{\textbf{TOD Data Generation.}} 
% Training TOD systems presents considerable challenges due to the need for highly specialized datasets that accurately capture diverse user goals, system actions, and constraints. These limitations not only reduce the breadth of available training data, but also require careful system design. In this context, we will explore two critical aspects of TOD systems, namely training and simulation.

\subsubsection{Predefined Language Templates}\label{sec:predefined_templates}

Utilizing predefined language templates, referred to as an \textit{agenda-based} or \textit{schema-based} approach, ensures that the generated conversations follow a predetermined structure. The system then utilizes a series of responses that align with specific conversational goals or tasks, guiding the dialogue according to a pre-established \textbf{outline} or \textbf{schema}, which have pre-defined language templates for each intent and slot type. For example, the intent \textit{<book$\_$movie>} can be associated with the template \textit{"Book movie with [name="value"] and [date="value"].} For a specific movie and date, such as \textit{inform<intent=book$\_$movie, name=Inside Out, date=tomorrow>}, the template is filled and generates the turn \textit{"Book movie with name Inside Out and date is tomorrow."}. Often, agenda- or schema-based TOD systems contain a human-centered paraphrasing step, or an NLG module, to convert these generations into more diverse and intelligible human dialogue.

One paper that follows an agenda-based approach is ABUS~\cite{li2017ABUS}. The input consists of a set of rules (a stack-like representation of user states), and example dialogues (used for building an accompanying user simulator). The simulator sends real-time simulated utterances to the system, which is trained to reply using an RL policy. This approach addresses the challenge of training dialogue systems to respond accurately and effectively in real-world scenarios, facilitated by a publicly available simulation framework.
A drawback of agenda-based approaches is that they are task-dependent. M2M~\cite{Shah2018M2M} has access to multiple \textit{APIs} alongside a task specification, where each API has access to a task schema, outlining the scope and restrictions of interactions for the task in question. The generation process of M2M can be divided into two parts: first, using a \textbf{task specification} that contains a schema corresponding to an API; second, using \textbf{task-independent information} that maps the task specification to a set of dialogues.
Moreover, SGD~\cite{Rastogi2020SGD} addresses the fact that in the real world, multiple services have overlapping functionality. The authors advocate for building a single unified dialogue model for all services and APIs by allowing access to a schema that provides a dynamic set of intents and slots, along with their natural language descriptors. Using dynamic schemas for each API allows for sharing knowledge between the services. SGD builds on top of M2M~\cite{Shah2018M2M} by generalizing it to multiple user-system simulations, called agents, spanning over 26 services covering 16 domains, and resulting in a 16k dialogue dataset.
%Similarly to M2M and ABUS~\cite{li2017ABUS}, the framework leverages simulator interactions to generate dialogue outlines. While they still leverage crowd-sourcing for natural language generation, they introduce an intermediate step of converting the generated outlines into short utterances. More precisely, the set of generated outlines is first mapped to utterances such as ``INFORM(location=”Los Angeles”)” to  ``Which city are you in?” Then, the crowd workers paraphrase the list of utterances to ensure naturalness and coherence. This approach addresses the challenge of reducing API calls by leveraging overlapping functionalities among services, reducing redundancy, which minimizes costs and response times.

\subsubsection{Natural Language Around Input}\label{sec:around_input}

As a response to being limited by predefined language templates, ~\citet{kim2021neuralwoz} define three strategies. First, augmenting the knowledge base (KB) to enrich the semantic complexity of the dialogue system without altering its operational constraints. Second, modifying the schema itself by either introducing new constraints, removing existing ones, or altering the values these constraints can assume. Third, which is the focal point of this subsection, is directly generating dialogues in natural language form, extending the functionality of dialogue systems to previously unseen domains. This method circumvents the limitations imposed by rigid schema constraints and opens future research directions for more dynamic and versatile interactions.

The Neural User Simulator (NUS) ~\cite{kreyssig2018neural} eliminates hand-crafted rules for natural language generation by utilizing a corpus-driven approach to generate natural language utterances. NUS leverages dynamic goal generation, meaning that the system can dynamically change the user goal with more dialogue context, assuming the user would want to shift their goal mid-conversation. It uses the ontology created with the dialogue State Tracking Challenge 2 (DSTC2)~\cite{henderson2014DSTC2} dataset's evolving states, from which the system can extract different slots and, together with the dialogue history, can generate a dynamic goal. %The fundamental distinction from ABUS lies in NUS's ability to generate language: it does not resort to stitching predefined natural language templates to slot-value pairs but instead learns to construct utterances through a sequence-to-sequence model, aligning goal states with the ontology at every step.
NeuralWOZ~\cite{kim2021neuralwoz} presents a novel system composed of two sequential neural models, namely the Collector and the Labeler. The Collector constructs dialogues given as input a user goal instruction and API call results. These elements synthesize into a belief state encapsulating both informable and requestable slot and slot values. The Collector generates the dialogue, and Labeler then annotates it with possible slots and slot values from the belief state extracted from a pre-defined list, creating a coherent dataset for dialogue state tracking.

\subsubsection{Natural Language Conditioned on Input}\label{sec:conditioned_on_input}

The Hierarchical User Simulator (HUS)~\cite{gur2018HUS} is part of the same family as the previously discussed ABUS and NUS. HUS employs a multifaceted encoding scheme to transform different dialogue features in different vector representations: (1) the user goal, (2) the current dialogue turn, and (3) the entire dialogue history. Moreover, HUS has access to a KB from which it extracts information about the task and domain of the dialog. %Although HUS has demonstrated competence in generating successful dialogues, the authors advocate for a process that is close to human-like dialogue generation. 
Due to the deterministic nature of HUS, the model will yield the same dialogue each time given identical user goals and system turns. To introduce variability, the authors propose a variational inference framework VHUS, by sampling an unobserved state from a latent space at each turn, with the latent space being positioned between the encoding of the dialogue history and the decoding of the user turn. Compared to VHDA, which is discussed in Section \ref{sec:variational_inference}, the latent space in VHUS enriches the dialogue generation with diversity without modifying the underlying user state and goal generation. Therefore, while VHUS contributes to dialogue variety through latent sampling, it does not compromise the factuality of the generated slots and values.

%%%%%%% [TUS]
Similarly, TUS~\cite{lin2021TUS} maps different inputs to different representations in the feature space, however it also distinguished itself by its unique domain-agnostic feature representations. %The input feature representation of each turn-level slot consists of a set of subvectors. The \textit{basic information} models multiple aspects of the system, such as the belief state, user goal, whether there are constraints in the current turn generation if previous constraints have been fulfilled. The \textit{system action} models the current system action, and extra domain-specific values. The \textit{user action} encodes similar information with the system action, but also includes the history of the user system, and the \textit{domain and slot index features} to avoid the situation when the first three subvectors are the same for different domains, this vector adds an index. The authors address the challenge of having a domain-dependent dialogue system by separately modeling the domain-specific data with the \textit{domain and slot index features}.
A similar work is JOUST ~\cite{tseng2021DSUSRL}, which combines two simulator agents trained on a collection of source domain dialogues to converse with each other through natural language. The pre-trained agents are fine-tuned in a low resource setting on target domain data using RL. %The authors study two aspects of JOUST: 1) capabilities of domain adaptation, and 2) single-to-multiple domain transfer. Although using two simulator agents has been previously proposed, this paper adds novelty by training them on multi-domain human-generated dialogues. 
Similarly to TUS, it is simulation-based, it has access to a knowledge base, it does not generate new values, and keeps track of the dialogue state simultaneously with the natural language dialogue turn generation. 
%We reflect upon the TUS methodology more because, compared to JOUST, it also models the dialogue state to be domain-independent.
Another paper that follows analogous work is Unified-US~\cite{wan2022unified}, %During both training and fine-tuning, the dialogue system has access to the following inputs: 1) a task description provided in natural language, 2) the dialogue context, and 3) the user goal in natural language. 
which leverages a simulator model composed of two parts, namely the user and system agents. 
The authors first pre-train both agents using a multi-domain dialogue dataset without user goals. Subsequently, they are further pre-trained on the same dialogue dataset, however, this time incorporating the user goals, and fine-tuning on a target domain (NeuralWOZ~\cite{kim2021neuralwoz}). %It is important to note that Unified-US distinguishes itself from prior works such as NeuralWOZ by removing API calls for extracting slots and values. Instead, it directly uses the slots and their values from annotated datasets.

IND ~\cite{ahmad2023ina} is a simulation-based generated dataset on negotiation dialogues. The authors create the Integrative Negotiation Dataset (IND) using GPT-J with human-in-the-loop for quality check. %This process is done in 5 steps. First, for background data, the system has inputs such as knowledge-grounded information, for which the system has access to a database consisting of data on 10 electronic items, each item having features such as Product Name, Product Description, Product Features, Price, Accessory List, and Accessory Description. Then, a set of 11 negotiating intents are defined, such as Ask, Inform, Negotiate-Price-Increase, etc. For each intent, few-shot prompts are designed to describe the task and the product data. Then, the dialogue flow generation takes each intent and its prompt and associates the relevant information from the product background data. For example, for intent \textit{Negotiate-Add-X}, the authors associate the item name, while for intent \textit{Negotiate-Price-Decrease}, the proposed price is associated. Then, they prompt GPT-J with the intents, instructions, and associated product details to create dialogue turn-by-turn. In the end, data correction is applied by involving human-in-the-loop, where experienced experts make edits to the generated dialogue to ensure grounding in the provided background database. Moreover, the experts were tasked to correct or remove intent, action, and negotiation flow inconsistencies, as well as grammatical errors. dialogues with low fluency were finally dropped from the database. 
The second part of this study uses the IND dataset to train the negotiating agent INA, which is further discussed in Section \ref{sec:in_context_learning}. The challenge of this data generation problem is that negotiation strategies are highly context-dependent, involving interactions about aspects such as price, product, delivery, quality, etc, while also answering with knowledge-grounded information such as manufacturing details about the product, making it quite a challenge for generating high-quality dialogues. The authors construct the dialogues based on the ``Integrative” approach to negotiation, i.e., a win-win situation where each party understands the other’s needs and the goal is to reach mutual satisfaction. 

\subsubsection{Variational Inference}\label{sec:variational_inference}

%%%%%%% [VHDA]
VHDA~\cite{yoo2020VHDA} is a method that receives as input human-generated dialogues, and constructs utterance-level natural language outputs together with dialogue acts, modeling latent variables over all dialogue aspects to extrapolate and generate novel slot values. This characteristic allows the system to produce attributes beyond those encountered in the training data, demonstrating an ability to innovate within the dialogue context and suggesting a higher degree of system adaptability and generative capacity. These features highly differentiate VHDA from the previous research. The authors build on top of the idea of VHCR~\cite{parl2018vhcr}, encapsulating multiple encoder and decoder modules, where each models a specific dialogue feature. %The encoder is employed as a self-attention bidirectional LSTM for mapping all previously mentioned latent spaces. As a training objective, the ELBO is optimized on the goal-oriented dialogue samples. 
VHDA can be closely compared to VHUS~\cite{gur2018HUS}, however, the main and very important difference between the two is that VHUS samples from the latent space of natural language turn generation, while VHDA also models the user goal, and dialogue state in the latent space.

\subsubsection{Few-shot learning}\label{sec:few_shot}

In this section we discuss papers that leverage the capabilities of Large Language Models (LLMs) to understand the underlying syntax and semantics of the dialogue data. Here we differentiate between learning with few-shot examples, and in-context learning. Both techniques require limited data.

% \begin{itemize}
\begin{itemize}[leftmargin=*]
    \item \textit{Few-shot learning}: the ability of a model to generalize when provided a very small dataset for \textbf{training} or \textbf{fine-tuning}.
    \item \textit{In-context learning}: the ability of a model to generalize when provided very few examples in the input prompt \textbf{without explicit training} or \textbf{fine-tuning}.
\end{itemize}

Simulated-Chat ~\cite{mohapatra2021simulatedchats} leverages the abilities of large pre-trained models to follow instructions and generate synthetic conversations. The system is split into user and agent simulators. The input of the user simulator is a set of instructions based on which it can generate a dialogue utterance, while the input of the agent simulator is a knowledge base. Given the current dialogue context, the agent generates a belief state. The two simulators hold a conversation conditioned on both the instructions and knowledge base, trained by fine-tuning GPT-2 and Longformer, first on human-generated dialogues, and then on additionally self-generated simulated dialogues. %The problem involves generating a dialogue $\mathcal{D}$ based on a given set of instructions $\mathcal{I}$ and a knowledge base $\mathcal{KB}$. The dialogue consists of user and agent utterances, and at each turn, both systems model their responses based on the dialogue context and available information. The joint distribution of the dialogue is determined by equations that model the user and agent systems' behavior. 
To use this framework for TOD generation, a belief state generator is added after the agent system. This generator produces key-value pairs that are matched against the knowledge base to retrieve relevant information, with the belief state updated using greedy methods.

% The problem can be formulated as following: Assume a dialogue $D = (u_1, a_1,...,u_n, a_n)$ where $a_i$ and $u_i$ represent a utterance generated by the user, respectively by the agent system. A sequence $c_m = (u_1, a_1,...,u_{m-1}, a_{m-1})$ is called the context dialogue at turn $m$. Given a set of instruction $\mathcal{I}$ and the knowledge base $\mathcal{KB}$, the goal is to generate a dialogue $\mathcal{D}$. The framework maps this problem in the following manner: \textbf{User system}: At each turn $i$, the user system can model the current utterance as $p(u_i|c_i, \mathcal{I}).$; \textbf{Agent system}: At each turn $i$, the agent system has access to the dialogue context, the previously generated utterance by the user system, and the knowledge base. Therefore it can model the current utterance as $p(a_i|c_i, u_i, \mathcal{KB}).$ The joint distribution of dialogue $\mathcal{D}$ can be therefore written as Equation \ref{eq:simulated_chat_dialogue_model}.

% \begin{equation}
%     p(\mathcal{D}|\mathcal{I}, \mathcal{KB}) = \prod_{i=1}^n p(u_i|c_i, \mathcal{I})p(a_i|c_i, u_i, \mathcal{KB})
%     \label{eq:simulated_chat_dialogue_model}
% \end{equation}

% For using this generator as TOD, one model must be added after the agent system, namely a belief state generator. The belief state generator outputs a sequence of key-value pairs that will be mapped over the knowledge base $\mathcal{KB}$, to return a set of entities that match the query. The current belief state is sampled with greedy approaches.

\subsubsection{In-context learning}\label{sec:in_context_learning}

% \begin{figure}[t]
%     \centering
%     \includegraphics[width=0.85\textwidth]{figs/TOD/simulation.png}
%     \caption{Two-sided simulation as presented in Simulated-Chat~\cite{mohapatra2021simulatedchats}}.
%     \label{fig:tod_simulation}
% \end{figure}

% We identify two types of simulators. Agenda-based simulators generate dialogue given a pre-defined set of entity-based rules and restrictions (see Section \ref{sec:predefined_templates}), and data-driven simulators create dialogues by learning from real data, matching patterns for creating realistic conversational flows (see Section \ref{sec:around_input}-\ref{sec:in_context_learning}).

%%%%%%%%%%%%%%%%%%%%%%%%%%%%%%%%%%

%%%%%%% [ICL-US]
ICL-US~\cite{Terragni2023ICLUS} presents a method to generate TOD dialogues through in-context learning, receiving as input a few example dialogues to perform a new task, without needing any fine-tuning. The framework is composed of a \textit{user simulator} and a \textit{dialogue system}. Given a user goal and a few in-context dialogue examples, ICL-US generates a user utterance per turn. %Including the dialogue history, each turn is conditioned on 1) a set of example dialogues $K$ shots examples, where each dialogue contains by the user goal and natural language dialogue, 2) a user goal, and 3) the conversation history. While this method eliminates the need for human annotations, the authors needs to use an existing dataset for extracting the user goal and dialogue examples. 
In-context learning requires the dialogue examples to be similar to the desired generation. The authors apply two sampling techniques to select the few-shot dialogues for each generation: 1) random sampling, and 2) sampling based on Jaccard similarity between the user goal and the candidate dialogue goals. 
Similarly, Dialogic~\cite{li2022DIALOGIC} uses as input a few in-context dialogue examples, and a simulator to build TOD data using in-context learning. The authors demonstrate that on MultiWOZ2.3, their approach leads to better performance than when training on human-generated dialogue. %The setup of Dialogic requires two inputs: a small dataset that contains a few annotated dialogues, and an ontology from which the user goal is extracted. Given the ontology, a model generates $G_i$ which contains both the user goal and belief state, represented by a set of triplets $\{domain, slot\_name, slot\_vlaue\}$. The authors sample dialogues from the dataset that have a similar user goal as $G_i$. To sample relevant dialogue examples from $D_S$, the authors maximize the similarity between goals $G_i$ and $G_j$ for all examples $j$ in $D_S$. This is done by calculating the Jaccard similarities of two dialogues on domain, and slot levels, similarly to ICL-US. 
Dialogic applies a Controllable Dialogue Generation step: This is an auxiliary step to verify the reliability of the GPT-3 generated dialogue. Once prompted, GPT-3 will generate a belief state and user utterance. If the belief state contains something that the dialogue turn generated utterance does not, it is called it is called over-generation. If the belief state lacks something in the utterance, it is called de-generation. 
%For tackling de-generation, the authors apply an auxiliary TOD model. Overall, the two methods attack the TOD generation problem in a very similar way. The only obvious differences being sampling the user goal directly from the dataset, or using an ontology to extract it.
INA ~\cite{ahmad2023ina} leverages GPT-2 for in-context generation of dialogues, using as examples the IND dataset generated in the same paper. This is achieved by fine-tuning GPT-2 in a supervised fashion with a cross-entropy loss and with a novel RL reward function using the PPO loss. More details about both IND and INA are presented in Section \ref{sec:conditioned_on_input}.

\note{\paragraph{Intent generation} There is a line of work that specializes in data generation for intents in a task-oriented setting. For example, \citet{fang2023chatgptdataaugmentationcompositional} study the challenge of identifying unseen intents that are compositions of previously observed ones, and uses ChatGPT as a data augmentation method to enhance generalizability. Along similar lines, \citet{dai2023auggptleveragingchatgpttext} and \citet{yu2023largelanguagemodelattributed} explore augmentation strategies for rephrasing around intents and expanding domain coverage. Building on this idea, \citet{lin2024generaterefinedataaugmentation} proposes a two-stage pipeline for zero-shot intent detection: utterances are first generated with an open-source LLM, and then refined by a smaller sequence-to-sequence model fine-tuned on seen domains. Their “generate-then-refine” approach improves both the diversity and utility of synthetic data, outperforming zero-shot generation alone. Meanwhile, \citet{sauer-etal-2022-knowledge} tackles the problem of limited labeled data by combining few-shot learning with knowledge distillation for intent classification, showing that compact models can still achieve competitive performance in data-scarce scenarios.}

\subsection{Quality Filtering}\label{sec:quality_check}

%     %%% Step 3:
In this subsection, we explore quality filtering with an emphasis on checking the accuracy of the generated data. \note{Similar to ODD and CIS dialog types, quality filtering focuses on lexical and consistency checks. However, TOD has an extra challenge in filtering dialogs for the quality of the generated intents, entities, and entity attributes. }Methods employing a schema, ontology, or knowledge graph in the generation process inherently ensure the accuracy of entities and entity attributes, as they are directly extracted from the input representations. This principle holds for the majority of papers discussed in this study, including ABUS, M2M, SGD, NUS, NeuralWOZ, HUS, VHUS, JOUST and Unified-US ~\cite{li2017ABUS, Shah2018M2M, Rastogi2020SGD, kreyssig2018neural, kim2021neuralwoz, gur2018HUS, tseng2021DSUSRL, wan2022unified}. We consider these to have factuality granted. On the other hand, methods that discover slots and values in the latent space do not have factuality covered. Dialogic~\cite{li2022DIALOGIC} controls the generation through an extra step called automatic revision, correcting for potential annotation errors by comparing the GPT-3 generated belief state with the current utterance, and avoiding any de-generation or over-generation issues. ICL-US~\cite{Terragni2023ICLUS} adds an evaluation step by comparing the dialogue acts extracted from the generated System and User NLU components at each turn. IND and INA~\cite{ahmad2023ina} apply a human-in-the-loop component where experts make edits at the utterance level of the generated dialogue to ensure the information is grounded in the provided background database, while also correcting minor grammatical errors. Meanwhile, VHDA ~\cite{yoo2020VHDA} ensures the semantic logic of the dialogue turns but does not apply quality checks for filtering generations.

\section{Open Domain Conversational Data Generation}~\label{sec:odd}

% === Introduction: 
\note{Open-Domain Dialogues (ODD) represent a specific type of conversational data that contains conversations across a wide variety of topics without being confined to specific tasks or domains~\cite{Ni2023Recent}.} When developing a synthetic dataset for ODD, it is essential to incorporate key features of ODD interactions~\cite{Ding23UltraChat}. Firstly, conversational coherence must be ensured, meaning each turn in the conversation should meaningfully connect to previous ones~\cite{Ni2023Recent}. Secondly, the dataset should promote response diversity, avoiding bland and repetitive replies like "I don't know" to foster more engaging interactions~\cite{zhao2019LaRL}. Additionally, the dataset should encompass a broad spectrum of topics, enhancing the system's generality. Lastly, it is crucial to generate conversations that aim to elicit informative responses, ensuring that conversations are both knowledgeable and relevant. For example, in knowledge-grounded dialogue systems, responses should draw on external knowledge bases, while in personalized dialogue systems, they should align with user profiles~\cite{gu2021chaincqg, do2022cohsCQG, HwangL22CQAGAR, hwang2022multiCQG}.

% \begin{figure}[t]
%     \centering
%     \includegraphics[width=0.98\textwidth]{figs/ODD/odd_example.pdf}
%     \label{fig:odd_example}
% \end{figure}

% == Our framework
This section reviews existing work on generating conversational data for ODD systems and systematically outlines the generation process in three components: input generation, utterance generation, and quality filtering.
% Step 1: 
The first component, \emph{input generation}, is tasked with providing the fundamental information necessary to initiate conversations. It processes a set of unstructured data and produces structured information cards, each serving as the basis for a conversation. In the literature on data generation, these information cards are referred to as ``conversation seeds.'' For instance, when creating a conversational dataset from a knowledge graph (KG)~\cite{kim2022soda}, the input generation component ingests a large KG, samples a triple, and through several steps, transforms it into a concise description that forms the foundation of a sample conversation. 
% Step 2: 
The second component, \emph{utterance generation}, utilizes the conversation seed to generate a multi-turn conversation. For example, using the description obtained from a KG triplet, this component instructs an LLM to generate a multi-turn conversation.
% Step 3: 
The third component, \emph{quality filtering}, aims to eliminate samples that do not meet the required quality standards and could potentially degrade the performance of the conversational system trained on this dataset. For instance, if conversation seeds or the resulting dialogue samples fail to meet quality criteria, they are discarded using heuristic or automated tools. 

\begin{figure}[t]
    \centering
    \includegraphics[width=0.93\textwidth]{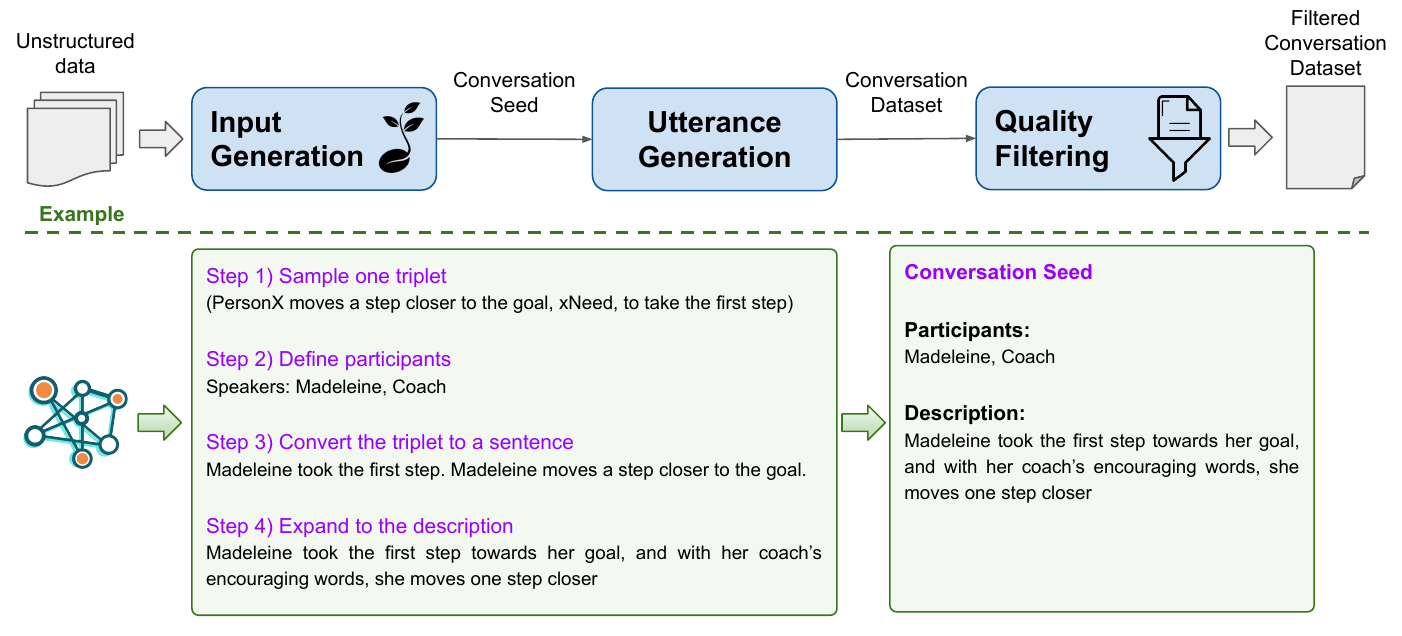}
    % \shrink
    \caption{Overview of data generation in ODD. In this example, adapted from SODA~\cite{kim2022soda}, knowledge graph triplets are used to construct a short description that serves as a conversation seed. An LLM is then prompted to generate a conversation based on this description.}
    \label{fig:odd_example}
    % \shrink
\end{figure}

This section provides a detailed explanation of each component and discusses the methods associated with them.
Figure~\ref{fig:odd_all} summarizes the methods employed in the literature for the aforementioned components, and Figure~\ref{fig:odd_example} illustrates the complete generation process with an example.

% \begin{figure}[t]
%     \centering
%     \includegraphics[width=0.94\textwidth]{figs/ODD/odd_all.png}
%     \caption{ODD all}
%     \label{fig:odd_all}
% \end{figure}

\begin{figure}[t]
\centering % Centers the figure
\begin{tikzpicture}[node distance=0.75cm]

% === Main Nodes (layer 1) ==
\node (odd_root) [root] {Open Domain Conversation Generation};
\node (2_utr) [odd_layer1_2, right=-0.25cm and 0.5cm of odd_root] {(2) Utterance Generation};
\node (1_inp) [odd_layer1_1, above=1.3cm of 2_utr] {(1) Input Generation};
\node (3_flt) [odd_layer1_3, below=1.0cm of 2_utr] {(3) Quality Filtering};

% === Layer 2 ===============
\node (1_2_tkg) [odd_layer2_1, right=0.5cm of 1_inp] {Triplet from KG};
\node (1_1_dia) [odd_layer2_1, above=0.4cm of 1_2_tkg] {Existing Dialogues};
\node (1_3_prf) [odd_layer2_1, below=0.4cm of 1_2_tkg] {Personalized Profile};
\node (2_1_1go) [odd_layer2_2, above right=-0.25cm and 0.5cm of 2_utr] {One GO};
\node (2_2_tbt) [odd_layer2_2, below right=-0.25cm and 0.5cm of 2_utr] {Turn by Turn};
\node (3_1_nos) [odd_layer2_3, above right=-0.25cm and 0.5cm of 3_flt] {Noise \& Lexical Filtering};
\node (3_2_con) [odd_layer2_3, below right=-0.25cm and 0.5cm of 3_flt] {Consistency Filtering};
% \node (3_3_per) [odd_layer2_3, below=0.4cm of 3_2_con] {Persona Filtering};

% === Layer 3 ===============
\node (1_1_1_ppr) [odd_layer3_1, right=0.5cm of 1_1_dia] {WEAKDAP$^{\large *}$~\cite{chen2022weakly}, AUGESC$^{\large *}$~\cite{Zheng23AugESC}, GCN~\cite{Lin22GCN}, BOTSTALK~\cite{kim2022botstalk}, PLACES$^{\large *}$~\cite{chen2023places}};
\node (1_2_1_ppr) [odd_layer3_1, right=0.5cm of 1_2_tkg] {SODA$^{\large *}$~\cite{kim2022soda}};
\node (1_3_1_ppr) [odd_layer3_1, right=0.5cm of 1_3_prf] {PERSONACHATGEN$^{\large *}$~\cite{lee2022personachatgen}, SPC$^{\large *}$~\cite{Jandaghi23spc}};

\node (2_1_1_ppr) [odd_layer3_2, right=0.5cm of 2_1_1go] {PLACES$^{\large *}$~\cite{chen2023places}, SODA$^{\large *}$~\cite{kim2022soda}, AUGESC$^{\large *}$~\cite{Zheng23AugESC}, GCN~\cite{Lin22GCN}};
\node (2_2_1_ppr) [odd_layer3_2, right=0.5cm of 2_2_tbt] {PERSONACHATGEN$^{\large *}$~\cite{lee2022personachatgen}, BOTSTALK~\cite{kim2022botstalk}, WEAKDAP$^{\large *}$~\cite{chen2022weakly} SPC$^{\large *}$~\cite{Jandaghi23spc}};

\node (3_1_1_ppr) [odd_layer3_3, right=0.5cm of 3_1_nos] {SODA$^{\large *}$~\cite{kim2022soda}, AUGESC$^{\large *}$~\cite{Zheng23AugESC}, PERSONACHATGEN$^{\large *}$~\cite{lee2022personachatgen} };
\node (3_2_1_ppr) [odd_layer3_3, right=0.5cm of 3_2_con] {SPC$^{\large *}$~\cite{Jandaghi23spc}, PERSONACHATGEN$^{\large *}$~\cite{lee2022personachatgen}, BOTSTALK~\cite{kim2022botstalk}};
% \node (3_3_1_ppr) [odd_layer3_3, right=0.5cm of 3_3_per] {};

% % Arrows
\draw (odd_root.east) -- ++(5pt,0) |- (1_inp.west);
\draw (odd_root.east) -- ++(5pt,0) |- (2_utr.west);
\draw (odd_root.east) -- ++(5pt,0) |- (3_flt.west);

\draw (1_inp.east) -- ++(5pt,0) |- (1_1_dia.west);
\draw (1_inp.east) -- ++(5pt,0) |- (1_2_tkg.west);
\draw (1_inp.east) -- ++(5pt,0) |- (1_3_prf.west);
\draw (2_utr.east) -- ++(5pt,0) |- (2_1_1go.west);
\draw (2_utr.east) -- ++(5pt,0) |- (2_2_tbt.west);
\draw (3_flt.east) -- ++(5pt,0) |- (3_1_nos.west);
\draw (3_flt.east) -- ++(5pt,0) |- (3_2_con.west);
% \draw (3_flt.east) -- ++(5pt,0) |- (3_3_per.west);

\draw (1_1_dia) -- (1_1_1_ppr);
\draw (1_2_tkg) -- (1_2_1_ppr);
\draw (1_3_prf) -- (1_3_1_ppr);
\draw (2_1_1go) -- (2_1_1_ppr);
\draw (2_2_tbt) -- (2_2_1_ppr);
\draw (3_1_nos) -- (3_1_1_ppr);
\draw (3_2_con) -- (3_2_1_ppr);
% \draw (3_3_per) -- (3_3_1_ppr);

\end{tikzpicture}
% \shrink
\caption{Overview of approaches used for Open-domain conversation generation. The conversation generation process is comprised of three main components, described in subsections~\ref{sec:odd:input}--\ref{sec:odd:filter}. \note{Methods using LLM are specified with ${\large *}$}.}
\label{fig:odd_all}
\shrink \miniskip
\end{figure}

\subsection{Input Generation}~\label{sec:odd:input}
Everyday dialogues usually revolve around specific topics, starting with a general question that gradually moves to more specific aspects of the topic~\cite{kim2022soda}. Therefore, to create a conversation sample, it is  necessary to first define an information card, referred to as "conversation seed," which specifies the main topic, subtopics, and key topic details for a conversation \note{(see example in the electronic appendix)}. 
% \todo{Figure~\ref{fig:odd_conv_seed_example} from the ESConv dataset~\cite{Liu21ESConv} shows a conversation seed next to the corresponding conversation.} 
Thus, the input generation component processes the data source with the goal of producing conversation seeds. The differences among input generation methods stem from the types of input sources used, the techniques employed in the generation process, and the specific content of each conversation seed. The subsequent sections will discuss various methods for generating conversation seeds.

% Existing Dialogue Datasets: WEAKDAP, AUGESC, GCN / BOTSTALK / PLACES
\subsubsection{Existing Dialogues}
% While human-written seeds~\cite{chen2023places} are of high quality, they also inherit the limitations of crowdsourcing methods, such as being time-consuming and labor-intensive. To address the human-written approach's limitations, existing conversational datasets can be employed. 
% In this approach, information from existing crowdsourced conversation datasets is utilized to generate conversation seeds. 
% Crowd-sourcing datasets typically contain background information for each conversation, which can effectively serve as conversation seeds. Thus, the process involves selecting an appropriate crowdsourced dataset tailored to the specific final task, eliminating the need for specialized generation techniques. Consequently, conversation seeds in this method are primarily composed of task descriptions, background knowledge, and one or two initial turns of the conversation.
In this method, information from existing crowdsourced conversation datasets is used to generate conversation seeds. These datasets typically include background information for each conversation, which serves as an effective starting point for creating new synthetic conversations. The process involves selecting a crowdsourced dataset that is well-suited to the specific end task. As a result, conversation seeds in this approach primarily consist of task descriptions, background knowledge, and the initial one or two turns of the conversation.

% Some works use this approach tailored to specific goals. 
In this realm, WEAKDAP~\cite{chen2022weakly} focuses on classifying emotions and conversational actions and detecting intentions in Spanish, and it utilizes the DailyDialog~\cite{li2017dailydialog} and FBTOD~\cite{Schuster19Cross} datasets for building it's conversation seeds. AUGESC~\cite{Zheng23AugESC} aims to generate conversations for Emotional Support Conversation (ESC) task. It employs the EmpatheticDialogues (ED)~\cite{rashkin2019towardsEmpathetic} dataset, which offers a rich variety of dialogue descriptions tagged with emotions and detailed descriptions of emotional states. GCN~\cite{Lin22GCN} is dedicated to creating knowledge-based conversations and uses the TopicalChat dataset~\cite{Karthik19Topical}, where each conversation is linked to relevant facts or articles found on Wikipedia, Reddit, or in The Washington Post. 
% \todo{in a knowedge graph?}.
BOTSTALK~\cite{kim2022botstalk}, designed for multi-skill conversational systems, integrates multiple behaviors and skills into a single conversation to enable appropriate interactions with diverse users and situations. BOTSTALK selects ConvAI2~\cite{Dinan2019ConvAI2}, WoW (Wizard of Wikipedia)~\cite{Dinan19wow}, and ED~\cite{rashkin2019towardsEmpathetic} datasets and combines them, using the conversation description and the first turn of the conversation sample as seeds.

Unlike previous approaches that use utterances from existing conversational datasets, PLACES~\cite{chen2023places} creates background information and conversations using human workers. Initially, it compiles a list of topics and tasks from the Feedback for Interactive Talk \& Search  (FITS) Dataset~\cite{Xu23Learning}, which includes 52 conversational topics (e.g., "nutrition," "philosophy") and 315 subtopics (e.g., "Italian food," "Soren Kierkegaard"). For each subtopic, PLACES manually generates background information. Additionally, it constructs a series of ten dialogues between two speakers for selected topics, illustrating everyday conversations with proper grammar. Consequently, the conversation seed in PLACES comprises the topic, background information, and a selection of sample conversations, all produced by human effort.

\if 0
\begin{figure}[t]
    \centering
    \includegraphics[width=0.94\textwidth]{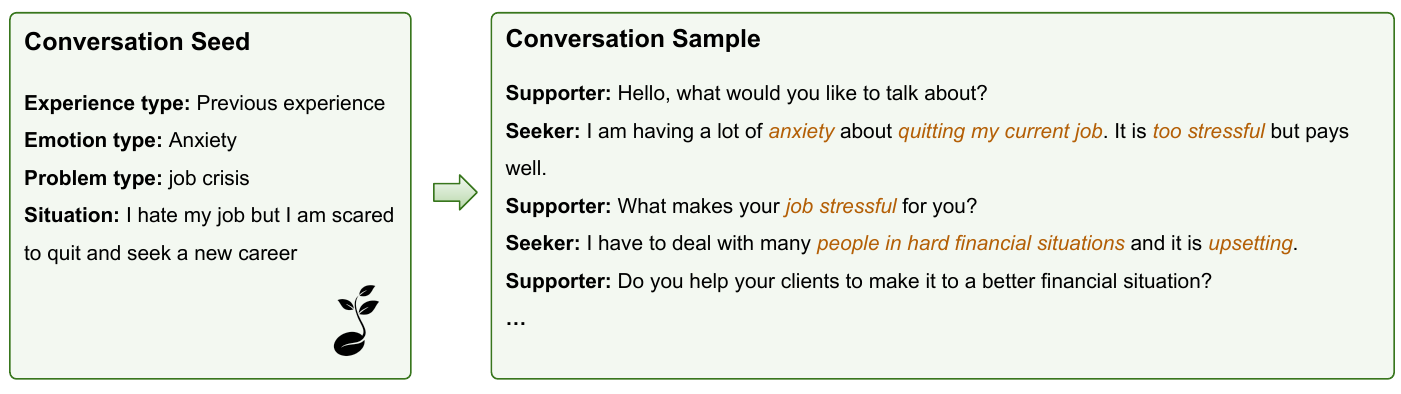}
    % \shrink
    \caption{Conversation seed and the corresponding dialogue from the ESConv Dataset~\cite{Liu21ESConv}. This dataset focuses on emotional support conversations. The example  shows a conversation seed that outlines the emotional problem and current situation of a support seeker. The corresponding dialogue develops around the information provided in the seed.}
    \label{fig:odd_conv_seed_example}
    % \shrink
\end{figure}
\fi

% ==== Using other resources like KG
\subsubsection{Triplet from KG}
In addition to existing conversational datasets, other types of resources, such as knowledge graphs (KGs), can be used to construct conversation seeds. Each triplet in a KG comprises two entities, referred to as the head and the tail, connected by a relation. The motivation for using textual resources beyond conversational datasets stems from the scarcity of conversational material in some domains. This makes it beneficial to leverage other textual resources, like KGs, for generating conversations. This approach effectively enhances the use of available textual resources in creating conversation datasets.
SODA~\cite{kim2022soda} utilizes KG triplets~\cite{west22atomic} to generate a concise description. The process involves several steps. Initially, SODA samples a socially relevant triplet, since the goal is to generate social conversations. It then transforms the triplet into a simple sentence using predefined templates tailored to each type of relation. To enhance the naturalness of the sentences, SODA substitutes variables representing individuals with the most common names from U.S. Social Security Number applicants.\footnote{\url{catalog.data.gov/dataset/baby-names-fromsocial-security-card-applications-national-data}}. Lastly, using GPT-3.5~\cite{Ouyang22gpt35}, SODA expands this sentence into a short description of two or three sentences. The description then serves as the conversation seed.

\subsubsection{Personalized Profile} 
Personalized chat systems are a distinct type of chat system that tailors responses based on a user profile (UP). Each UP consists of multiple profile sentences (PSs) that contain personalized information about the user. These PSs enable a dialogue system to generate personalized responses.
To generate conversational data for such systems, the fundamental approach involves developing UPs for two agents and then generating conversations based on these profiles. As a result, the conversation seed primarily consists of UPs. The process of generating these seeds begins with a list of topics and unfolds in two main steps. The first step is the preparation or creation of a set of PSs. The second step involves grouping several PSs together to form a complete UP. Ultimately, these user profiles comprise the conversation seed and serve as the foundation for generating personalized conversations.

% PERSONACHATGEN~\cite{lee2022personachatgen} generates conversational data for personalized systems. Initially, it defines a taxonomy of hierarchical persona topics. Following this, they use GPT-3 to generate PSs based on the defined topics. Once a set of PSs is established, PERSONACHATGEN groups them by calculating a contradiction score to select relevant PSs, and creates UPs. The process includes quality assurance measures for the UPs created. Firstly, heuristic filters are applied to remove samples that do not adhere to the key-value pattern and eliminate duplicates. Secondly, PERSONACHATGEN employs automatic scoring to assess the compatibility between the topic and the PS or among PSs within a UP.
% Synthetic-Persona-Chat (SPC)~\cite{Jandaghi23spc} starts by obtaining PSs from the PersonaChat dataset~\cite{zhang2018personachat}. It then prompts an LLM to generate additional PSs. Subsequently, SPC organizes these PSs using an NLI classifier to create the UP, ensuring that the profiles are coherent and contextually relevant.
% 
\shorten{PERSONACHATGEN~\cite{lee2022personachatgen} generates conversational data for personalized systems by first defining a hierarchical taxonomy of persona topics and using GPT-3 to generate PSs. These PSs are grouped into UPs using contradiction scores, followed by quality control via heuristic filtering and automatic compatibility scoring. Synthetic-Persona-Chat (SPC)~\cite{Jandaghi23spc} starts from PSs in the PersonaChat dataset~\cite{zhang2018personachat}, expands them using an LLM, and organizes the resulting PSs into coherent UPs with an NLI classifier.}

\subsection{Utterance Generation}\label{sec:gen}

\if 0
\begin{figure}[t]
    \centering
    \includegraphics[width=0.94\textwidth]{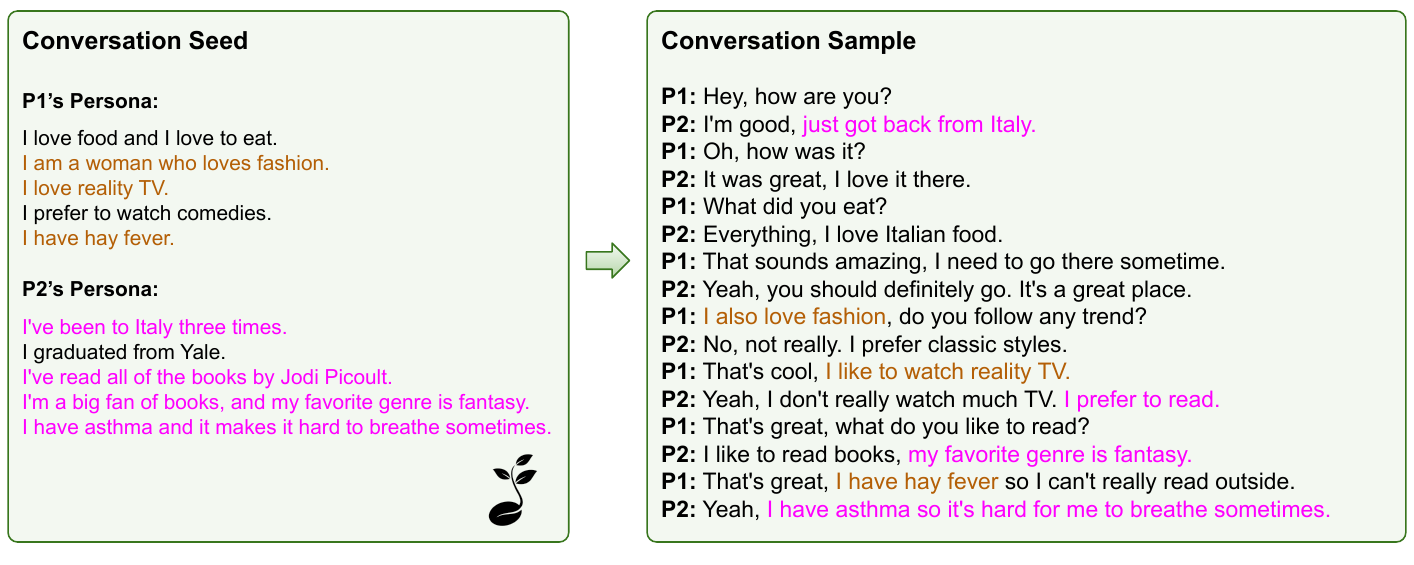}
    % \shrink
    \caption{Examples of generated dialogue from PersonaChatGen~\cite{lee2022personachatgen} based on a conversation seed containing two user profiles. Utterances in \textcolor{orange}{orange} are directly related to Person 1, while those in \textcolor{magenta}{magenta} relate to Person 2.}
    \label{fig:odd_utr_example}
    % \shrink \miniskip
\end{figure}
\fi

% === Intro
The primary goal of this step is to transform a conversation seed into a complete conversation. Thus, the input for this component consists of conversation seeds generated by the previous component, and its output is synthetically generated conversations. 
% \todo{Figure~\ref{fig:odd_utr_example} illustrates an example input and output of this component.}
% 
For ODD systems, the primary method for generating a conversation involves prompting an LLM with a conversation seed to generate a multi-turn conversation at once. This approach is referred to as \emph{One Go} in this paper. Some studies, however, employ a \emph{Turn by Turn} generation approach, where the conversation is constructed sequentially, one turn at a time. We will discuss the reasons why the turn-by-turn method is used.

\subsubsection{One GO}
This method generates a multi-turn conversation from a conversation seed, primarily by prompting an LLM. There distinct methods are used to generate conversation in one go.
The first method uses In-context Learning (ICL), where a pre-trained LLM is directly prompted without any fine-tuning~\cite{chen2023places, kim2022soda}. PLACES~\cite{chen2023places} generates conversations by using a prompt that initially includes a topic along with its human-written background information and a few randomly selected conversation samples as few-shot examples. Subsequently, another topic along with its corresponding background information is added to the prompt, and the LLM is tasked with generating a conversation for it. SODA~\cite{kim2022soda} employs GPT-3.5 to create multi-turn conversations from descriptions generated based on a triplet from a KG.

In the second method, an LLM is first fine-tuned and then used to generate conversations.
For instance, AUGESC~\cite{Zheng23AugESC} initially fine-tuned GPT-J~\cite{Wang21gptj} on a dialogue completion task using 100 samples from the ESConv dataset~\cite{Liu21ESConv} in one epoch. After fine-tuning, a conversation description along with the first utterance is provided to the LLM, which is then tasked with generating the remainder of the conversation.

The third method employs a Generator-Critic Architecture~\cite{Madaan23Self} and uses a combination of prompting and fine-tuning iteratively for generation. In this method, an LLM first generates multiple conversation candidates. Then, a Critic, which is another LLM, evaluates these candidates based on predetermined policies and selects the best ones. These top conversations are then incorporated into the dataset for the next generation cycle. This iterative process of selecting and adding the best candidates gradually enhances the Generator's performance.
Similarly, SPC~\cite{Jandaghi23spc} utilizes this approach for personalized conversation generation, prompting the LLM with a pair of user profiles.

\subsubsection{Turn By Turn}
Some studies generate conversations incrementally, often involving the interaction of two or more LLMs. There are several reasons for using the turn-by-turn method.
The first reason is to generate a personalized conversation based on two user profiles. In this scenario, two LLMs, each equipped with a user profile, engage in a dialogue with each other, generating the conversation turn by turn. PERSONACHATGEN~\cite{lee2022personachatgen} utilizes this approach by employing two GPT-3 models prompted by user profiles.
The second reason involves merging multiple datasets. In this approach, the choice of which dataset to use is determined turn by turn, based on the dialogue history. BOTSTALK~\cite{kim2022botstalk} uses this method as it merges three datasets to create a multi-skill conversational dataset.
The third reason is to increase the quantity and diversity of the data. By generating dialogues turn by turn, various scenarios can be created for generation. For example, WEAKDAP~\cite{chen2022weakly} defines three strategies for generation: 
First, Conversation Trajectory Augmentation (CTA) starts with the initial turn and substitutes the subsequent turns with generated ones. Second, All Turn Augmentation (ATA) generates a turn, considers the current subset as a conversation sample, then retains the original turn and generates the next. This results in \(n-1\) new conversations of varying lengths, from 2 to \(n\), for a conversation with \(n\) turns. Third, Last Turn Augmentation (LTA), a specific instance of ATA, focuses on replacing only the last turn of a conversation with a generated utterance.

\subsection{Quality Filtering}
\label{sec:odd:filter}

The primary goal of the filtering component is to eliminate conversations that lack key ODD features such as correctness, consistency, diversity, and informativeness. The existing filtering methods can be broadly categorized into two types.
The first type, Noise \& Lexical Filtering, primarily focuses on assessing correctness and diversity.
In this type, the filter removes samples that exhibit various issues, such as unfinished conversations~\cite{Zheng23AugESC}, deviations from desired patterns~\cite{kim2022soda, lee2022personachatgen}, repetitive patterns~\cite{lee2022personachatgen}, inappropriate turn lengths (either too short or too long)~\cite{Zheng23AugESC}, conversations with more than two participants~\cite{kim2022soda}, dangerous content~\cite{lee2022personachatgen}, toxic content with social biases, and offensive material~\cite{kim2022soda, lee2022personachatgen}.
In the second type of filtering, the focus is on checking the consistency of conversations. BOTSTALK~\cite{kim2022botstalk} ensures dialogue consistency by examining each newly added turn to verify that it aligns with the rest of the conversation. Similarly, in personalized-based models~\cite{lee2022personachatgen, Jandaghi23spc}, the consistency between persona sentences in one user profile is also assessed. To achieve this, a Natural Language Inference (NLI) classifier is primarily utilized.

\section{Information Seeking Conversational Data Generation}
\label{sec:cis}

% === Info-seeking intro ==========
% Main goal of CIS
\note{Conversational Information Seeking (CIS) conversations are multi-turn dialogues between one or more users and an information system, conducted primarily in natural language, whose goal is to resolve evolving information needs~\cite{Zamani2022CIS}.}
% The main goal of Conversational Information Seeking (CIS) systems is to fulfill users' information needs. These systems allow users to search information using natural language dialogue, instead of traditional search queries~\cite{Zamani2022CIS}.
% Features and challenges
However, CIS systems face several challenges in accurately addressing a user's request. 
The first challenge to address is ensuring the system maintains control of the conversation, allowing it to progress coherently until it meets the user's needs, while also offering users the flexibility to specify what they want to find or how they want the information to be presented~\cite{Zamani2022CIS}. The challenge intensifies when the user's queries are about various subjects or responses need to be drawn from different resources during the conversation. In such scenarios, the conversation may experience topic shifts, necessitating the system's ability to recognize when the topic changes and accordingly switch the response source~\cite{adlakha2022topiocqa}. Additionally, users may pose complex questions that require multi-hop reasoning across multiple sources to formulate a response. This demands more sophisticated retrieval models and reasoning capabilities, as the system needs to synthesize information from diverse resources~\cite{Zhang23Beam, Li24MultiDoc}.
Another major challenge is query ambiguity, where the system may not understand the user's request or might have multiple potential responses but cannot determine which one is most relevant. In such cases, the system need to ask clarifying questions. This interaction, where both the user and the system can initiate queries and drive the conversation, is known as a mixed-initiative system~\cite{Wu22INSCIT, Deng2023ProCoT}. 
To effectively develop a CIS system capable of handling these challenges, it is essential to generate training data that encapsulates these scenarios.

\begin{figure}[t]
    \centering
    \includegraphics[width=0.95\textwidth]{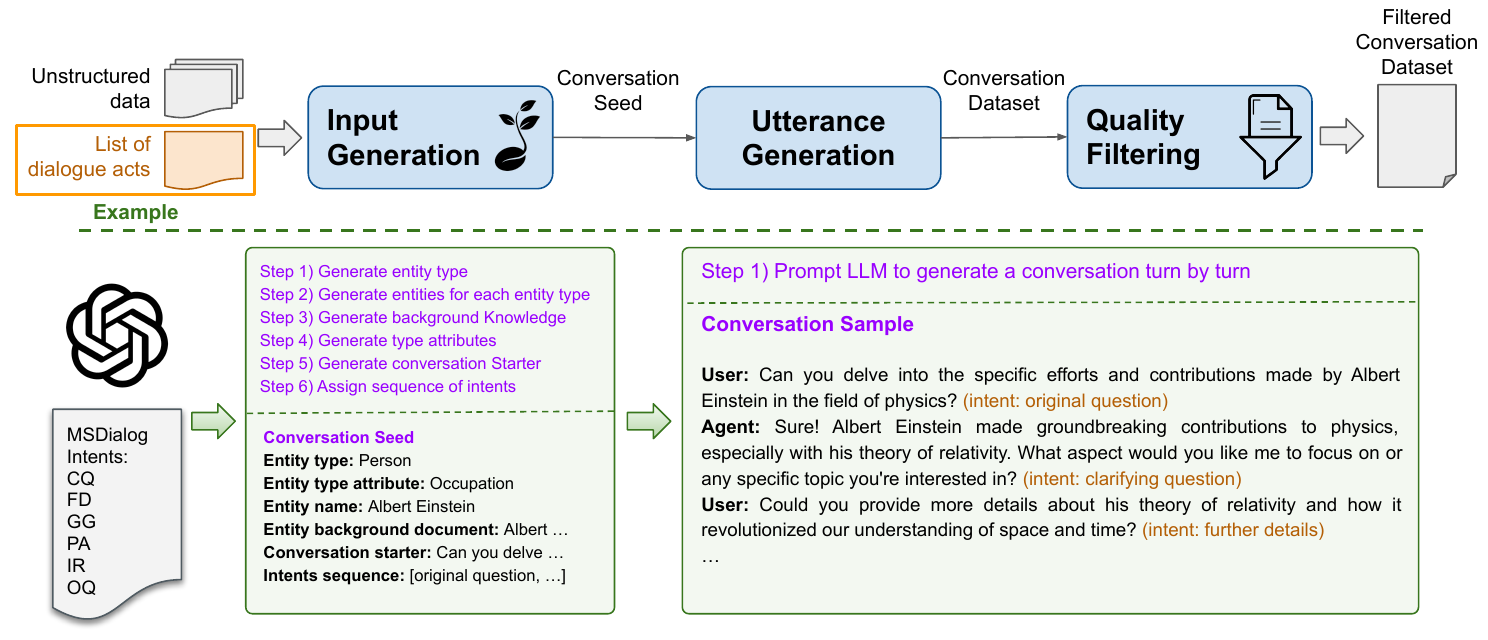}
    % \shrink 
    \caption{Overview of data generation in CIS. 
    % This process comprises three main components. Inputs and outputs for each component are depicted, alongside an example from SOLID~\cite{askari2024self}. 
    In this example from SOLID~\cite{askari2024self}, a list of intents from the MSDialog dataset~\cite{qu2018msdial} is fed into the input generation component. This component sequentially prompts an LLM to generate the necessary background information and selects a random sequence of intents. After constructing the conversation seed with the background information and intent sequence, it prompts an LLM to generate intent-aware conversation turns sequentially.}
    \label{fig:cis_example}
    % \shrink \miniskip
\end{figure}

% Our contributions
This section reviews existing work on generating conversational data for CIS systems and systematically describes the generation process through three main components: input generation, utterance generation, and quality filtering; see Figure~\ref{fig:cis_example}.
The first component, \emph{input generation}, is responsible for providing the fundamental information needed to initiate conversations; e.g., topic and evidence document. It also defines a conversation flow, which is used to guide the conversation to address users' information needs.
%the role of each turn in a conversation. This sequence controls the dialogue flow to address the user's information needs. 
Both elements, fundamental information and conversation flow, constitute what is referred to as a "Conversation Seed." For example, a conversation seed may include an entity name with relevant background information, followed by a sequence of intents like [original question, clarifying question, further details].
The second component, \emph{utterance generation}, aims to generate a conversation based on the "Conversation Seed."
The third component, \emph{quality filtering}, seeks to eliminate samples that do not meet the required quality standards, which could potentially degrade the performance of the conversational system trained on this dataset. This section provides a detailed explanation of each component and discusses the methods associated with them. The methods employed for these components are illustrated in Figure~\ref{fig:cis_all}.

\subsection{Input Generation}
\label{sec:cis:input}

% they have structure 
CIS conversations are designed to respond to queries or gather information about a topic, typically following a structured sequence. In the literature on conversation generation, it is generally assumed that an information-seeking conversation starts with the user's query, progresses through a series of interactions between the agent and the user, and concludes with the delivery of the final answer.
% named dialogue flow
Therefore, each turn in these conversations can be labeled with an act name, resulting in a multi-turn conversation being categorized by a sequence of dialogue acts, commonly referred to as "dialogue flow." In other words, the dialogue flow offers a comprehensive perspective of the conversation, outlining the role of each turn and how it contributes to reaching the final answer.
% Seed included, with an example
Thus, a dialogue flow, coupled with information about the topic of conversation and corresponding background information, forms the "Conversation Seed" for CIS systems \note{(see example in the electronic appendix)}. 
% \todo{Figure~\ref{fig:cis_seed_example} illustrates an example of a conversation seed alongside its conversation sample from the Doc2Dial dataset~\cite{Feng2020Doc2Dial}.}
%The conversation seed includes a title, a relevant document, and a sequence of dialogue acts.
The conversation sample is generated based on the title and document and is structured around the dialogue flow to meet the user's information needs. In the following, we discuss  existing approaches for input generation and categorize them into two main categories of document-driven and sequence-grounded.%, based on input data, generation techniques, and the content of the seed as the output.

\begin{figure}[t]
\centering % Centers the figure
\begin{tikzpicture}
% [node distance=0.5cm]

% Main Nodes
\node (convcis) [root] {Conversation Info. Seeking Generation};

\node (2_utr) [cis_layer1_2, right=1.4cm and 0.5cm of convcis] {(2) Utterance Generation};
\node (1_inp) [cis_layer1_1, above=1.7cm of 2_utr] {(1) Input Generation};
\node (3_flt) [cis_layer1_3, below=1.7cm of 2_utr] {(3) Quality Filtering};

\node (1_1_doc) [cis_layer2_1, above right=-0.35cm and 0.3cm of 1_inp] {Document-Driven};
\node (1_2_seq) [cis_layer2_1, below right=-0.35cm and 0.3cm of 1_inp] {Sequence-Grounded};

% \node (2_1_tbt) [cis_layer2_2, above right=0.2cm and 0.3cm of 2_utr] {Turn by Turn};
% \node (2_2_ogo) [cis_layer2_2, below right=0.2cm and 0.3cm of 2_utr] {One Go};

\node (2_2_inp) [cis_layer2_2, right=0.4cm and 0.3cm of 2_utr] {Inpainting};
\node (2_1_pip) [cis_layer2_2, above=0.4cm and 0.3cm of 2_2_inp] {Answer Extraction \\ \& Question Generation};
\node (2_3_sim) [cis_layer2_2, below=0.4cm and 0.3cm of 2_2_inp] {LLM guided/simulated};

\node (3_1_nos) [cis_layer2_3, above right=-0.35cm and 0.3cm of 3_flt] {Noise \& Lexical};
\node (3_2_fac) [cis_layer2_3, below right=-0.35cm and 0.3cm of 3_flt] {Factuality Check};

% \node (2_1_2_inp) [cis_layer3_2, right=0.4cm and 0.3cm of 2_1_tbt] {Inpainting};
% \node (2_1_1_pip) [cis_layer3_2, above=0.4cm and 0.3cm of 2_1_2_inp] {Answer Extraction \\ \& Question Generation};
% \node (2_1_3_sim) [cis_layer3_2, below=0.4cm and 0.3cm of 2_1_2_inp] {User/Agent Simulation};

\node (1_1_1_1_ppr) [cis_layer4_1, right=0.2cm of 1_1_doc] {DG2~\cite{wu2022dg2}, SynDG~\cite{bao2023SynDG}, Inpainting~\cite{Dai22Inpainting}, MultiCQAG~\cite{hwang2022multiCQG}, CQAG-AR~\cite{HwangL22CQAGAR}, SimSeek~\cite{kim2022simseek}, CONVERSER$^{\large *}$~\cite{Huang23CONVERSER}, SimQuAC$^{\large *}$~\cite{abbasiantaeb2024let}};
\node (1_2_1_1_ppr) [cis_layer4_1, right=0.2cm of 1_2_seq] {ConvTrans~\cite{Mao22ConvTrans}, SOLID$^{\large *}$~\cite{askari2024self}, TopDial$^{\large *}$~\cite{Wang23topdial}, MusicSyn$^{\large *}$~\cite{leszczynski2022conversational}, TtWMusic$^{\large *}$~\cite{leszczynski2023talk}, %MATHDIAL~\cite{macina2023mathdial}, 
LAPS$^{\large *}$~\cite{Joko2024LAPS}, ConvSDG$^{\large *}$~\cite{Mo24ConvSDG}};

\node (2_1_1_1_ppr) [cis_layer4_2, right=0.2cm of 2_1_pip] {DG2~\cite{wu2022dg2}, SimSeek~\cite{kim2022simseek}, MultiCQAG~\cite{hwang2022multiCQG}, CQAG-AR~\cite{HwangL22CQAGAR}, CONVERSER$^{\large *}$~\cite{Huang23CONVERSER}};
\node (2_1_2_1_ppr) [cis_layer4_2, right=0.2cm of 2_2_inp] {Inpainting~\cite{Dai22Inpainting}, MusicSyn$^{\large *}$~\cite{leszczynski2022conversational}, TtWMusic$^{\large *}$~\cite{leszczynski2023talk}, SynDG~\cite{bao2023SynDG}};
\node (2_1_3_1_ppr) [cis_layer4_2, right=0.2cm of 2_3_sim] {ConvTrans~\cite{Mao22ConvTrans}, LAPS$^{\large *}$~\cite{Joko2024LAPS}, %MATHDIAL~\cite{macina2023mathdial}, 
TopDial$^{\large *}$~\cite{Wang23topdial}, SimQuAC$^{\large *}$~\cite{abbasiantaeb2024let}, SOLID$^{\large *}$~\cite{askari2024self}, ConvSDG$^{\large *}$~\cite{Mo24ConvSDG}};

\node (3_1_1_1_ppr) [cis_layer4_3, right=0.2cm of 3_1_nos] {TopDial$^{\large *}$~\cite{Wang23topdial}, %MATHDIAL~\cite{macina2023mathdial}, 
SOLID$^{\large *}$~\cite{askari2024self}};
\node (3_2_1_1_ppr) [cis_layer4_3, right=0.2cm of 3_2_fac] {DG2~\cite{wu2022dg2}, SimSeek~\cite{kim2022simseek}, MultiCQAG~\cite{hwang2022multiCQG}, CQAG-AR~\cite{HwangL22CQAGAR}, SimQuAC$^{\large *}$~\cite{abbasiantaeb2024let}, SynDG~\cite{bao2023SynDG}, CONVERSER$^{\large *}$~\cite{Huang23CONVERSER}};

\draw (convcis.east) -- ++(5pt,0) |- (1_inp.west);
\draw (convcis.east) -- ++(5pt,0) |- (2_utr.west);
\draw (convcis.east) -- ++(5pt,0) |- (3_flt.west);
\draw (1_inp.east) -- ++(5pt,0) |- (1_1_doc.west);
\draw (1_inp.east) -- ++(5pt,0) |- (1_2_seq.west);
\draw (2_utr.east) -- ++(5pt,0) |- (2_1_pip.west);
\draw (2_utr.east) -- ++(5pt,0) |- (2_2_inp.west);
\draw (2_utr.east) -- ++(5pt,0) |- (2_3_sim.west);
\draw (3_flt.east) -- ++(5pt,0) |- (3_1_nos.west);
\draw (3_flt.east) -- ++(5pt,0) |- (3_2_fac.west);

\draw (2_1_pip.east) -- ++(5pt,0) |- (2_1_1_1_ppr.west);
\draw (2_2_inp.east) -- ++(5pt,0) |- (2_1_2_1_ppr.west);
\draw (2_3_sim.east) -- ++(5pt,0) |- (2_1_3_1_ppr.west);
% \draw (2_2_ogo.east) -- ++(5pt,0) |- (2_2_1_1_ppr.west);

% \draw (1_1_doc.east) -- ++(5pt,0) |- (1_1_1_1_ppr.west);
% \draw (1_2_seq.east) -- ++(5pt,0) |- (1_2_1_1_ppr.west);
% \draw (2_1_1_pip.east) -- ++(5pt,0) |- (2_1_1_1_ppr.west);
% \draw (2_1_2_inp.east) -- ++(5pt,0) |- (2_1_2_1_ppr.west);
% \draw (2_1_3_sim.east) -- ++(5pt,0) |- (2_1_3_1_ppr.west);

\draw (3_1_nos.east) -- ++(5pt,0) |- (3_1_1_1_ppr.west);
\draw (3_2_fac.east) -- ++(5pt,0) |- (3_2_1_1_ppr.west);

\end{tikzpicture}
\caption{Overview of methods for information seeking conversational data generation. The conversation generation process is comprised of three main components, described in subsections~\ref{sec:cis:input}--\ref{sec:cis:filter}. \note{Methods using LLM are specificed with ${\large *}$}.}
\label{fig:cis_all}
\shrink
\end{figure}

\subsubsection{Document-driven}

\if 0
\begin{figure}[t]
    \centering
    \includegraphics[width=0.94\textwidth]{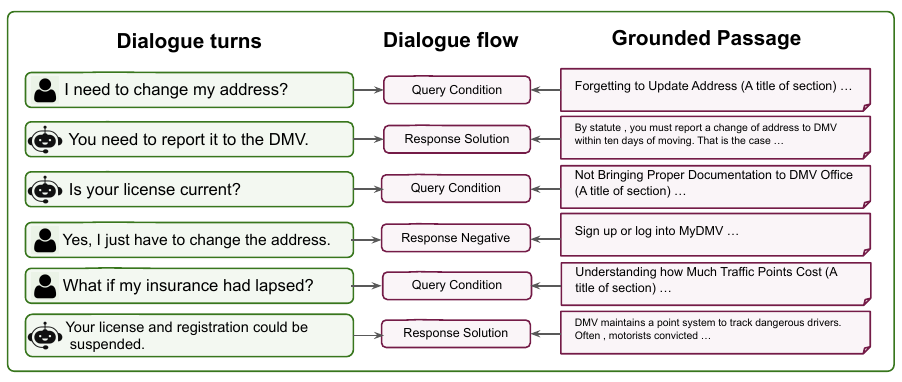}
    \shrink
    \caption{Conversation Seed and Corresponding Dialogue from the Doc2Dial Dataset~\cite{Feng2020Doc2Dial}. This dataset is dedicated to generating conversations based on governmental documents. The displayed example features a conversation seed comprising a  title, a document, and a sequence of dialogue acts. In the corresponding conversation sample, each turn is labeled with a specific dialogue act.}
    \label{fig:cis_seed_example}
    \shrink \miniskip
\end{figure}
\fi

% === Intro
High-quality documents, such as those found on platforms like Wikipedia or PubMed, are typically written or edited by experts, who invest significant effort in refining their content and addressing potential reader questions preemptively~\cite{Dai22Inpainting}. 
These documents often begin with broad overviews of general concepts and gradually delve into more detailed aspects of the subject~\cite{gao2019CFnet, gu2021chaincqg, do2022cohsCQG, wu2022dg2}. The sequential organization of documents coupled with the rich background information of documents serve as a valuable resource for modeling dialogue flow of information-seeking conversations. 
Therefore, in the document-driven approach,  the implicit flow of a document or a ranked list of document passages is used to guide the dialogue. 
%Consequently, a document itself can form a conversation seed, as it contains both background information and structure.
% 
In what follows, we discuss three methods of constructing a dialogue flow in a document-driven manner.

% Type 1) Passage retrieval: during the converation / or predefinded DG2, SynDG
% Type 2) Inpainting idea: Inpainting 
% Type 3) No flow: SimSeek, simQuAC

% === type 1) Passage retrieval
% The first method of document-driven input generation segments documents into passages and then selects a passage to generate a conversation turn, thereby creating a dialogue flow based on the sequence of passages. DG2~\cite{wu2022dg2} employs a turn by turn retrieval-based approach for dialogue flow generation. For each turn, it utilizes a "passage selector" component responsible for ranking these passages and selecting the most relevant one based on the conversation history to generate the current turn. 
% 
The first document-driven input generation method segments documents into passages and generates dialogue turns by selecting passages sequentially. DG2~\cite{wu2022dg2} follows a turn-by-turn retrieval-based approach, using a passage selector to rank passages based on the conversation history and choose the most relevant one for each turn.
%
%a In the first type, a document is segmented into multiple passages. Subsequently, a passage is selected to generate each turn of the conversation, thereby creating a dialogue flow consisting of a sequence of passages~\cite{wu2022dg2}. 
SynDG~\cite{bao2023SynDG} utilizes "knowledge pieces" instead of mere passages. A knowledge piece can be any textual segment; e.g., a sentence from a Wikipedia article or a persona sentence used in a personalized dialogues to describe a user's preferences. %The dialogue flow in SynDG is predefined, based on knowledge pieces driven from existing dialogues.
Segmenting a document can obscure its positional context, even though dialogues naturally shift toward later, more detailed parts of a document. To address this, DG2 incorporates both dialogue-turn and passage-position information. It marks speaker turns with prompts like “user{num}:” or “agent{num}:” to track dialogue progression, and assigns each passage an index indicating its location in the document. Combining these signals helps the model begin conversations near the document’s start and gradually move toward later sections as the dialogue unfolds.
% 
% Paper 2: SynDG
SynDG initially defines a sequence of knowledge pieces as a pre-defined dialogue flow. It applies this pre-defined dialogue flow in two instances: the first instance uses the persona sentences from the PersonaChat dataset~\cite{zhang2018personachat} as knowledge pieces, and the second instance utilizes passages from the WoW dataset~\cite{Dinan19wow}.

% === type 2) Inpainting
In the second method of docuemnt-driven input generation, documents are conceptualized as dialogues between the writer and an imaginary reader, where each sentence in the document is viewed as the writer’s response to a hypothetical question from the reader. Thus, in this approach, the dialogue flow consists of the document's sentences. This concept, called Inpainting, was first introduced by \citet{Dai22Inpainting}, and subsequently adopted in other studies~\cite{leszczynski2022conversational, leszczynski2023talk}.

% === type 3) No flow
As the third method, a specific flow for the conversation is not defined. Instead, a complete document of or background information is provided, and it is left to the "utterance generation" component to decide which part of the document to use for dialogue generation~\cite{kim2022simseek, abbasiantaeb2024let, hwang2022multiCQG, HwangL22CQAGAR}.

% \subsubsection{Intent Flow}
\subsubsection{Sequence-Grounded}
% Intro
In this approach, the conversation seed comprises a topic, its corresponding background information, and a sequence of dialogue acts~\cite{liu2021durecdial, Wang23topdial}, sometimes referred to as intents~\cite{qu2018msdial, askari2024self}. The conversation seed closely resembles that used in ODD but additionally includes the sequence of acts. For this setup, two types of input resources are necessary: a collection of topics with their background knowledge and a list of dialogue acts.
Furthermore, given that a sequence of dialogue acts is selected from a pool, verifying the validity of this sequence is crucial.
\shorten{TopDial~\cite{Wang23topdial} generates synthetic recommendation dialogues using a role-playing framework where two LLMs act as the system and the user. Each role is initialized with a conversation seed: the system seed includes background information, an initial utterance, dialogue acts, and a target from DuRecDial 2.0~\cite{liu2021durecdial}, while the user seed consists of profile attributes and personality features. User personalities are simulated using randomly assigned positive or negative descriptions of the Big-5 traits, openness, conscientiousness, extraversion, agreeableness, and neuroticism~\cite{goldberg1993structure}. Although only the final target is predefined, the system typically begins by asking preference-related questions and gradually introduces topic-specific attributes as the dialogue unfolds.}
% 
% SOLID~\cite{askari2024self} proposes a method for generating data for intent-aware information-seeking dialogues. It involves sequentially prompting an LLM to generate entity types and corresponding entities for each type, background documents, and  the first utterance in a conversation, which collectively form the necessary information for the conversation seed. For the dialogue flow, SOLID employs a list of intents from MSdialog~\cite{qu2018msdial}, adhering to the same intent distribution found in the main dataset.
% 
\shorten{SOLID~\cite{askari2024self} generates intent-aware information-seeking dialogues by sequentially prompting an LLM to produce entity types and entities, background documents, and an initial utterance, which together form the conversation seed. The dialogue flow follows intent sequences from MSdialog~\cite{qu2018msdial}, preserving the original intent distribution.}
% 
% Similarly, MusicSyn~\cite{leszczynski2022conversational} and TtWMusic~\cite{leszczynski2023talk} use  fixed sequences of dialogue acts. They are designed to generate information-seeking conversations for music recommendation systems and make use of the CPCD dataset~\cite{Beyond23Chaganty}, which contains background information on artist playlists. To create a sequence of playlists that serves as the dialogue flow, they compute the embedding vectors of these playlists. They then sample a sequence based on closeness, utilizing cosine distance to ensure consistency.
\shorten{MusicSyn~\cite{leszczynski2022conversational} and TtWMusic~\cite{leszczynski2023talk} generate music recommendation dialogues using fixed dialogue act sequences and the CPCD dataset~\cite{Beyond23Chaganty}. They construct dialogue flows by embedding artist playlists and sampling sequences based on cosine similarity to maintain topical consistency.}
\note{ConvTrans~\cite{Mao22ConvTrans} reorganizes raw web search sessions into a heterogeneous session graph, where nodes represent queries and edges represent relationships between queries. The graph includes three types of edges (response-induced, topic-shared, and topic-changed) capturing three common types of query relations in conversational search. ConvSDG~\cite{Mo24ConvSDG}, on the other hand, generates conversations by taking a topic, its corresponding description, and a partial conversation as input.}

% 2) Not pre-defined seq: [MATHDIAL][LAPS]
Some studies do not specify a fixed sequence; instead, outline a set of actions and select one before generating each turn. These studies focus on collaboration between LLM and humans, employing semi-synthetic methods for conversation generation. LAPS~\cite{Joko2024LAPS} uses a set of primary dialogue acts for personalized recommendation tasks, (1) greeting, (2) preference elicitation, (3) recommendation, (4) follow-up questions, and (5) goodbye. It then guides the conversation follow by generating a guidance for each turn, specifying the next action that should be taken in the conversation based on the conversation history and user preferences. This guidance is then utilized by a crowd worker to compose the next utterance for the conversation.
%MATHDIAL~\cite{macina2023mathdial} introduced a dataset specifically designed for dialogue-based tutoring systems concentrating on mathematics problems. In this system, the flow of the dialogue is not predetermined due to the interactive human element. However, before generating each turn, the human is expected to select a dialogue act that incorporates different teaching techniques. 
% that  turn turn-by-turn dialogue acts using LLMs to  create multi-session information-seeking conversations that integrate user preferences.

\subsection{Utterance Generation}
\label{sec:cis:gen}

The primary goal of this component is to transform a conversation seed into a multi-turn conversation. However, the content of the seed in CIS differs from that in ODD. The input for this component is the conversation seeds containing a topic and dialogue flow, and its output is a conversational dataset. In the following, we will explore the approaches used for utterance generation.
% 
% In CIS, the primary method for generating conversations is turn-by-turn, facilitated by the pre-existence of a sequence of dialogue acts. In this approach, each turn is generated based on the background information, the current dialogue act, and the conversation history. This is achieved either by using a fine-tuned LLM or by prompting an LLM. Additionally, one-go generation is also employed for specific purposes, which will be discussed in further detail.
% \subsubsection{Turn by Turn}
% Most works generate conversations turn by turn, and a limited number use one-go generation. We categorize the turn by turn-based works into three types.

\subsubsection{Answer Extraction and Question Generation}
% 1) Answer Extraction & Question Generation [DG2][SimSeek][CQAG-AR][MultiCQAG]
This approach utilizes two sequential components, known as Answer Extraction and Question Generation, which are based on the pipeline method for question-answering (QA) generation introduced by \citet{alberti2019synthetic}. Typically, the first component extracts a text span that could potentially serve as an answer, and the second component generates a question corresponding to the extracted answer.
% [DG2][CQAG-AR][MultiCQAG]
DG2~\cite{wu2022dg2} retrieves a passage to use as a conversation seed for generating a turn. It then employs a span extraction component to identify a rationale that could potentially serve as an answer to a question. Following this, it sequentially generates user and agent utterances based on the extracted rationale and the conversation history. MultiCQAG~\cite{hwang2022multiCQG} and CQAG-AR~\cite{HwangL22CQAGAR} employ a similar approach, with the distinction that after extracting the span, they determine which type of question to generate: open-ended, closed-ended, or unanswerable.
% 
% [SimSeek]
SimSeek~\cite{kim2022simseek} explores two scenarios of generating synthetic conversations from documents. 
%the question of whether knowing an answer in advance can better encourage information-seeking behavior when asking questions. To address this, SimSeek establishes two scenarios. 
In the first scenario, called "SimSeek-sym," an evidence passage is provided from which an answer is extracted first, followed by the generation of a question based on the extracted answer. In the second scenario, "SimSeek-asym," only a topic and background information are supplied to the question generator component. Then an answer finder component attempts to identify the answer within the evidence passage. In this approach, the full Wikipedia passage is considered as the conversation seed.
\note{CONVERSER~\cite{Huang23CONVERSER} generates a sequence of queries within a session, grounded on one or more passages. It also incorporates passage switching, where the current passage is replaced with a related passage at each turn with a certain probability.}

% 2) Dialogue reconstruction / 
\subsubsection{Dialogue Inpainting}
The core concept of dialogue inpainting is to reconstruct missing parts of a dialogue~\cite{Dai22Inpainting}; e.g., given a conversation containing agent's responses, the inpainting method generates the corresponding user's questions. \citet{Dai22Inpainting} train a dialogue inpainting model by fine tuning an LLM on dialogues with randomly masked utterances.  The inpainter then learns to predict the missing utterances of a dialogue. Assuming that sentences of a document represent agent responses in an imaginary conversation, the trained inpaiter model then generates user questions for all system responses and transforms a document to a an information seeking conversation.
The concept of inpainting can be also seen in SynDG~\cite{bao2023SynDG}, where  a fine-tuned sequence-to-sequence converts knowledge pieces (e.g., Wikipedia passages or sentences from a user's profile) into consistent turns of a conversation.
Additionally, MusicSyn~\cite{leszczynski2022conversational} and TtWMusic~\cite{leszczynski2023talk} apply this technique to generate conversations for a music recommendation system. The system's responses consist of a list of playlists, and the inpainting model is tasked with generating the user's questions to complement these responses. 

% . Here, the seed may consist of a sequence of passages or knowledge pieces, with the reconstruction model aiming to transform these textual pieces into dialogue turns. Alternatively, the seed could be a sequence of system responses, and the model's goal is to generate a complete dialogue from a partial one.
% % 
% SynDG~\cite{bao2023SynDG} fine-tunes a pre-trained sequence-to-sequence model for a dialogue reconstruction task. Using this fine-tuned model, it converts knowledge pieces—which can be Wikipedia passages or sentences from a user's profile—into consistent turns of a conversation.
% % 
% Inpainting~\cite{Dai22Inpainting} introduces a technique designed to reconstruct missing parts of a dialogue. In this context, the conversation seed consists of the agent's responses, necessitating the generation of the user's questions. The method trains a dialogue inpainting model by providing it with a complete dialogue from which one utterance is randomly masked. The model then learns to predict the missing utterance. During inference, the model operates autoregressively: it starts by generating the first question of the conversation, replaces the first masked utterance with the generated question, and continues to generate subsequent questions in a similar fashion.
% 

% 3) User/Agent Simulation
\subsubsection{LLM guided/simulated}
% LAPS
% MathDial
% TopDial
% SimQuAC
% SOLID - SOLID-RL

The third category of CIS utterance generation methods employs LLMs to steer the conversation generation process either by guiding humans or by directly simulating user roles to generate utterances.  

% === [LAPS]: LLM-Guided Human Generation
In the first approach, the LLM generates guidance, and humans create conversation utterances based on this guidance. LAPS~\cite{Joko2024LAPS} utilizes this approach to generate large scale human-written conversational data for preference elicitation and personalized recommendation. It generates multi-session conversations using an LLM to dynamically provide personal guidance for crowd workers who play both roles of  user and agent. The personal guidance is generated dynamically per turn based on the conversation history and elicited  preferences of the crowd worker in previous conversation sessions. It enables crowd workers to overcome the high cognitive load of generating conversational utterances in complex and lengthy conversations and craft their own utterances through dialogue self-play. 
%engage in the task.uch as those in multi-session recommendation systems, crowd workers often exhibit low engagement due to the high cognitive load involved in generating conversational utterances. To mitigate this issue, LAPS proposes coaching crowd workers throughout the conversation with automatically generated personalized guidance, allowing them to craft their own utterances through dialogue self-play.
% 
LAPS consists of four key components:  dialogue act classification, guidance generation, utterance composition, and preference extraction. For dialogue act classification, LAPS utilizes an LLM to identify the next act that should be taken in the dialogue. The primary dialogue acts in this framework are: (1) greeting, (2) preference elicitation, (3) recommendation, (4) follow-up questions, and (5) goodbye. In the guidance generation component, it uses GPT-3.5-turbo to create personalized guidance, with the template for guidance prompts incorporating dialogue history, preference memory, and the identified dialogue act. 
The utterance composition is handled by a human agent who crafts the conversation utterances for both the user and the assistant, using dialogue self-play. After the completion of a dialogue session, the preference extraction process extracts user preferences from the dialogue and verify them by the same human agent who managed the conversation. 

% === [MATHDIAL]: LLM-Human collaboration
%In the second approach, an LLM takes on the role of one participant in a conversation.
%MATHDIAL~\cite{macina2023mathdial} seeks to create a tutor dialogue dataset through the integration of human and LLM interactions. In this work, the conversation seed contains a math problem accompanied by a list of actions that the teacher can perform. It uses an LLM for the student role and a human for the teacher role. Specifically, it employs InstructGPT~\cite{Ouyang22gpt35} to generate student turns, prompting the model with the previous dialogue history and grounding information for the upcoming turn. The prompts include the math word problem (MWP) based on the GSM8k collection~\cite{Cobbe21gsm}, the initial student confusion, and the student profile, which provides details about the nature of the confusion and the student’s persona. Based on the student’s response, the teacher selects one of the dialogue acts and continues the conversation by addressing the student’s confusion.

% === [TopDial][SimQuAC]
In the second approach, multiple LLMs serve as both participants of the conversation, with no human involvement.
% [TopDial]
% TopDial~\cite{Wang23topdial} proposes generating conversations for a target-guided recommendation system. It employs three LLMs: one representing the user, another for the system, and a third to manage and supervise the dialogue. The user's conversation seed includes their profile and personality, while the system's seed comprises domain knowledge, dialogue acts, and the final target. For these roles, three instances of ChatGPT (gpt-3.5-turbo version) are utilized.
% 
TopDial~\cite{Wang23topdial} generates conversations for target-guided recommendation by simulating user, system, and supervisor roles using three instances of ChatGPT. The user is seeded with a profile and personality, while the system is conditioned on domain knowledge, dialogue acts, and the target goal.
SimQuAC~\cite{abbasiantaeb2024let} developed to emulate the crowdsourcing methodology originally used for creating the QuAC dataset~\cite{choi2018quac}, replacing human participants with LLMs. It employs zero-shot learning LLMs to simulate teacher-student interactions. The dialogue revolves around a wikipedia topic; the student only has access to the topic and background information and asks questions pertinent to it. The teacher, with full access to the article text, responds by selecting relevant text spans from the article to answer the student's questions, focusing on providing information directly from the source rather than generating responses from memory.

% === [SOLID]
In the final approach, a single LLM is responsible for generating all utterances in a conversation.
For example, 
% SOLID~\cite{askari2024self} uses a LLM to create a dataset for intent-aware information seeking dialogues. The conversation seed for this paper includes topic information and an intent sequence. SOLID defines an instruction for each intent. To generate each turn, an instruction is automatically created based on dialogue history, current intent, and the background information, prompting the LLM to produce an utterance. While turn-by-turn generation is commonly used for creating CIS datasets, an extension of SOLID, named SOLID-RL, employs a one-go generation method. SOLID-RL, a variant of the original method, underscores the benefits of one-go generation, such as improved naturalness and consistency in the conversation, as well as faster generation speeds. Furthermore, SOLID-RL introduces a reinforcement learning (RL)-guided approach to train LLMs in a zero-shot manner, enabling the generation of conversation data in a single step.
\shorten{SOLID~\cite{askari2024self} uses an LLM to generate intent-aware information-seeking dialogues from conversation seeds containing topics and intent sequences, with each turn produced via intent-specific instructions conditioned on dialogue history. While the original method generates dialogues turn by turn, its extension SOLID-RL adopts a one-go generation approach, improving naturalness, consistency, and generation speed. SOLID-RL further introduces an RL-guided zero-shot framework to generate entire conversations in a single step.}
\note{ConvTrans~\cite{Mao22ConvTrans} introduces a conversational query rewriter that transforms ad-hoc queries in the graph into natural language conversational queries, thereby bridging the gap between different query forms. It then employs a specialized random walk sampling algorithm to generate pseudo search sessions from the query-transformed session graph.}

% % ==================== W / Wo fine-tuning ==================
% Beyond the generation methodology, another criterion for differentiating methods is whether the generator model operates \textbf{\textit{with or without fine-tuning}}. The "passage-grounded" approaches predominantly employ fine-tuned models because they involve components tasked with specific functions. Exceptionally, SimQuAC~\cite{abbasiantaeb2024let} utilizes an LLM with prompting for generation. Both the "intent flow grounded" and "starting point" strategies primarily rely on the prompting method. However, SOLID-RL~\cite{askari2024self} deviates by fine-tuning the model to facilitate the generation of multi-turn conversations in a single step.

\subsection{Quality Filtering}
\label{sec:cis:filter}

The primary goal of the filtering component is to remove conversations that lack key CIS features, particularly correctness and consistency, whether at the turn level or throughout the entire conversation. This step improves the quality of synthetic data and prevents the introduction of noise into downstream tasks. We categorize the filtering methods into two groups.

\subsubsection{Noise \& Lexical Filtering}
The first type primarily focuses on ensuring correctness and maintaining diversity. This filtering is typically applied using heuristic methods to check correctness~\cite{Wang23topdial, askari2024self, Joko2024LAPS}. For instance, TopDial~\cite{Wang23topdial} implements fact-checking and makes corrections based on domain-specific knowledge. 
%Meanwhile, MATHDIAL~\cite{macina2023mathdial} employs a safety filter to eliminate conversations that contain sensitive content.
\note{These filtering methods are commonly used for ODD and are particularly necessary for approaches that rely on LLMs for utterance generation. Since LLMs have a greater degree of freedom in generating utterances, they are more likely to produce inappropriate or irrelevant content. Therefore, Noise \& Lexical filtering plays a crucial role in ensuring the quality and consistency of the generated conversations.}

\subsubsection{Factuality Check}
In this type of filtering, the consistency of the generated utterances is evaluated. Document-driven methods predominantly include a factuality checking step to assess the performance of the utterance generator components. A notable technique is roundtrip consistency, initially introduced for QA generation~\cite{alberti2019synthetic} and later adapted for conversation data generation~\cite{wu2022dg2, kim2022simseek, Huang23CONVERSER}. This method uses a model to verify whether the answer span remains consistent with the original span that prompted the question.

DG2~\cite{wu2022dg2} incorporates passage selector and span extractor components in the utterance generation part. To implement filtering, DG2 develops additional passage selector and rationale extractor models. It then compares the turns generated by the two models and eliminates inconsistent samples. SimSeek~\cite{kim2022simseek} also employs additional answer extractor and question generator models but relaxes the filtering criteria from an exact match to word-level similarity (i.e., F1 score) to better accommodate the typically longer answers found in conversational QA as opposed to those in single-turn QA. CQAG-AR~\cite{HwangL22CQAGAR} focuses on the consistency between an answer and a question after the generation of each turn to avoid propagation errors in subsequent turns of a conversation. It introduces \emph{answer revision} module, which immediately revises the extracted answer after a question is generated. Similarly, MultiCQAG~\cite{hwang2022multiCQG} proposes a \emph{Hierarchical Answerability Classification} module to address the error propagation issue by determining whether a question can be answered based on the given passage. If a question is deemed unanswerable, the module substitutes the answer with "unknown" to prevent erroneous continuations of conversation.

% SimQuAC~\cite{abbasiantaeb2024let} incorporates \emph{Question Validation} and \emph{Answer Validation \& Regeneration} components to enhance the conversation's quality, generated by two LLMs. The question validation step ensures the structural integrity of the questions generated by the LLM, focusing on their syntactical correctness and coherence. The answer validation \& regeneration component reinforces the criteria for answer generation, prompting the teacher to produce responses that are consistent with the discussed topic. 
% 
\shorten{SimQuAC~\cite{abbasiantaeb2024let} improves LLM-generated conversations using \emph{Question Validation} and \emph{Answer Validation \& Regeneration} modules, ensuring syntactic correctness, coherence, and topic-consistent responses.}
SynDG~\cite{bao2023SynDG} implements a \emph{two-stage filtering} process including flow-level and conversation-level checks. At the conversation level, another T5 model~\cite{Raffel20t5} is fine-tuned for a task that involves predicting masked utterances, similar to roundtrip consistency. Additionally, another model is fine-tuned to verify the consistency of the dialogue flow before generating the conversation turns.

\note{Compared to the factuality-check filtering in ODD, both approaches share the ultimate goal of ensuring consistency between the turns of a conversation. However, in ODD, the focus is on verifying the consistency of the user persona or the coherence of selected turns originating from different datasets. In contrast, in CIS, the consistency is evaluated across conversation turns and question–answer pairs, as these interactions are grounded in a specific document.}

% \input{sections/6-generation-Intruction}
% \newpage
\section{Discussion}
% 1) compare synthetic datasets with human-annotated ones
% 2) adaptation to other domains

In this section we discuss the proposed conversational data generation methods from different aspects.

\begin{table}[t]
\centering
\setlength{\tabcolsep}{2.5pt}
\small
\caption{\note{Summary of human-created and synthetic datasets, including their tasks/domains and the number of samples.}}
\label{tab:datasets}
\begin{threeparttable}

\begin{tabular}{llc|llc} \hline
\multicolumn{3}{c|}{\textbf{Human-created Datasets}} & \multicolumn{3}{c}{\textbf{Synthetic Datasets}} \\\hline\hline
Dataset & Task & Size & Dataset & Task & Size \\\hline

\multicolumn{6}{l}{\textbf{Task-oriented Dialogue System}}\\\hline
DSTC2~\cite{kimDSTC2} & State Tracking & 2.5K           & Sim‑GEN~\cite{Shah2018M2M} & Movie tickets & 120K     \\
DialEdit~\cite{DialEdit} & Image editing & 129          & NeuralWOZ~\cite{kim2021neuralwoz} & Multi-domain & 6K \\
M2M~\cite{Shah2018M2M} & Restaurant \& Movie & 3K       & SimulatedChats~\cite{mohapatra2021simulatedchats} & Multi-domain & 10K \\
WoZ2.0~\cite{Woz2.0} & Restaurant Search & 1.2K         & IND~\cite{ahmad2023ina} & Negotiation Strategies & 4K \\ 
MultiWOZ~\cite{budzianowski2018multiwoz} & Multi-domain & 10K & DIALOGIC~\cite{li2022DIALOGIC} & Multi-domain & 8.4K  \\
SGD~\cite{Rastogi2020SGD} & Multi-domain & 16K \\ 
\hline
\multicolumn{6}{l}{\textbf{Open-Domain Dialogue System}} \\
\hline
DailyDialog~\cite{li2017dailydialog}  & Emotion \& Act Class. & 13.1K & AUGESC~\cite{Zheng23AugESC} & Emotional Support & 65K \\
Topical-Chat~\cite{Karthik19Topical} & Chit-chat & 11.3K              & BSBT~\cite{kim2022botstalk} & Blended Skill & 300K \\
ESConv~\cite{Zheng23AugESC}        & Emotional Support & 1.3K         & SODA~\cite{kim2022soda}     & Social Dialogue & 1.5M\\
ConvAI2~\cite{kim2022botstalk}     & Persona-based Chitchat & 20K     & PERSONACHATGEN~\cite{lee2022personachatgen} & Persona-based Chitchat & 1.6K  \\  
Wizard of Wikipedia~\cite{Dinan19wow}  & Knowledge-grounded & 22K     & Synthetic-Persona-Chat~\cite{Jandaghi23spc} & Persona-based Chitchat & 10.3K \\
Empathetic Dialogues~\cite{rashkin2019towardsEmpathetic} & Empathetic & 25K \\
Prosocial Dialog~\cite{kim2022soda}         & Social Dialogue & 58K \\
PERSONACHAT~\cite{zhang2018personachat}     & Persona-based Chitchat & 11K \\
\hline
\multicolumn{6}{l}{\textbf{Conversational Information Seeking}} \\
\hline
Doc2Dial~\cite{Feng2020Doc2Dial} & Goal-oriented & 4.8K                    & DG2~\cite{wu2022dg2} & Goal-oriented & 4.8K \\
OR-QuAC~\cite{choi2018quac} & CQA & 5.6K                                   & WikiDialog~\cite{Dai22Inpainting} & CQA & 11.4M \\
QReCC~\cite{Anantha21QReCC}   & CQA & 13.6K                                & WebDialog~\cite{Dai22Inpainting}  & CQA &  8.4M \\
CAsT 19-22~\cite{DaltonCAsT19, Dalton20CAsT, Owoicho22CAsT2022} & CQA & 93 & CQAG-AR~\cite{HwangL22CQAGAR} & CQA & 4.7K \\
% CAsT-20~\cite{Dalton20CAsT} & CQA & 25 \\
% CAsT-22~\cite{Owoicho22CAsT2022} & CQA & 18 \\
iKAT-23~\cite{Aliannejadi24iKAT} & CQA & 25                                & WIKI-SIMSEEK~\cite{kim2022simseek} & CQA & 213K \\
CoQA~\cite{HwangL22CQAGAR} & CQA & 8K                                      & CONVERSER~\cite{Huang23CONVERSER} & CQA & 31K \\
TopiOCQA~\cite{adlakha2022topiocqa} & CQA & 4K                             & SimQuAC~\cite{abbasiantaeb2024let} & CQA & 334 \\
MSDialog~\cite{Yang18MSDialog2} & CIS & 2.2K                               & ConvTrans~\cite{Mao22ConvTrans} & Conv. Search & 75K \\
MANtIS~\cite{Penha19MANtIS} & Conv. Search & 1.3K                          & SOLISpeak~\cite{askari2024self} & Conv. Search & 316K \\
DuRecDial~\cite{liu2021durecdial} & Conv. Recom. & 6K                      & TOPDIAL~\cite{Wang23topdial} & Conv. Recom. & 18K \\
CPCD~\cite{leszczynski2022conversational} & Conv. Recom. & 16K             & TtWMusic~\cite{leszczynski2022conversational} & Conv. Recom. & 1M \\
MG-ShopDial~\cite{Bernard23ShopDial} & E‑commerce & 64                     & ConvSDG~\cite{Mo24ConvSDG} & Conv. Search & 1.7K \\
LAPS~\cite{Joko2024LAPS} & PMCS\tnote{*} & 1.4K \\\hline

\hline
\end{tabular}
\begin{tablenotes}
\footnotesize
\item[*] Personalized Multi-Session Conversational Search
\end{tablenotes}
\end{threeparttable}
\end{table}

% ==========================

\note{\textit{\textbf{Human and syntactically  generated data}}: We list the human-curated and LLM-generated datasets for TOD, ODD, and CIS in Table~\ref{tab:datasets}, along with their corresponding tasks/domains and sample sizes, to provide an overview of the existing datasets. Generally, human-curated datasets contain fewer samples, while the size of synthetic datasets depends on the authors’ choices. Moreover, we observe that synthetic datasets have been created for a wide range of tasks and domains, whereas the creation of synthetic data often relies on the existence of human-curated data in the same domain or task~\cite{kim2022soda, askari2024self, Wang23topdial}. 
In general, synthetic data are useful in domains or tasks where there is a lack of conversational data or only a small amount of training data available. In such cases, other textual sources can be leveraged to generate a large amount of conversational data, thereby strengthening the training process. Another advantage of synthetic data is that it can introduce new challenges that are often absent in human-created data, such as topic shifts and lengthy conversations~\cite{kim2022botstalk, kim2022soda}.
}

\note{\textit{\textbf{Quantity and Quality}}: Existing studies mainly assess synthetic datasets based on two key dimensions: their quantity and quality.
Quantity refers to features such as the number of samples, number of turns, average answer length, and average question length. The quantity of generated samples often depends on the author’s choice and the task at hand~\cite{abbasiantaeb2024let, Dai22Inpainting}. For example, SOLID~\cite{askari2024self} generated 316K samples, which is 158 times larger than its original seed dataset. In contrast, SimQuAC~\cite{abbasiantaeb2024let} generates a controlled number of samples, matching the size of the original dataset. Other generation features, such as the number of turns and tokens, are typically controlled to align with the original dataset, as these characteristics naturally reflect human-generated conversations~\cite{kim2022botstalk, wu2022dg2}.}
\note{Quality of generated datasets is commonly assessed through metrics such as proactivity, coherence, personalization, target success rate, correctness, naturalness, and completeness~\cite{Wang23topdial, Joko2024LAPS, kim2022simseek}. For this evaluation, samples from both LLM-generated conversations and seed data are randomly selected and compared pairwise by either LLM evaluators or human annotators. For instance, TopDial~\cite{Wang23topdial} and SimQuAC~\cite{abbasiantaeb2024let} employed such evaluations and found that generated datasets achieved comparable or slightly higher win rates than the original datasets.
Another approach to evaluating dataset quality is to measure performance on downstream tasks, such as passage retrieval or answer generation~\cite{wu2022dg2}. A general conclusion across studies is that using only synthetic samples alone cannot outperform scenarios where human-generated samples are utilized. However, effectively combining original data with synthetic data can improve system performance on downstream tasks. This is because the challenges and nuances present in human-generated data remain difficult for models to fully capture, and synthetic data alone cannot fully replicate these complexities~\cite{abbasiantaeb2024let, askari2024self}.}

\note{\textit{\textbf{LLM-based vs non-LLM-based Methods:}}}
\note{In this section, we discuss the transition of methods from non-LLM-based approaches to those leveraging LLMs.}
\note{For TOD, early approaches primarily relied on rule-based systems and smaller sequence-to-sequence models. Methods like ABUS~\cite{li2017ABUS}, M2M~\cite{Shah2018M2M}, and SGD~\cite{Rastogi2020SGD} had access to predefined templates and schemas; neural approaches such as NUS~\cite{kreyssig2018neural}, HUS~\cite{gur2018HUS}, and VHDA~\cite{yoo2020VHDA} kept track of the dialogue states using neural architectures, with added modules for language generation. Introducing LLMs made a significant shift in how TOD generations are made, as this type of data relies a lot on exact information, such as restaurant opening hours and real geographical locations. The robustness of LLMs alongside their in-context learning and reasoning capabilities made it possible to generate both dialogue content and slot-value entries.}

\note{For ODD, one common approach in the absence of LLMs  is to mix existing datasets. In this case, the main consideration is maintaining consistency between consecutive turns that originate from different datasets~\cite{kim2022botstalk}. Another approach involves training a small model, specifically for the conversation data generation task~\cite{Lin22GCN}.
In the LLM-based methods for ODD, the process typically involves providing the LLM with a topic and description, and instructing it to generate a conversation sample~\cite{chen2023places, zhang2018personachat, chen2022weakly}. In some cases, in-context learning is applied, where the model is given examples and the initial part of an existing conversation, and is then asked to generate the remaining turns~\cite{Zheng23AugESC}.}

\note{For CIS, most document-driven methods rely on LMs for utterance generation. This is because the generation process is controlled by a sequence of documents, and producing utterances grounded in these documents can be effectively handled by smaller LMs~\cite{wu2022dg2, kim2022simseek, Huang23CONVERSER}. However, in sequence-grounded methods, LLMs are predominantly used. This is because such methods require converting a keyword or intent into a coherent utterance, a task that demands the reasoning capabilities of LLMs.}

\note{\textit{\textbf{Domain Adaptation:}}}
\note{For TOD, several methods show promise for generalization across different domains, although it remains a critical challenge as some approaches require additional domain-specific schemas, and the generations have to be grounded. SGD~\cite{Rastogi2020SGD} and M2M~\cite{Shah2018M2M} demonstrate cross-domain transfer by creating dynamic schemas across services and domains, with SGD demonstrating that many domains actually have an overlap of entities and attributes, making it more easily adaptable. However, this is highly dependent on how domains are semantically close to each other. Nevertheless, these methods need either access to pre-defined domain-specific ontologies and schemas, or having sufficient overlap with a new domain to make use of dynamic modeling of entities and their features. Given LLM models, generalizability seems more straightforward with in-context learning; however, there is still a need for domain-specific examples and a robust way of evaluating generations.}

\section{Biases and ethical Considerations}
%% Bias on synthetic dialog generation 
\note{Synthetic dialog generation can potentially introduce biases, such as underrepresentation of real-world topics, user needs, and intents~\cite{askari2024self, kim2022soda}. While many of the approaches reviewed in this work leverage real-world datasets for training a model to generate synthetic dialogs, overly specialized corpora may lead to issues of poor generalizations and generations that are not applicable to broader or unseen scenarios. }

%% How LLMs make it worse
\note{LLMs introduce additional biases and may raise ethical concerns by generating dialogs highly influenced by their parametric knowledge acquired during pretraining on often a wide range of unknown data. A primary concern is that LLMs tend to avoid generating content involving long-tail entities, focusing more on information about which it is certain and has been seen more during pretraining~\cite{long_tail}. A second challenge lies in generating hallucinated facts. While the generated dialogs can somehow be filtered and their factuality can be checked by comparison with ground truth statements, that often defeats the purpose of LLM-based generations and constrains creativity~\cite{hallucinations}. A last vital issue is that dialogs may be generated on undesirable subjects, given that LLMs inherently have socio-economic, cultural, racial and gender biases. Ideally, LLM generations would help to enrich the data to a calibrated distribution that can further help to eliminate these biases~\cite{etnic_biases}. However, that is hard to control and evaluate in mass generation.}

\section{Conclusion}~\label{sec:conclusion}
% In this survey paper, we extensively review the body of work on creating conversational data. Given the vast array of research related to data generation for dialogue systems, we narrow our focus to methods that generate complete multi-turn dialogues. We adopt the taxonomy of dialogue systems from previous surveys~\cite{Yang:2025:survey, Zamani2022CIS, yi2024survey, Jinjie2023Recent}, organizing our discussion around task-oriented dialog (ToD), open domain dialog (ODD) and conversational information seeking (CIS) dialogue systems.
% 
This survey paper reviews methods for generating conversational data, focusing specifically on approaches that produce complete multi-turn dialogues. Following prior taxonomies~\cite{Yang:2025:survey, Zamani2022CIS, yi2024survey, Jinjie2023Recent}, we organize the discussion around task-oriented dialog (ToD), open-domain dialog (ODD), and conversational information seeking (CIS) systems.

Within the domain of TOD, the advancement and effectiveness of conversational systems, for example in performing specific tasks like booking a flight or making a reservation, heavily rely on the generation of robust and factually accurate dialogues. To address these constraints, research has focused on developing sophisticated data generation and augmentation methodologies that leverage structures such as entities, entity attributes, dialog acts, and intents within complex graphical representations, such as Ontologies, Schemas, and Knowledge Graphs, ensuring the generation of diverse, plausible, and relevant dialogues.

In ODD, general-purpose chatbots such as ChatGPT and GPT-4 have exhibited high quality and performance primarily in general tasks and concepts, rather than in specialized domains~\cite{mallen2023trust, Maria24Agriculture}. Moreover, access to these models is usually restricted via APIs. To overcome these limitations and improve the performance of LLMs in specific domains and languages, such as emotional support~\cite{Zheng23AugESC}, Spanish intent detection~\cite{chen2023places}, and multi-skill conversations~\cite{kim2022botstalk}, data augmentation followed by fine-tuning has been suggested as an effective approach~\cite{Ovadia23Injection, Soudani24Fine}. Additionally, some crowdsourced datasets may contain outdated concepts and lack newer ones. Synthetic data generation has been proposed to update the knowledge stored in LLMs and mitigate these issues~\cite{lee2022personachatgen}.
% In the realm of ODD, it has been observed that general-purpose chatbots like ChatGPT and GPT-4, despite performing exceptionally well, are proprietary and access to them is restricted via APIs. To address this constraint, researchers have developed conversational datasets to train open-source, robust chatbots. Additionally, certain studies aim to enhance dialogue systems for specific challenges, domains, and tasks.

CIS systems represent a promising type of dialogue systems that encompasses a wide range of tasks, challenges, and applications. The significant number of studies proposing methods for generating conversational data suggests that CIS systems have attracted considerable interest. Various approaches to creating dialogue data have been explored, including document-driven and sequence-grounded methodologies for tasks such as intent prediction, negotiation systems, and recommendation systems. Overall, researchers aim to manage the conversation generation process to better respond to users' needs. They achieve this by defining control variables such as the initial topic utterance and the dialogue flow~\cite{abbasiantaeb2024let, askari2024self, Joko2024LAPS}.
% CIS systems have garnered considerable interest, with a significant amount of research dedicated to them. Various approaches to creating dialogue data have been explored, including document-grounded, intent-flow grounded, and data generation for instruction-tuning purposes. Moreover, these efforts target a diverse range of tasks, from general-purpose datasets centered around Wikipedia pages to more specialized objectives like intent prediction, negotiation systems, and recommendation systems that are target-oriented.

\section{Future Directions}~\label{sec:future_direction}

A crucial aspect of the conversation generation process is the control over the output data, which pertains to the extent to which the generation can be monitored and how many variables can be actively managed. Current research shows that the primary variables controlled are typically the topic and the dialogue flow. Despite the large volume of synthetic data produced, its quality often remains inadequate when evaluated through downstream tasks. This highlights the need to improve the quality of generated data so that even smaller datasets can yield results comparable to or better than those from human-generated data. Enhanced controllability could also increase data diversity, fairness, and safety~\cite{FACTS-IR}, reduce bias~\cite{Gerritse:2020:BCS}, improve factuality using uncertainty quantification methods~\cite{Soudani25Uncertainty, soudani2025uncertainty2, vashurin-etal-2025-benchmarking} and create datasets that pose more significant challenges for dialogue systems~\cite{samarinas2024simulating}. 

Another limitation of synthetic conversational data is the transferability limitation of dialogs across domains. Often models trained on synthetic data from one domain struggle to generalize to other settings due to differences in structure, required knowledge and intent distribution. However, synthetic approaches allow for generation in large quantities, leveraging the possibility of creating cross-domain datasets that can encompass multiple domains.
These capabilities are particularly valuable in developing proactive systems that involve clarification questions, dialogues with shifting targets, mixed-initiative interactions, and reasoning over complex questions centered around entities and knowledge graphs~\cite{Magdalena:2021:RLR, Joko:2021:CEL, Joko:2022:PEC}. Therefore, future research should enhance existing challenges of conversational AI such as knowledge groundedness, personalization, safety and bias, and focus on refining the quality and control of the generation process to ensure that the resulting datasets align more closely with the desired characteristics.

\note{Recent studies show that using LLM-generated dialogues as training data and relying on LLM-based evaluators can create a self-reinforcing loop in dialogue system development~\cite{Wataoka24Self, Panickssery24Evaluators}. LLM evaluators often display self-preference bias, consistently rating their own outputs higher than those of other models or humans~\cite{Balog:2025:LLMjudge, Dietz:2025:tropes, joko:2025:face}. This bias, driven by familiarity and lower perplexity, reduces response diversity and amplifies model weaknesses. As a result, dialogue systems trained and evaluated mainly on LLM-generated data may converge toward homogenized behaviors that reflect evaluator preferences rather than true quality improvements. Overcoming this self-reinforcement bias is an important direction for building more robust and unbiased dialogue systems.}

\note{Generating large-scale data for complex and personalized tasks, such as tutoring or modeling user personas, remains a challenge~\cite{Abbasiantaeb25Conversational, Mok25Exploring, Liu25One}. Recent work on scaling persona generation and personalization benchmarks still struggles to capture the evolving nature of user needs and long-context interactions. Tasks like personalized tutoring require adapting to individual learning styles, emotions, and progress, while persona-driven dialogues need consistent yet flexible trait modeling. Current pipelines often lack diversity, context continuity, and privacy-aware personalization. Developing scalable methods to produce rich, context-aware, and user-aligned data is an important future research.}
\bibliographystyle{ACM-Reference-Format}
\bibliography{main}

\clearpage
\appendix
\section{Examples Of Dialogue Types}
\begin{table}[h!]
    \centering
    % \renewcommand{\arraystretch}{1.1}
    % \caption{\note{Examples of Different Dialog Types}}
    \label{tab:dialog_types_examples}
    \begin{tabular}{@{}p{3cm}p{11cm}@{}}
        \toprule
            \textbf{Dialogue Type} & \textbf{Example Dialogue} \\
            \midrule
            \textbf{Task-Oriented (TOD)} &
            \textbf{User:} Book a restaurant in Orlando for 4 people. \newline
            \textbf{System:} What type of food and price range should I look for? \newline
            \textbf{User:} I'd like a moderately priced taiwanese restaurant. \newline
            \textbf{System:} I found a restaurant in the center of Orlando. They have availability at 8PM and 8:30PM. \\
        \addlinespace
            \textbf{Open-Domain (ODD)} &
            \textbf{User 1:} Hi there! Are you looking for someone to take care of your pet while you’re out of town? \newline
            \textbf{User 2:} Yes, I’m going on vacation for a week and I need someone to watch my cat. \newline
            \textbf{User 1:} That’s no problem at all! I love taking care of animals, and I have experience with cats. I’ll be happy to take care of your cat while you’re gone. \newline
            \textbf{User 2:} That sounds great. How much do you charge? \\
            
        \addlinespace
            \textbf{Information-Seeking (IS)} &
            \textbf{User:} I am interested to learn more about taiwanese cousine, it's my favourite. \newline
            \textbf{System:} Of course! I can help with that. What would you like to know? \newline
            \textbf{User:} I really like the five spice powder. Can you tell me what it is made of, in what proportions and what are common areas each gets cultivated? \newline
            \textbf{System:} Five-spice powder typically includes star anise, cloves, Chinese cinnamon (cassia), Sichuan peppercorns, and fennel seeds. The exact proportions vary by region and recipe, but a common blend might use equal parts of each, or slightly more star anise and cinnamon for warmth and sweetness. In terms of cultivation, .. \\
        \bottomrule
    \end{tabular}
\end{table}

\section{Examples Of Information Seeking Conversational Data Generation}
\begin{figure}[h]
    \centering
    \includegraphics[width=0.94\textwidth]{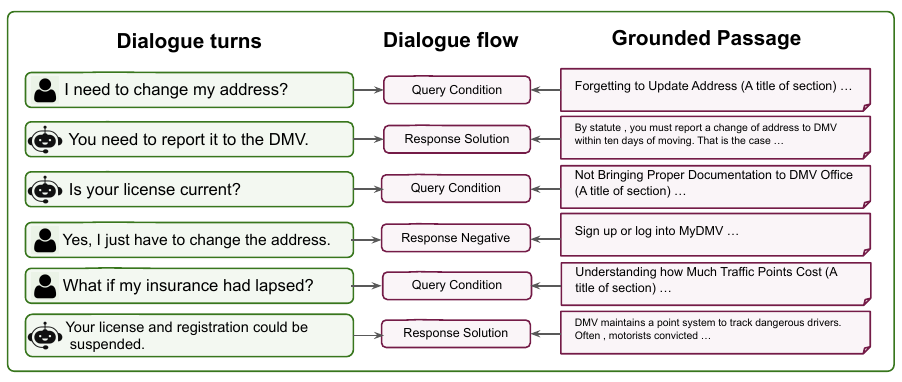}
    \caption{Conversation Seed and Corresponding Dialogue from the Doc2Dial Dataset~\cite{Feng2020Doc2Dial}. This dataset is dedicated to generating conversations based on governmental documents. The displayed example features a conversation seed comprising a  title, a document, and a sequence of dialogue acts. In the corresponding conversation sample, each turn is labeled with a specific dialogue act.}
    \label{fig:cis_seed_example}
\end{figure}

\section{Examples Of Task-oriented Conversational Data Generation}
\begin{figure}[H]
    \centering
    \includegraphics[width=0.94\textwidth]{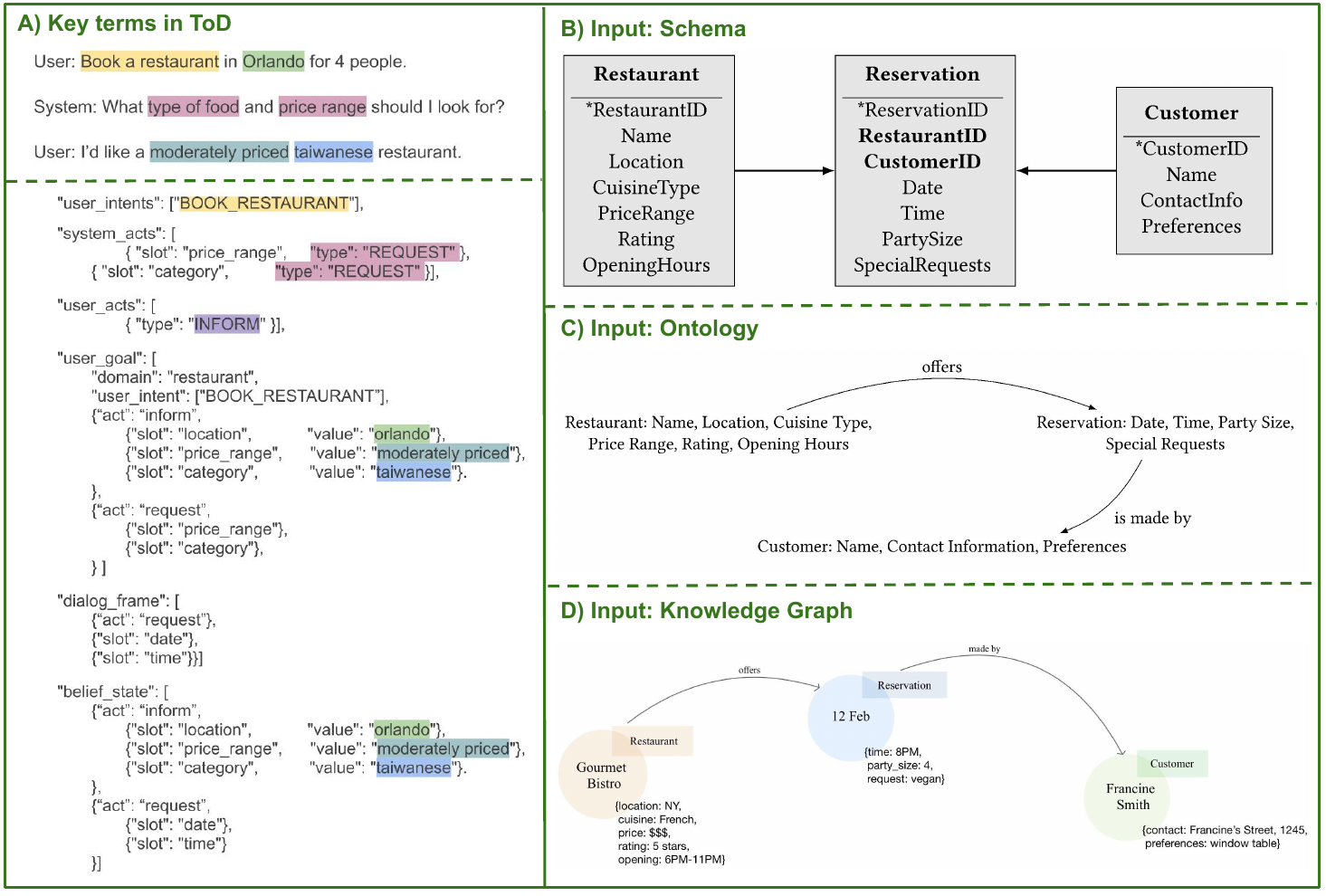}
    \caption{This figure illustrates several aspects of a task-oriented dialogue (TOD): (A) key terms, slots, and associated values used in a TOD dialogue; (B) an example entity–attribute schema for the task of restaurant table reservation; (C) an example ontology with relational information for the same task; and (D) an example knowledge graph (KG) constructed for this task.}
    \label{fig:tod_structure}
\end{figure}

\section{Examples Of Open Domain Conversational Data Generation}
\begin{figure}[htbp]
    \centering
    \includegraphics[width=0.94\textwidth]{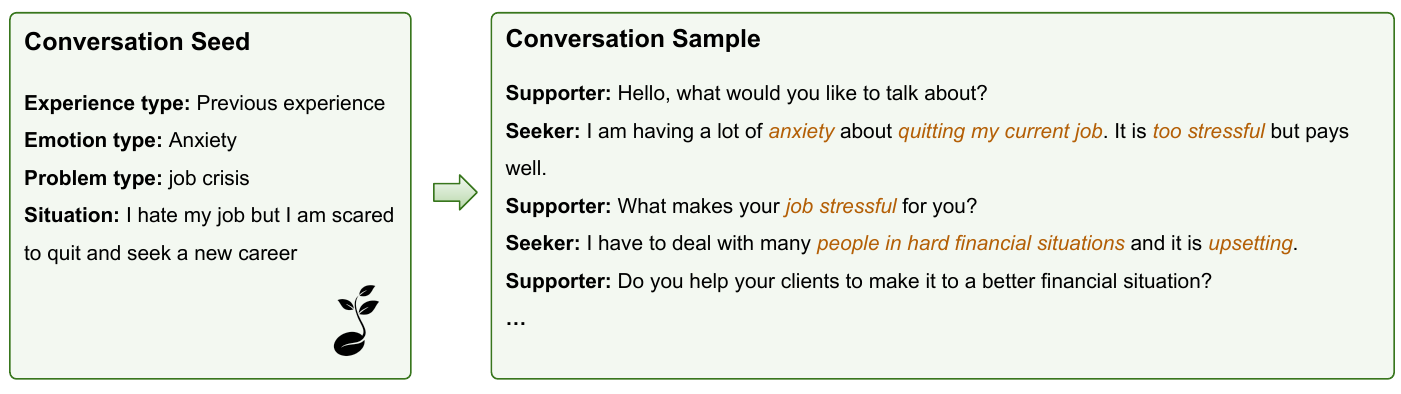}
    \caption{Conversation seed and the corresponding dialogue from the ESConv Dataset~\cite{Liu21ESConv}. This dataset focuses on emotional support conversations. The example  shows a conversation seed that outlines the emotional problem and current situation of a support seeker. The corresponding dialogue develops around the information provided in the seed.}
    \label{fig:odd_conv_seed_example}
\end{figure}

\begin{figure}[h]
    \centering
    \includegraphics[width=0.94\textwidth]{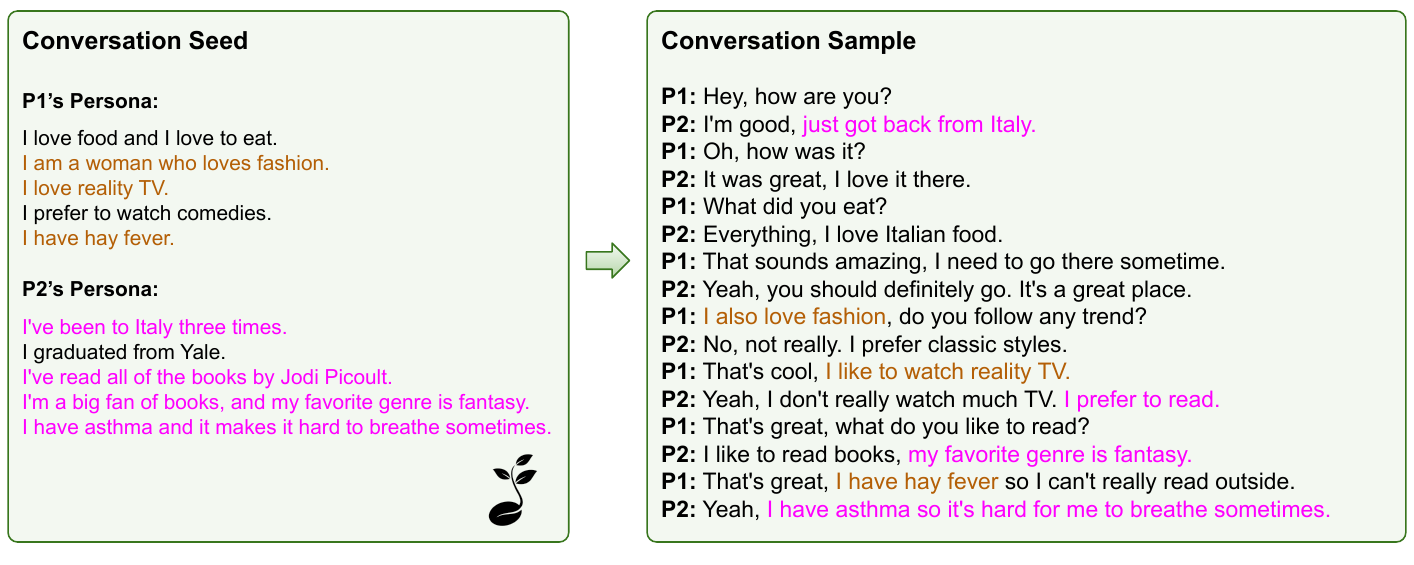}
    \caption{Examples of generated dialogue from PersonaChatGen~\cite{lee2022personachatgen} based on a conversation seed containing two user profiles. Utterances in \textcolor{orange}{orange} are directly related to Person 1, while those in \textcolor{magenta}{magenta} relate to Person 2.}
    \label{fig:odd_utr_example}
\end{figure}

\end{document}